\documentclass[lettersize,journal]{IEEEtran}
\usepackage{amsmath,amsfonts}
\usepackage{algorithmic}
\usepackage{algorithm}
\usepackage{array}
\usepackage{textcomp}
\usepackage{stfloats}
\usepackage{url}
\usepackage{verbatim}
\usepackage{graphicx}
\usepackage{cite}

\usepackage{amssymb}
\newcommand\norm[1]{\left\lVert#1\right\rVert}
\newcommand*{\prob}{\mathsf{p}}
\DeclareMathOperator*{\argmax}{argmax} 
\DeclareMathOperator*{\argmin}{argmin} 
\usepackage{xcolor,colortbl}
\usepackage[caption=false]{subfig}
\usepackage{threeparttable}
\usepackage{multirow}
\usepackage{hhline}
\usepackage{tikz}
\usepackage[switch]{lineno}
\def\checkmark{\tikz\fill[scale=0.4](0,.35) -- (.25,0) -- (1,.7) -- (.25,.15) -- cycle;}

\newcommand{\revise}[1]{{\color{black}{#1}}}
\newcommand{\revisetwo}[1]{{\color{black}{#1}}}

\begin{document}

\title{Graph-based Thermal-Inertial SLAM with Probabilistic Neural Networks}

\author{Muhamad Risqi U.~Saputra,
        Chris Xiaoxuan~Lu,
        Pedro Porto B. de~Gusmao,
        Bing~Wang,
        Andrew~Markham,
        and~Niki~Trigoni

\thanks{This work was supported by EPSRC grant entitled "ACE-OPS: From Autonomy to Cognitive assistance in Emergency OPerationS" (EP/S030832/1) and by the US National Institute of Standards and Technology (NIST) grant entitled "Pervasive, Accurate, and Reliable LBS for Emergency Responders" (No. 70NANB17H185). The authors would like to thank John G. Rogers III and Arthur Schang from CCDC Army Research Laboratory (ARL) USA for their assistance on using SubT-tunnel dataset. Most parts of this work conducted when M. R. U. Saputra was a postdoc at University of Oxford. \textit{(Corresponding author: M. R. U. Saputra.)}}
\thanks{M. R. U. Saputra is with Monash University, Indonesia. Email: risqi.saputra@monash.edu.}
\thanks{C. X. Lu is with the School of Informatics at the University of Edinburgh, United Kingdom (UK). Email: xiaoxuan.lu@ed.ac.uk.}
\thanks{Pedro P. B. de Gusmao is with the Department of Computer Science and Technology at the University of Cambridge, United Kingdom (UK). Email: pp524@cam.ac.uk.}
\thanks{B. Wang, A. Markham, and N. Trigoni are with the Department of Computer Science, University of Oxford, United Kingdom (UK). Email: \{bing.wang, andrew.markham, niki.trigoni\}@cs.ox.ac.uk.}
}

\markboth{Journal of \LaTeX\ Class Files,~Vol.~14, No.~8, August~2021}%
{Shell \MakeLowercase{\textit{et al.}}: A Sample Article Using IEEEtran.cls for IEEE Journals}

\IEEEpubid{0000--0000/00\$00.00~\copyright~2021 IEEE}

\maketitle

\begin{abstract}
Simultaneous Localization and Mapping (SLAM) system typically employ vision-based sensors to observe the surrounding environment. However, the performance of such systems highly depends on the ambient illumination conditions. In scenarios with adverse visibility or in the presence of airborne particulates (e.g. smoke, dust, etc.), alternative modalities such as those based on thermal imaging and inertial sensors are more promising. In this paper, we propose the first complete thermal-inertial SLAM system which combines neural abstraction in the SLAM front end with robust pose graph optimization in the SLAM back end. We model the sensor abstraction in the front end by employing probabilistic deep learning parameterized by Mixture Density Networks (MDN). Our key strategies to successfully model this encoding from thermal imagery are the usage of normalized 14-bit radiometric data, the incorporation of hallucinated visual (RGB) features, and the inclusion of feature selection to estimate the MDN parameters. To enable a full SLAM system, we also design an efficient global image descriptor which is able to detect loop closures from thermal embedding vectors. \revise{We performed extensive experiments and analysis using three datasets, namely self-collected ground robot and handheld data taken in indoor environment, and one public dataset (SubT-tunnel) collected in underground tunnel. Finally, we demonstrate that an accurate thermal-inertial SLAM system can be realized in conditions of both benign and adverse visibility.}
\end{abstract}

\begin{IEEEkeywords}
Thermal-inertial SLAM, loop closure detection, probabilistic deep neural networks, pose graph optimization.
\end{IEEEkeywords}

\section{Introduction}
\IEEEPARstart{S}{imultaneous} Localization and Mapping (SLAM) is an important task in robotics and autonomous systems. It enables a mobile agent to explore an unknown environment by simultaneously estimating the position of the agent whilst constructing a representation of the environment, termed a \textit{map}. This task is a precursor to many other robotic tasks such as navigation, exploration, or manipulation, making accurate SLAM estimation a fundamental need for autonomous systems.

A SLAM framework typically consists of a \textit{front end} and a \textit{back end}. The front end acquires sensor data and transforms it into an abstraction that is more amenable for inference, while the back end estimates the states of the agent given the abstracted data from the front-end. The back end is also responsible for optimizing the agent states and generating a globally consistent representation of the environment \cite{cadena2016past, li2019development}.

Most front ends in SLAM systems utilize range (e.g. depth, Lidar) or vision (RGB) sensors to sense the surrounding environment. Notable examples include ORB-SLAM \cite{mur2015orb} and LOAM \cite{zhang2014loam} which employ RGB and Lidar sensors respectively for their SLAM front end. While these range- and vision-based SLAM systems can generally work well in a wide range of applications, their performance largely depends on the benign visibility. When it comes to the adverse illumination conditions and/or in the presence of airborne particulates (e.g. dust, soot, smoke, etc.), using existing range and vision sensors for SLAM estimation is problematic. For instance, it is widely known that RGB cameras cannot operate in darkness while depth cameras are sensitive to glare and strong illumination \cite{debeunne2020review, ruppelt2016stereo, khattak2019keyframe}. The same visibility issues also applies to RGB, depth, and even Lidar sensors when operating in environments with airborne particulates \cite{bijelic2019seeing} or thick fog/mist. In contrast, thermal imaging cameras are not affected by illumination conditions and the presence of most airborne particulates \cite{brunner2014perception}. Instead, they capture the Long Wave Infrared (LWIR) data emitted from  objects in the environment. These advantages make thermal imaging cameras a viable alternative modality for SLAM application in visually-denied environments.
\IEEEpubidadjcol 

However, realizing a full thermal SLAM system comes with a set of challenging tasks. One of the most fundamental is how to abstract or encode the thermal data so as to maximally aid the graph optimization process. This is an intrinsically challenging task as thermal cameras capture the temperature profile of the environment instead of environmental appearance and geometry. The problem is even more pronounced with the fact that the re-scaled 8-bit resolution of thermal data has lower contrast, making standard feature matching and data association difficult. Moreover, thermal cameras periodically require suspension of camera operation for approximately 0.5-1 second to perform Non-Uniformity Correction (NUC) (also known as Flat Field Correction (FFC) in other literature \cite{mouats2015thermal, delaune2019thermal}) in which a uniform temperature is presented to the sensor to estimate the fixed-pattern noise correction parameters. Together, these issues mean that traditional methods developed for other optical sensors fall short in the typical front-end abstraction pipeline (e.g. feature extraction, data association, estimating an odometry prior, etc.).

The past decade has witnessed the rapid development of Deep Neural Networks (DNN) as a strong non-linear function approximator. It has been seen in the recent works that DNNs can be successfully used in visual odometry \cite{wang2017deepvo, saputra2019learning, li2019net, liang2019salientdso} and (re-) localization estimation \cite{clark2017vidloc, wang2019atloc, barsan2018learning}. We therefore hypothesize that one can model the abstraction or encoding of thermal data for a SLAM front end using DNN. In particular, by employing a type of probabilistic neural network, i.e. Mixture Density Networks (MDN), we can fully model the front end by constructing both odometry and loop closure constraints along with their covariance as a metric of uncertainty. In this way a more traditional back end graph-based optimizer can be used to generate a global trajectory. In a nutshell, our key and novel insights in building a reliable pose graph from thermal imagery include the usage of normalized radiometric (14-bit resolution) thermal data to avoid re-scaling, the incorporation of hallucination networks as complementary information \cite{saputra2020deeptio}, the inclusion of selective fusion module \cite{chen2019selective} which filters out reliable features, and the use of a probabilistic DNN. We also present a novel approach to neural loop closure estimation. Combined with outlier rejection in the back end to filter noisy loop closure constraints, we demonstrate that is possible to achieve a complete thermal-inertial SLAM system which produces globally consistent trajectory estimation, in spite of the above mentioned challenges.

The work described in this article builds on our previous work in \cite{saputra2020deeptio} which presented the first system for deep thermal-inertial odometry. The new contributions here can be summarized as follows:
\begin{itemize}
    \item We demonstrate the first complete thermal-inertial SLAM (TI-SLAM) system in the literature, which combines robust pose graph optimization in the back end with neural abstraction in the front end generated by probabilistic neural networks.
    \item We construct odometry and loop closure constraints in the pose graph by using a Mixture Density Network (MDN) parameterized through hallucination and feature selection network given normalized 14-bit radiometric thermal data as the input. We also combine the odometry network with IMU measurements to increase robustness in unknown scenes or when the thermal imaging is performing NUC calibration. 
    \item  We present an efficient global descriptor-based neural loop closure detection based on thermal embedding vectors output by a DNN.
    \item \revise{We perform extensive experiments and analysis under both benign and poorly-illuminated conditions on in-house ground robot and handheld data (self-collected), and on a public ground robot data (SubT-tunnel) taken in underground tunnel. The code and in-house datasets are released to the community.\footnote{https://github.com/risqiutama/ti-slam}}
\end{itemize}

\section{Related Work}
\subsection{Conventional Visual SLAM}
Visual SLAM was originally solved by filtering algorithm \cite{davison2007monoslam}. Notable examples include MonoSLAM \cite{davison2007monoslam} and its variants \cite{sunderhauf2007using, holmes2008square, civera2008inverse}, in which every frame is processed by Extended Kalman Filter (EKF) to jointly estimate the camera pose and landmark locations. However, due to the nature of EKF algorithm which accumulates linearization errors across multiple frames, keyframe-based Bundle Adjustment (BA) approach is more widely used in the past decade since it has been shown to be more accurate than filtering \cite{strasdat2012visual}. Prominent examples from this category include PTAM \cite{klein2007parallel} and ORB-SLAM \cite{mur2015orb} which employ point-based features in the front end or PL-SLAM \cite{pumarola2017pl} which utilizes line segment as the front end abstraction. These keyframe-based BA methods typically integrate hardware and algorithmic advances in the past decade by incorporating parallel computing, statistical model selection, loop closures detection based on bag-of-words place recognition, local BA, or other graph optimization approaches. However, despite their great performances in particular scenarios, these model based approaches are very sensitive to outliers (e.g. spurious correspondences, dynamic objects, etc.) \cite{saputra2018visual} and easily lose tracks when the environment has limited hand-engineered features \cite{wang2018end}.

\subsection{Deep Networks in the Context of SLAM}
In the last couple of years, there are many works that aim to replace the SLAM front end with learning-based approaches. The learning-based approaches, especially based on DNN, are typically more robust as it does not rely on point or line features, but directly learn to solve the task from abundant data. Among these approaches, some deals with feature correspondences \cite{detone2017toward, sarlin2020superglue}, some with odometry \cite{zhou2017unsupervised, liang2019salientdso, saputra2019learning, almalioglu2019ganvo}, global re-localization \cite{kendall2015posenet, wang2019atloc}, and place recognition \cite{arandjelovic2016netvlad, chen2017deep} or loop closure detection \cite{zhang2017loop}. Nevertheless, the developed system is secluded from each other and is not trivial to be combined together as a single SLAM system.

\subsubsection{Odometry Estimation} The first work on DNN based approach for Visual Odometry (VO) is pioneered by DeepVO \cite{wang2017deepvo} which models the camera pose estimation as an end-to-end pose regression problem. This was then followed by incorporating inertial (e.g. VINet \cite{clark2017vinet}, SelectFusion \cite{chen2019selective}), training the network by self-supervision (e.g. SfMLearner \cite{zhou2017unsupervised}, GANVO \cite{almalioglu2019ganvo}, etc.), or combining together model-based and deep learning-based approaches (e.g. SalientDSO \cite{liang2019salientdso}). However, none of them address thermal camera system except our work in \cite{saputra2020deeptio}.

\subsubsection{Place Recognition} NetVLAD \cite{arandjelovic2016netvlad} is a prominent place recognition algorithm which aggregates the statistics of local descriptors by computing the sum of residuals for each visual word. This approach was then typically improved by making it more robust across different environmental conditions by learning condition- and viewpoint-invariant features \cite{chen2017deep} or learning geometric features through depth generation \cite{piasco2019learning}. Existing work on loop closure detection or place recognition using thermal camera is typically performed by using standard feature-based approaches (e.g. FAB-MAP \cite{cummins2008fab}) and is aided by other modalities such as RGB camera \cite{maddern2012towards}.

\subsubsection{Global Relocalization} In the context of SLAM, global relocalization can be used to construct loop closure constraints. PoseNet \cite{kendall2015posenet} is a pioneer work in this category which regresses the global camera pose given a single image as the input. This was then followed by incorporating an attention mechanism \cite{wang2019atloc}, enforcing temporal information \cite{clark2017vidloc}, or fusing it with an additional sensor through variational inference \cite{zhou2021vmloc}. While deep global relocalization can be used as loop closure constraints, recent work \cite{sattler2019understanding} observes that it cannot generalize in unseen scenario as they implicitly save the `map' of the environment within the network. Different from the previous absolute pose regression methods, relative pose regression methods identifies the nearest neighbours in database images to the query image and recovers the relative pose between the reference images and the query. Specifically, NN-Net~\cite{laskar2017camera} utilises a neural network to estimate the pairwise relative poses between the query and the top N-ranked references. A triangulation-based fusion algorithm coalesces the predicted N relative poses and the ground truth of 3D geometry poses, and the absolute query pose can be naturally calculated. Furthermore, RelocNet~\cite{balntas2018relocnet} additionally exploits a frustum overlap loss to assist the learning of global descriptors that are suitable for camera localization. Motivated by these, CamNet~\cite{ding2019camnet} applies a two-stage retrieval, image-based coarse retrieval and pose-based fine retrieval, to select the most similar reference images for the finally precise relative pose estimation. We take this approach to construct loop closure constraints by first finding similar images in the sequence and then extracting relative poses between detected loop pair.

\subsubsection{SLAM} Recently, researchers have started to combine existing works on odometry, relocalization, and loop closure detection in a complete SLAM system. DeepSLAM \cite{li2020deepslam}, for example, combines self-supervised deep learning based monocular visual odometry with pose graph optimization. The system consists of three main modules in which each of them deals with odometry (Tracking-Net), mapping (Mapping-Net), and loop closure detection (Loop-Net). Despite their great performances in public visual odometry benchmark, there is no uncertainty estimation in the odometry and loop closure constraints, making it less flexible to balancing the constraints or inspecting failure modes. DeepFactors \cite{czarnowski2020deepfactors} is another example which tries to combine deep learning and factor graph optimization. The system was trained to learn a compact depth map representation for dense visual SLAM system. However, they only demonstrate the system in a small indoor environment (e.g. ScanNet dataset). Finally, despite some emerging works on combining model-based and deep learning-based approaches, none of them address thermal camera system.

\subsection{Thermal Odometry and SLAM}
\label{sec:related_work}
Realizing odometry and SLAM estimation using thermal imaging system remains a challenging problem due to the nature of thermal cameras which capture the heat distribution from the observed environment instead of the appearance and geometry. Nevertheless, some efforts have been made to construct thermal odometry, although it has been used for relatively short distances or yields sub-optimal performance compared to RGB-based odometry. Mouats et al. \cite{mouats2015thermal} utilized a Fast-Hessian feature extractor to estimate stereo thermal odometry for UAV tracking. Borges and Vidas \cite{borges2016practical} designed a practical thermal odometry by employing an automatic procedure to determine the correct time to perform the NUC operation. Nevertheless, the system can only work in outdoor scenario as it requires road lane estimation to compute the scale of the prediction.

Recent work on thermal odometry typically fused together thermal imaging systems with other modalities. Delaune et al. \cite{delaune2019thermal} combined thermal and inertial sensors for UAV tracking by using an EKF algorithm. They showed that by employing FAST and KLT tracker, the thermal-inertial odometry can work well during day and night. Similarly, Khattak et al. \cite{khattak2019keyframe, khattak2020keyframe} also construct thermal-inertial odometry for UAV tracking by using keyframe-based direct approach which minimizes radiometric error between two adjacent frames. They used raw radiometric data instead of the normalized grayscale data to avoid difficult data association as the scene dynamically changes based on the environment temperature.

Despite some work on thermal-inertial odometry estimation, to the best of our knowledge, there is no published work on thermal-inertial SLAM to date. Vidas and Sridharan \cite{vidas2012hand} realized a hand-held thermal SLAM by employing FAST-based feature tracking in the front end and bundle adjustment-based optimization in the back end. However, despite the claim of being a SLAM system, it is not a full SLAM system in a sense that there is no loop closure module which is used in state-of-the-art SLAM frameworks to generate a consistent trajectory and map (e.g. ORB-SLAM \cite{mur2015orb}, LSD-SLAM \cite{engel2014lsd}). Moreover, without an environment agnostic sensor like IMU, it is difficult to achieve robust estimation in an arbitrary environment. Shin and Kim \cite{shin2019sparse} recently proposed feature-based lidar-thermal SLAM. They enhanced thermal data with sparse range measurement from lidar to improve the scale estimation of the system. However, they demonstrated their system for operation in an autonomous car which typically has more thermal gradients than in indoor scenario (e.g. corridor with planar walls). Furthermore, all these works utilize a hand-engineered feature extractor which may lose track in environments with limited thermal gradients.

\begin{figure}
    \centering
    \includegraphics[width=0.8\columnwidth]{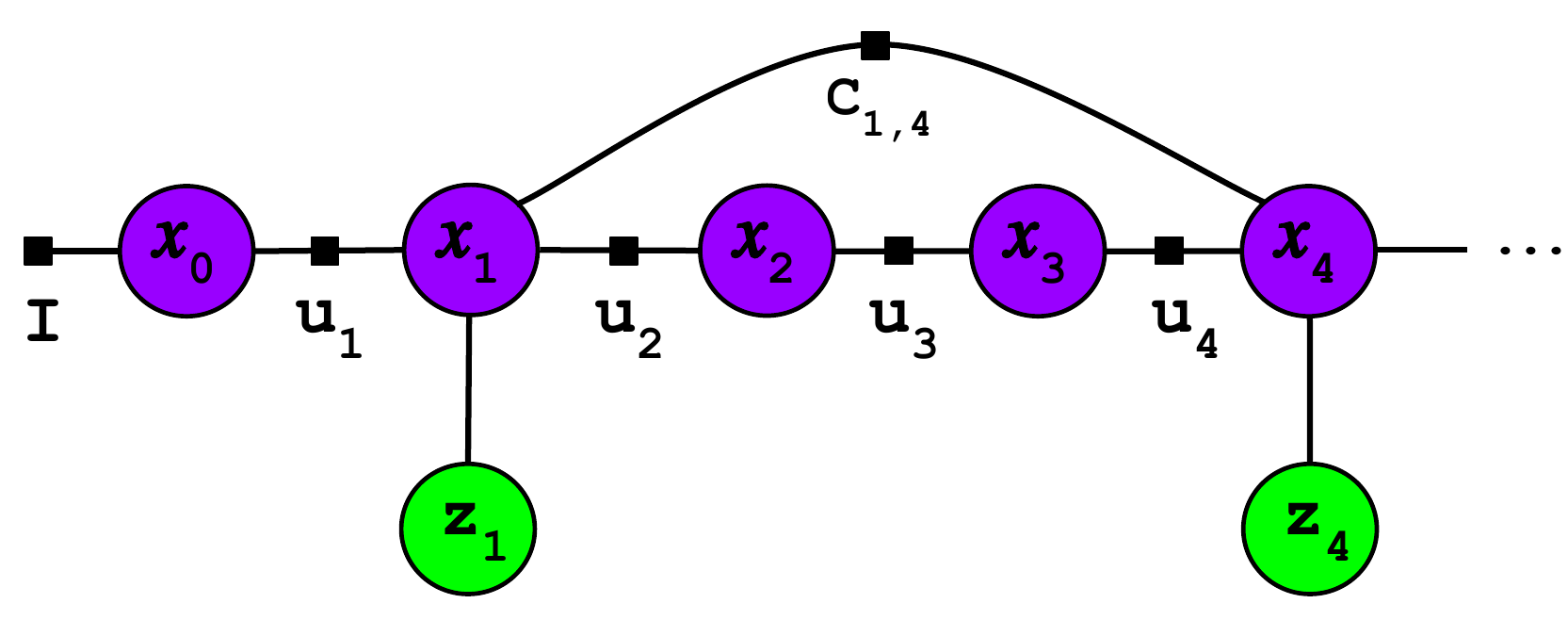}
    \vspace{-0.4cm}
    \caption{Factor graph representation of TI-SLAM.}
\label{fig:factor_graph}
\end{figure}

\section{SLAM Problem Formulation}
\label{sec:slam_formulation}
\subsection{Maximum a Posteriori (MAP) with Probabilistic Neural Networks}
It is widely known in the robotic community that we can solve the SLAM problem using a graph-based formulation. In this formulation, the SLAM estimation problem is simplified by abstracting the raw sensor measurements into edges in the graph \cite{grisetti2010tutorial}. To solve a graph-based SLAM problem, a Maximum a Posteriori (MAP) approach is typically employed. Let $\mathcal{X} = \{ x_i: i = 1, ..., m \}$ be an unknown variable (e.g. trajectory of the agent as discrete poses) that we want to estimate. Given a set of sensor measurements $Z = \{ z_i: i = 1, ..., m \}$ such that $z_i$ can be expressed as $z_i = h_i(x_i) + \epsilon_i$ where $h_i(.)$ and $\epsilon_i$ are measurement model and measurement noise respectively, we can compute $\mathcal{X}$ by estimating the assignment variables of $\mathcal{X}^*$ that yields the maximum of the posterior $\prob(\mathcal{X} | Z)$ as in the following equation \cite{cadena2016past} when the probability of each measurement is the same
\begin{align}
\mathcal{X}^* & \doteq \argmax_{\mathcal{X}} \prob ( \mathcal{X} | Z ) = \argmax_{\mathcal{X}} \prob(Z | \mathcal{X}) \prob(\mathcal{X}) .
\label{eq:MaP}
\end{align}
Note that the equality in Eq. (\ref{eq:MaP}) follows the rule in the Bayes theorem. Eq. (\ref{eq:MaP}) can then be factorized into the following form by assuming that the measurement $Z$ are independent
\begin{align}
\mathcal{X}^* = \argmax_\mathcal{X} \prob(\mathcal{X}) \displaystyle \prod_{i=1}^m \prob(z_i | \mathcal{X}) ,
\label{eq:MaP_factorized}
\end{align}
while both $\prob(\mathcal{X})$ and $\prob(z_i | \mathcal{X})$ are the \textit{factors} in the factor graph representation which encodes the probabilistic constraints among the nodes. In order to make Eq. (\ref{eq:MaP_factorized}) more explicit, we can assume that the measurement follows a Gaussian distribution with a zero-mean $\epsilon$ and information matrix $\Omega$ (the inverse of covariance matrix). Then, the likelihood function $\prob(z_i | \mathcal{X})$ will have the following form
\begin{align}
\prob(z_i | \mathcal{X}) \propto \text{exp} \left( -\frac{1}{2} \norm{h_i(x_i) - z_i }^2_{\Omega_i} \right) ,
\label{eq:MaP_likelihood}
\end{align}
where $\norm{e}^2_{\Omega} = e^{\intercal} \Omega e$. Since maximizing the posterior is essentially the same as minimizing the negative log likelihood, then the MAP estimation in Eq. (\ref{eq:MaP_factorized}) becomes
\begin{align}
\mathcal{X}^* & = \argmin_\mathcal{X} - \text{log} \left( \prob(\mathcal{X}) \displaystyle \prod_{i=1}^m \prob(z_i | \mathcal{X}) \right) \\
& = \argmin_\mathcal{X} \sum_{i=1}^{m} \norm{h_i(X_i) - z_i }^2_{\Omega_i} , 
\label{eq:MaP_neg_log}
\end{align}
where $h_i(.)$ represents a non-linear function. Eq. (\ref{eq:MaP_neg_log}) is widely known as a non-linear least square optimization problem and can be solved by using Gauss-Newton or Levenberg-Marquardt algorithm. Note that we omit $\prob(\mathcal{X})$ in Eq. (\ref{eq:MaP_neg_log}) since it is usually uninformative (e.g. modeled as uniform distribution) or does not contribute in determining the optimized value. 

\begin{figure*}
    \centering
    \vspace{-0.5cm}
    \includegraphics[width=1.95\columnwidth]{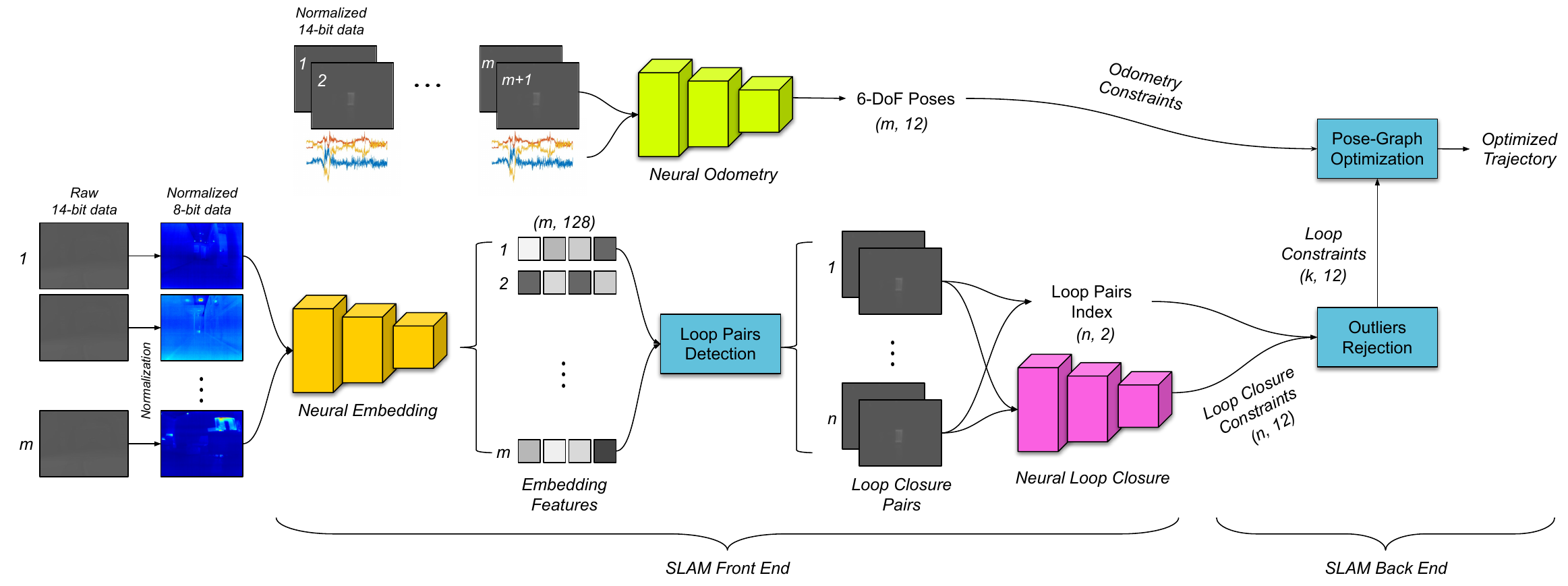}
    \vspace{-0.3cm}
    \caption{The high-level architecture of TI-SLAM. The front end is abstracted by probabilistic neural networks while the back end employs robust second order-based graph optimization.}
\label{fig:ti_slam_architecture}
\end{figure*}

To minimize Eq. (\ref{eq:MaP_neg_log}), in our formulation, we encode $h_i(.)$ with a deep, probabilistic neural network that estimates both mean and covariance. Then, from the perspective of the MAP estimator, our approach can be viewed as replacing the likelihood estimation with an abstraction from a deep neural network. From the optimization perspective, our deep network can also be viewed as the initial guess for the optimization. In this sense, we could combine the well-established formulation of SLAM problem with recent advances in DNNs as a strong non-linear function estimator to model a better abstraction of the sensor measurement.

To model the likelihood estimation with a deep neural network, we employ a Mixture Density Network (MDN) \cite{bishop1994mixture} which has been shown to work well for the camera (RGB) re-localization problem \cite{clark2017vidloc}. MDN allows the network to estimate a multi-modal posterior distribution which maps well to our problem of estimating a SLAM posterior from multi-modal sensor data. In MDN, the output is composed of a Gaussian Mixture Model (GMM) and the networks predict the GMM parameters mean $\mu_{k}$ and variance $\sigma_{k}$ where $k=1,...,K$ are indices of each Gaussian component $\mathcal{N}(\mu_{k},\sigma^{2}_{k})$. Then, given sensor measurement $Z$, the posterior $\prob(\mathcal{X} | Z)$ becomes
\begin{align}\
  \label{eq:mdn_estimates}
  \prob(\mathcal{X}|Z) = \sum_{k=1}^{K} \alpha_{k}(Z) \mathcal{N}(\mathcal{X}|\mu_{k}(Z),\sigma^{2}_{k}(Z)), 
\end{align}
where $\alpha_{k}$ are mixing coefficients constrained by $\sum_{k=1} \alpha_{k} = 1$ which is typically achieved by using a softmax function and learnt during training. Note that for training, we minimize the negative log-likehood of Eq. (\ref{eq:mdn_estimates}) such that $\mathcal{X}^* = \argmin_\mathcal{X} - \text{log} \left( \prob(\mathcal{X}|Z) \right)$. As we only estimate the variance instead of the full covariance, we assume that the output prediction (6-DoF poses) is independent of each other. This assumption has been used in \cite{wang2018end, kendall2016modelling} as well.

\subsection{SLAM Optimization Objectives}
Eq. (\ref{eq:MaP_neg_log}) can be interpreted as a general optimization objective for graph-based SLAM. In our implementation, we use a variant of this version which is called \textit{pose-graph} SLAM. \revise{In pose-graph SLAM, the variables to be inferred are the agent poses (positions and orientations) and each factor in the factor graph imposes a constraint between two poses (e.g. relative estimate between a pair of poses)}. In our SLAM problem, we define two factors, i.e. an \textit{odometry} factor and a \textit{loop closure} factor, both of which are inferred from a DNN. Fig. \ref{fig:factor_graph} represents the factor graph representation of our problem. The odometry factor $u_i$ imposes constraints between consecutive positions $x_i$ and $x_{i+1}$, while the loop closure factor imposes constraints between two distant poses that have a large portion of image or feature correspondence (e.g. $x_{1}$ and $x_{4}$), but not necessarily obtained at the exact same location. These definitions refers those used by feature-based visual SLAM in a sense that as long as we have sufficient correspondence, although not exactly at the same location, we can estimate the relative pose between those two locations and use it as an additional constraint. These loop closure pairs are detected via observing similar measurements (e.g. $z_{1}$ and $z_{4}$) and also encoded with a further neural network. Then, given the odometry and the loop closure factors, our SLAM optimization objective is described as follows
\begin{align}
\begin{split}
\mathcal{X}^* = \argmin_x \sum_{i} \underbrace{ \norm{h_u(x_i, u_i) - x_{i+1}}^2_{\hat \Sigma_i} }_\text{odometry constraints} + \\ \sum_{<i,j>} \underbrace{ \norm{h_c(x_i, c_{ij}) - x_j }^2_{\hat \Lambda_{ij}} }_\text{loop closure constraints},
\label{eq:slam_objective}
\end{split}
\end{align}
where $h_u(x_i, u_i)$ represents the estimated agent position at $i+1$ after composing the previous position $x_i$ with odometry estimation $u_i$ from deep neural network. Note that we use the covariance matrix $\hat \Sigma_i = \varrho \Sigma_i$ to characterize the uncertainty of the odometry estimates where $\varrho$ is a scale factor. $h_c(x_i, c_{ij})$ models the estimated position of the corresponding loop pair $x_j$ after composing $x_i$ with the relative poses between $x_i$ and $x_j$ ($c_ij$) with $\hat \Lambda_{ij} = \rho \Lambda_{ij}$ as the covariance and $\rho$ as the scale factor. Note that we use $\varrho$ and $\rho$ to balance the contribution of the covariance in odometry and loop closure constraints during optimization.

\section{Overview of The Thermal-Inertial SLAM System}
\label{sec:slam_overview}
Fig. \ref{fig:ti_slam_architecture} depicts the high-level architecture of our thermal-inertial SLAM system. As can be seen, the SLAM front end consists of three neural network branches to generate odometry and loop closure constraints. Odometry constraints (6-DoF relative camera poses and its variances) are estimated by a \textit{neural odometry} network given consecutive normalized 14-bit thermal images and IMU sequences. To generate loop closure constraints, we first extract an embedding feature for each normalized 8-bit thermal image via a \textit{neural embedding} network. These embedding vectors summarize the salient features that best describes a thermal image. By comparing these embedding features against all other embeddings, we can detect an image pair with sufficient correspondence and identify it as a potential loop pair. Lastly, the relative poses between these loop pairs are estimated by a \textit{neural loop closure} network - these are then regarded as the loop closure constraints. Note that as we do not have IMU data to generate robust loop closure constraints in the SLAM back end, we perform outlier rejection to discard noisy loop closure pose estimations and only keep the inliers. Finally, given both odometry and (inlier) loop closure constraints, the back end optimizer will optimize the entire pose-graph using Eq. (\ref{eq:slam_objective}) to generate an optimized trajectory.

\begin{figure*}
    \centering
    \vspace{-0.5cm}
    \includegraphics[width=1.8\columnwidth]{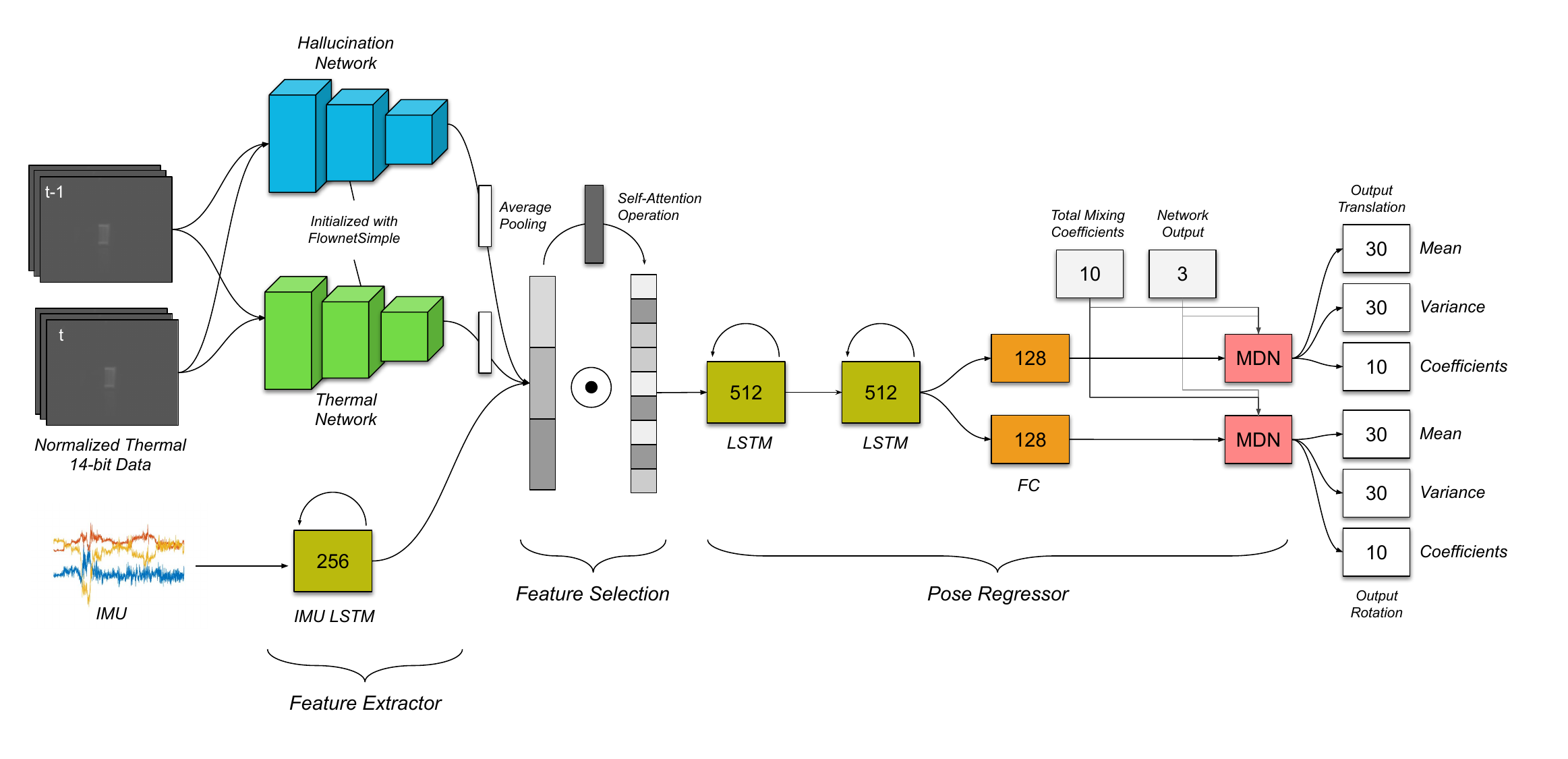}
    \vspace{-0.7cm}
    \caption{The network structure of neural thermal-inertial odometry which consists of a feature extractor, feature selection, and pose regressor. The network is trained to estimate the parameters of Mixture Density Network (MDN) given normalized radiometric thermal data (14-bit) as the input. To extract the 6-DoF poses during testing, we need to sample from the mixture models.}
\label{fig:neural_odom}
\end{figure*}

\section{SLAM Front End}
\label{sec:slam_frontend}
The SLAM front end is responsible for constructing the pose graph by abstracting the input from the thermal and inertial data using an MDN. In this section, we will detail the network structure and the training procedure for each neural network.

\subsection{Neural Thermal-Inertial Odometry}

\subsubsection{Network Architecture}
Fig. \ref{fig:neural_odom} depicts the architecture of our neural thermal-inertial odometry subsystem. This consists of three main parts, namely the feature extractor, the feature selector, and the pose regressor. The first part is the feature extractor, which is designed to distill geometrically meaningful features for odometry estimation. For images, this is typically implemented as optical flow estimation which captures the movement of pixels e.g. from edges of an object. However, since thermal images are inherently lacking in sufficient features to estimate dense optical flow (e.g. they are textureless), we follow the practice of \cite{saputra2020deeptio} which not only extract features from the thermal images, but also hallucinates visual features, simulating the ones extracted from a DNN-based visual odometry \cite{wang2017deepvo, saputra2019learning}. Given thermal images as the input, the hallucinated visual features will act as auxiliary information for the thermal features such that an accurate odometry can be inferred from textureless thermal sequences.

The second part is the feature selection module, which aims to select the most useful feature combination for odometry estimation. To this end, we follow the deterministic soft fusion structure in \cite{chen2019selective} to construct our feature selection. This feature selection is necessary because not all features are equally important at all times for accurate odometry estimation, e.g. each sensor comes with intrinsic noise. In particular, thermal data are plagued by fixed-pattern noise \cite{williams2019fixed}, while white random noise and sensor bias affect IMU data. Given these noisy features, the feature selection module will generate masks to re-weigh each feature by conditioning these over all input channels. The concatenated, re-weighted features, are then fed to the pose regressor network to estimate the parameters of the MDN which will be described in detail in the following section.

\subsubsection{Probabilistic Pose Regressor}
\revise{Instead of directly predicting 6-DoF agent poses as in the previous work \cite{wang2017deepvo, saputra2019learning}, the probabilistic pose regressor estimates the parameters of MDN: mean ($\mu$), variance ($\sigma$), and the mixing coefficients ($\alpha$) as seen in Eq. (\ref{eq:mdn_estimates}). To this end, we construct our pose regressor by stacking two LSTM layers followed by Fully Connected (FC) layers that decouple MDN parameters for translation and rotation. We employed LSTM since odometry is considered as a sequential motion estimation problem, in which implicitly modelling the temporal dependencies within the network is important. The two stacked LSTM layers will encapsulate the dependencies between the current and the previous frame in the latent states as described in \cite{wang2017deepvo, wang2018end}. For this purpose, we keep a one history of the previous hidden state in the LSTM although longer history is possible (yet it requires more computational time).}

\revise{To derive both the camera poses (6-DoF) and its uncertainty estimation (covariance matrix) through MDN, we model each component of the poses (e.g. translation in x direction) as a mixture of Gaussian. The number of mixing coefficients ($K$) stated in Eq. (\ref{eq:mdn_estimates}) are determined empirically and will be discussed in Section \ref{sec:prob_estimates}. The selection of K also determines the total paramaters of the MDN layer which are typically estimated via FC layers with $(3 * K)$ hidden units for the mean and the variances and $(K)$ hidden units for the mixing coefficients.} At test time, we can extract the 6-DoF poses together with the variances by sampling from the Gaussian mixture models.


\subsubsection{Learning Mechanism}
\label{sec:tio_learning_mechanism}
The neural odometry network is trained in two stages. In the first stage, we train the hallucination network and in the second stage, we train the remaining networks. As the visual hallucination network $\Psi_{H}$ is intended to imitate the visual features $\textbf{a}_{V}$ from real RGB images encoded by a visual encoder $\Psi_{V}$, we employ a deep Visual-Inertial Odometry (VIO) model as a pseudo ground truth to train $\Psi_{H}$. In particular, we use a VINet architecture \cite{clark2017vinet} to generate $\textbf{a}_{V}$ such that it can be used to train $\Psi_{H}$. Following \cite{saputra2020deeptio}, we employ the Huber loss \cite{huber1992robust} to train $\Psi_{H}$ to avoid the catastrophic impact of outliers due to periodic NUC operation in thermal camera. Then, by trying to minimize the discrepancy $\xi$ between the output activation from $\Psi_{H}$ and $\Psi_{V}$, our objective function $\mathcal{L}_{\text{H}}$ is defined as
\begin{align}
\begin{split}
    \mathcal{L}_{\text{H}} & = \frac{1}{n} \sum_{i=1}^{n} \mathcal{H}_{i} (\xi) \\
    \mathcal{H} (\xi) & = \begin{cases}
    \frac{1}{2}\norm{\xi}^{2} &  \text{for} \norm{\xi} \leq \delta, \\
    \delta ( \norm{\xi} - \frac{1}{2} \delta )              & \text{otherwise}
    \end{cases} \\
    \xi & = \Psi_{H}(\textbf{X}_{T};\textbf{W}_{H}) - \Psi_{V}(\textbf{X}_{V};\textbf{W}_{V}),
\label{eq:hallucination_loss}
\end{split}
\end{align}
where $\delta$ is a threshold, $n$ is the batch size during training, $\textbf{W}_{V}$ and $\textbf{W}_{H}$ are the weights for $\Psi_{V}$ and $\Psi_{H}$ respectively. Note that during this training process, we have previously trained VINet and freeze its weights.

In the second stage, we train the remaining part of the networks to estimate the parameter of MDN. As this is a supervised learning, we will provide the ground truth poses for the training. Then, the objective function is defined as follows
\begin{align}\
  \label{eq:mdn_pose_loss}
  \mathcal{L}_{MDN} = \sum_{i=1}^{n} \prob(\textbf{t}|Z)^-_i + \beta \sum_{i=1}^{n} \prob(\textbf{r}|Z)^-_i , 
\end{align}
where $\prob(\textbf{t}|Z)^-_i$ and $\prob(\textbf{r}|Z)^-_i$ are the negative log likelihood of Eq. (\ref{eq:mdn_estimates}) for each translation and rotation component respectively, with $\textbf{t} \in {\rm I\!R}^{3}$ and $\textbf{r} \in {\rm I\!R}^{3}$ are predicted translation and rotation. \revise{Note that we use Euler angle to represent rotation \textbf{r} as it is free from constraints and easier to converge as described by \cite{wang2017deepvo}. Note that the odometry motion is usually also constrained (e.g. the ground robot only perform rotation in yaw axis) which makes the usage of Euler angle safe from gimbal lock problem. We use $\beta$ to balance the loss between translation and rotation component as seen in \cite{wang2017deepvo, li2018undeepvo}.} Note that in this stage, we freeze the hallucination network $\Psi_{H}$ to avoid altering the learnt hallucination weights that have been trained in the first stage.


\subsection{Neural Embedding and Loop Closure Detection}

In the context of SLAM, the aim of loop closure detection is to identify whether the mobile agent has revisited a place. This information can then be used to constrain the odometry estimation and optimize the overall pose graph. 

Following the taxonomy described in \cite{garcia2015vision}, loop closure detection can be achieved through local or global image descriptors. Global descriptors represents an image in a holistic manner without the need to extract local features like SIFT \cite{lowe2004distinctive} or SURF \cite{bay2006surf}. Typical example includes representing the image as a colour intensity histogram as described in \cite{ulrich2000appearance} or other image statistics described in GIST \cite{oliva2001modeling}. On the other hand, local descriptor-based approaches extract keypoints (e.g. corners, blobs, or regions) and their corresponding descriptor vectors in which the measurements are typically taken from the vicinity of each keypoint. Then, aggregation methods such as those based on Bag-of-visual-words (BoW) \cite{angeli2008fast}, Vector of Locally Aggregated Descriptors (VLAD) \cite{jegou2010aggregating}, MAC \cite{radenovic2018fine}, or NetVLAD \cite{arandjelovic2016netvlad} can be used to summarize the descriptors.

Our neural embedding model follows the global descriptor approach as we do not rely on local features extracted from hand-engineered keypoints or aggregation methods, but directly generate a global descriptor from a thermal image through a neural network. We employ a global descriptor based approach due to the textureless nature of thermal images, in which two different features from an equivalent RGB image might be merged in a thermal image as they may have the same temperature. Thus, instead of focusing on clustering these ambiguous features (as been done by BoW or NetVLAD), we instead rely on global image information extracted by the deep network to improve the generalization.
 
\begin{figure}
    \centering
    \includegraphics[width=0.95\columnwidth]{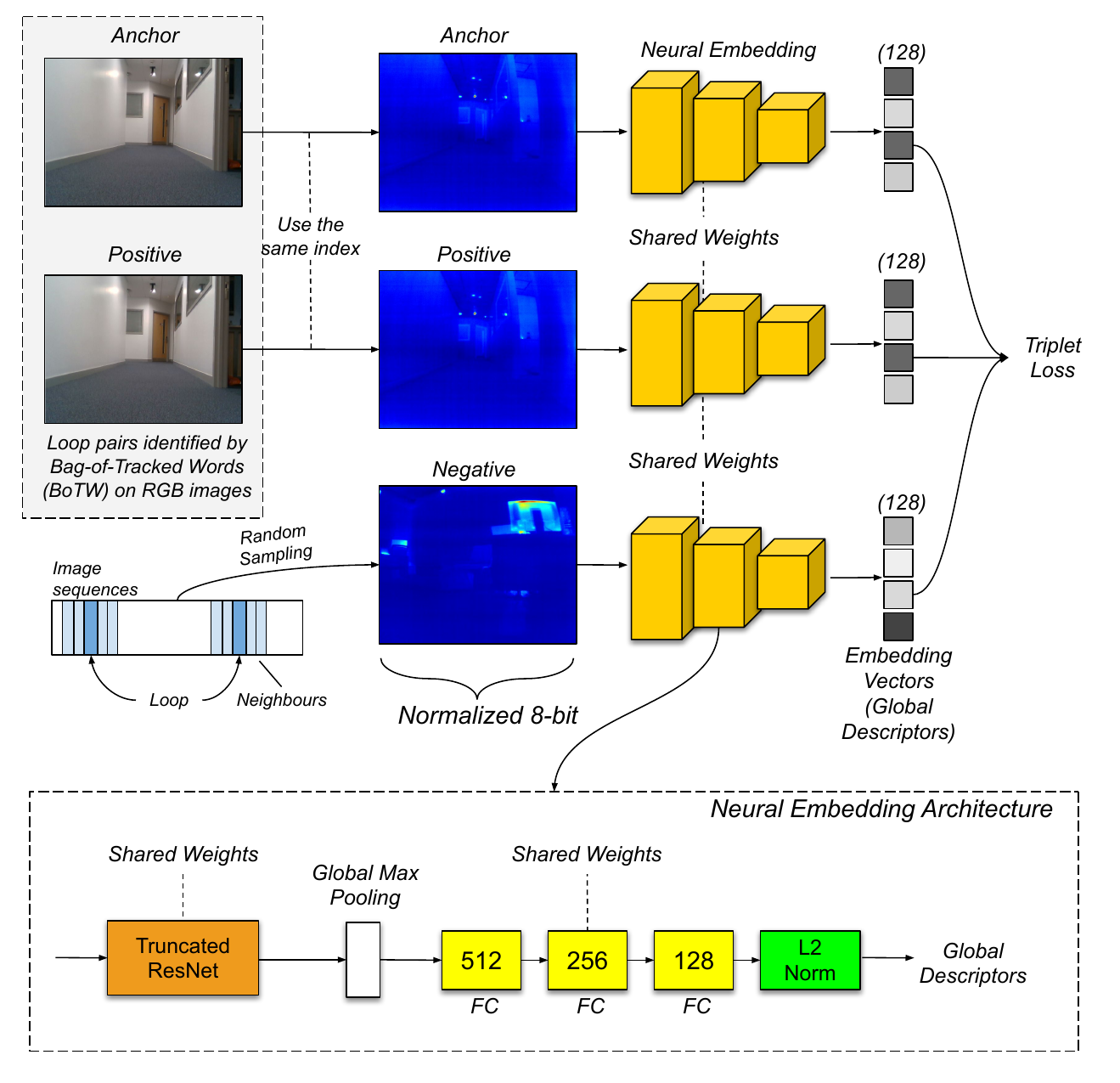}
    \vspace{-0.5cm}
    \caption{The network structure of neural embedding during training, which generates 128-D global descriptor for a single thermal image. For training, we obtain the list of anchor and positive examples from BoTW \cite{tsintotas2019probabilistic} applied on simultaneously captured RGB images such that our network can emulate BoTW performances on RGB.}
\label{fig:neural_embed}
\end{figure}

\subsubsection{Network Structure}
Fig. \ref{fig:neural_embed} depicts the structure of our neural embedding network. The network consists of Truncated ResNet, followed by global average pooling and fully connected layers. For the Truncated ResNet, we use ResNet50 structure \cite{he2016deep} up to the 49th layer to remove the classification part of the network and obtain a large spatial output dimension. The global max pooling layer is then used on top of ResNet50 to filter the most important part of the output vectors. We project this output embedding to a lower dimensional space by applying 3 FC layers. The number of hidden states in FC layers follows this decreasing rule $\text{FC}_1(\gamma * 4)$, $\text{FC}_2(\gamma * 2)$, $\text{FC}_3(\gamma)$, where $\gamma$ is the total output of the embedding vectors. In practice, we use $\gamma = 128$ to construct an efficient embedding vectors, following the practice in face recognition \cite{schroff2015facenet}. We presume that a small number of embedding vectors are sufficient to describe a thermal image since it has much less feature variation compared to the RGB images. Then, similar to the last layer in NetVLAD, we perform L2 normalization such that the entries of the embedding vectors will be sum to 1. To avoid training the entire network structure from scratch, we initialize the ResNet50 weights from ImageNet model and fine-tune all layers when training the network using our thermal data. 

\subsubsection{Learning Procedure}
To train the network, we follow the standard learning procedure to train a place recognition network based on triplet margin losses. A triplet $\{ \textbf{I}_T, \textbf{I}^+_T, \textbf{I}^-_T \}$ consists of an anchor image $\textbf{I}_T$, a positive example $\textbf{I}^+_T$, which represents a similar scene with sufficient correspondence with respect to the anchor (loop pair), and a negative example $\textbf{I}^-_T$, which represents an unrelated scene with no or minimum image correspondences with respect to the anchor. Given the triplet information, our triplet loss is defined as 
\begin{align}
\begin{split}
    \mathcal{L} (\textbf{I}_T, \textbf{I}^+_T, \textbf{I}^-_T) = \text{max} ( \lambda + \norm{d_{\textbf{W}_T}(\textbf{I}_T) - d_{\textbf{W}_T}(\textbf{I}^+_T)}^2 - \\ \norm{d_{\textbf{W}_T}(\textbf{I}_T) - d_{\textbf{W}_T}(\textbf{I}^-_T)}^2 , 0),
    \label{eq:triplet_loss}
\end{split}
\end{align}
where $d_{\textbf{W}_T}(.)$ is the neural embedding network, $d_{\textbf{W}_T}(\textbf{I}_T)$ is an embedding vector defining the global image descriptor of image $\textbf{I}_T$, $\textbf{W}_T$ is a shared trainable weights for the network, and $\lambda$ is a hyper-parameter to control the margin between positive and negative examples. By training the neural embedding network using Eq. (\ref{eq:triplet_loss}), the network is expected to produce similar embedding vectors when the mobile agent re-visits a place.

In order to provide the data for training the embedding network, we have to identify positive loop pairs amongst thermal image sequences. To avoid a manually laborious annotation task, we instead use the loop pair detected from a state-of-the-art place recognition algorithm applied on simultaneously captured RGB images as pseudo ground truth. In this sense, the embedding network can imitate how loop closures are formed from RGB correspondences, given thermal images as the input. Note that this is possible as both thermal and RGB cameras are placed in the same mobile agent with sufficient spatial correspondence. Then, to detect loop pairs amongst RGB images, we employ the state-of-the-art Bag-of-Tracked-Words (BoTW) \cite{tsintotas2019probabilistic}, an improved version of the standard Bag-of-Words (BoW) algorithm. The main difference between BoTW and Bow is that BoTW utilizes ``Tracked Words" among successive images rather than the standard histogram-based visual words in a single image, yielding more robust recognition performance. Nevertheless, despite its great performances on RGB images, BoTW performs very poor on thermal images (see the results in Section \ref{sec:experiments}). This is most likely because it relies heavily on point features, which are typically much scarcer on thermal images than in RGB images, especially when the images are captured in an uncluttered indoor environment. Hence, we use BoTW as our pseudo ground truth by applying it on RGB images rather than directly utilizing it to detect loops on thermal images. To improve the number of data during training, we interchange the anchor and the positive loop pair in a triplet and choose a random negative example that does not belong to the anchor and to the positive examples, nor is within adjacent frames with respect to the anchor and the positive examples (see illustration in Fig. \ref{fig:neural_embed}). This was done with expectation that the network should produce a similar embedding vector for adjacent frames, although these will not be considered as a loop pair.


\subsubsection{Loop Closure Detection}
\revise{After correctly training the embedding network, we can generate the embedding vectors for each thermal image. To detect a loop pair, we compute the discrepancy between each pair of embedding vectors $i$ and $j$ using cosine distance as follows} 
\begin{align}
  \label{eq:cosine_distance}
  \mathcal{S}_{ij} & = \frac{d_{\textbf{W}_T}(\textbf{I}_T^i) . d_{\textbf{W}_T}(\textbf{I}_T^j)}{\norm{d_{\textbf{W}_T}(\textbf{I}_T^i)} \norm{d_{\textbf{W}_T}(\textbf{I}_T^j)}} ,
\end{align}
where $\norm{.}$ is the magnitude of the embedding vectors. If $\mathcal{S}_{ij} < \zeta$, where $\zeta$ is a threshold, then the embedding pair is regarded as a loop closure pair. The selection of $\zeta$ is important as it determines the trade-off between the number of true positive and false positive pairs. \revise{In practice, we set $\zeta=0.045$ to generate around $85\%$ true positive loop pair although some false positive pair will also be detected (as depicted in ROC curve in Fig. \ref{fig:roc_curve}). Nevertheless, in offline operation, $\zeta$ can be tuned individually for different sequences as necessary.}

\begin{figure}
    \centering
    \includegraphics[width=0.95\columnwidth]{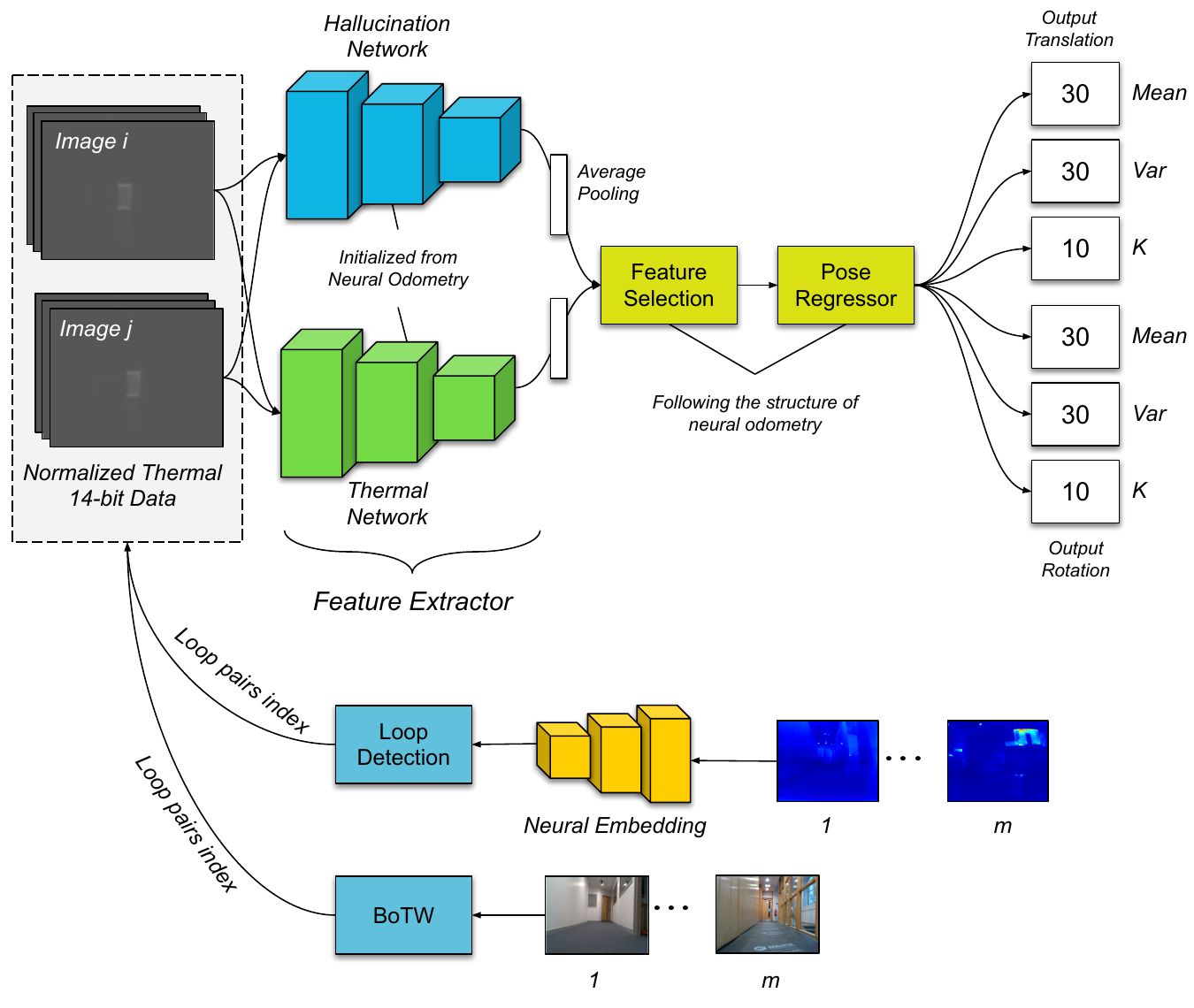}
    \vspace{-0.5cm}
    \caption{\revise{The network structure of \textit{neural loop closure} during training. This network is used to estimate the relative poses (and uncertainty) between a loop pair by using MDN. The structure resembles neural thermal-inertial odometry except that it does not have an IMU and the input from thermal images comes from frames $i$ and $j$ instead of successive frames. For training, the input loop pairs are obtained from both BoTW applied on RGB images and the neural embedding network applied on thermal images.}}
\label{fig:neural_loop}
\end{figure} 

\subsection{Neural Loop Closure}
Given a loop pair, we need to extract the relative poses between them such that we can inject it into the SLAM back end as loop closure constraints. \revise{To this end, we construct another deep network, termed Neural Loop Closure, that can estimate relative poses (and uncertainty) using MDN only from a pair of thermal images. Fig. \ref{fig:neural_loop} depicts the architecture of the network which resembles the neural thermal-inertial odometry network.} The main difference is that the input thermal pair does not come from consecutive images, but instead from a loop closure pair $i$ and $j$. Note that we do not have IMU data between them, making the training process even more challenging. 

For training, we provide the network with the list of loop pairs together with the ground truth poses. We utilize the list of loop pair generated by both BoTW applied on RGB images and Neural Embedding network applied on thermal images to provide more data during training. We initialize both thermal and hallucination network with the weights from neural thermal-inertial odometry to ease the optimization process. For testing, we sample from the mixture models to obtain 6-DoF relative poses between loop pairs. Given both odometry and loop closure constraints, now we can finally construct a complete pose graph to be optimized by the SLAM back end. 

\section{SLAM Back End}
\label{sec:slam_backend}
The SLAM back end is responsible for optimizing the whole trajectory given the pose graph constructed by the front end. However, the pose graph built by the front end is often subject to noise. For example, the odometry prior can be inaccurate and largely drift, or the loop closure constraints are erroneous as the poses are generated only from a pair of thermal images without the assistance of IMU. To mitigate such noise, we incorporate an outlier rejection module to filter and feed only the reliable ones to the subsequent graph-based optimization.  

\begin{figure*}
    \vspace{-0.3cm}
        \includegraphics[width=0.195\columnwidth]{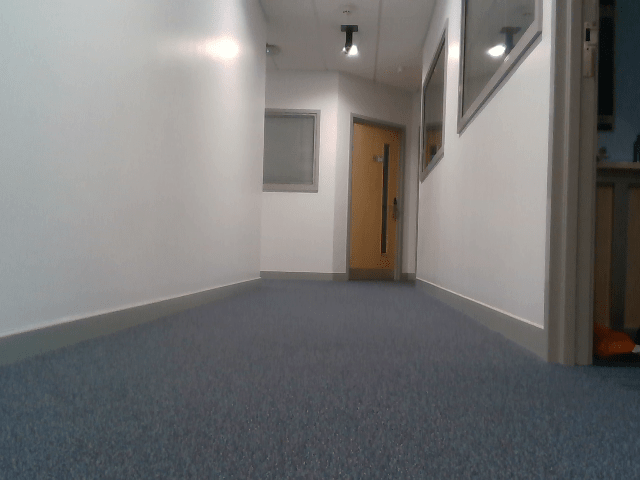}
        \includegraphics[width=0.195\columnwidth]{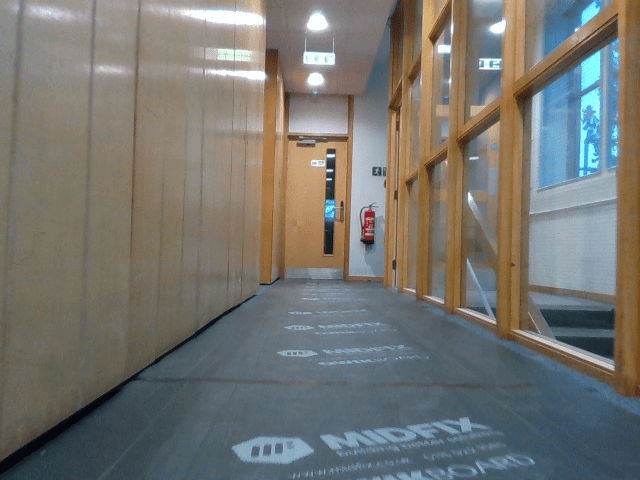}
        \includegraphics[width=0.195\columnwidth]{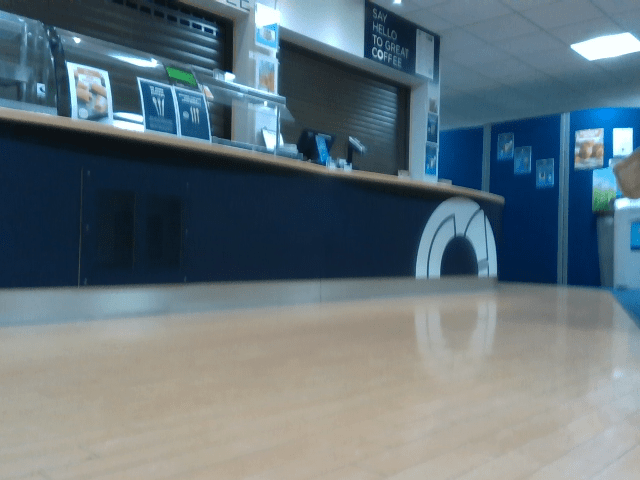}
        \includegraphics[width=0.195\columnwidth]{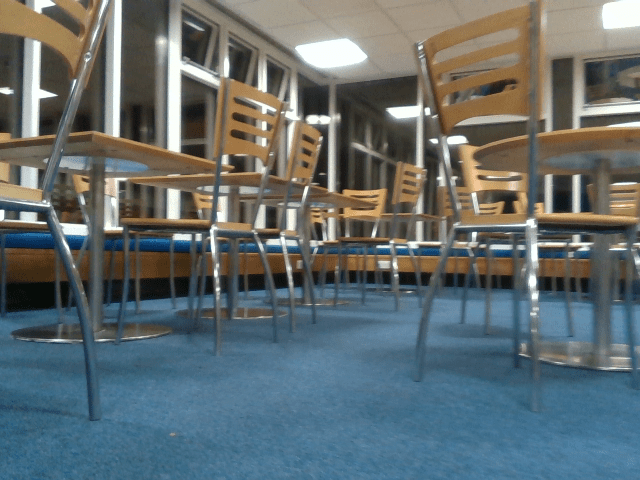}
        \includegraphics[width=0.195\columnwidth]{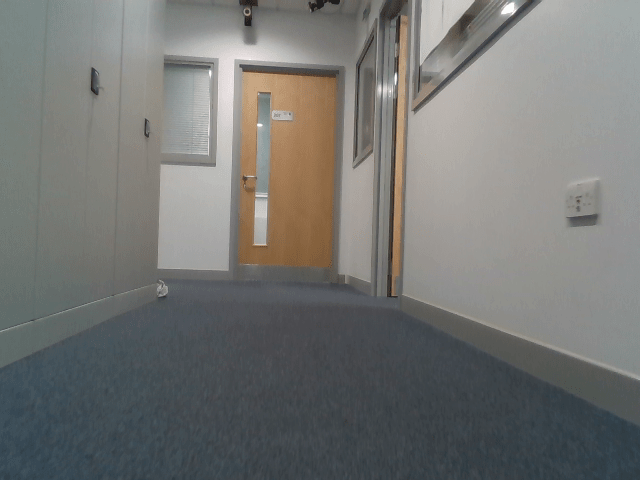}
        \includegraphics[width=0.195\columnwidth]{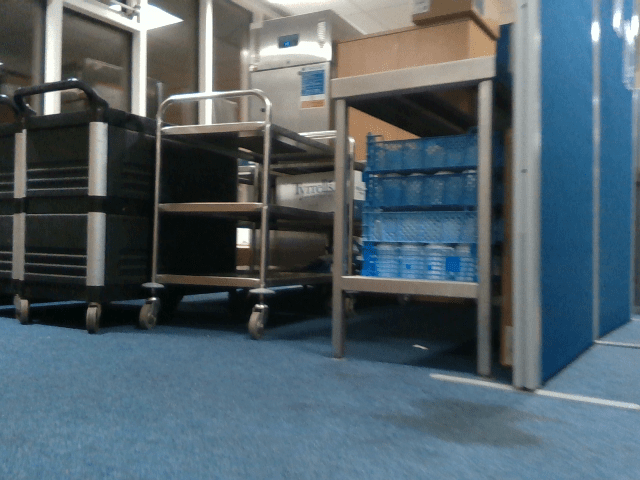}
        \includegraphics[width=0.195\columnwidth]{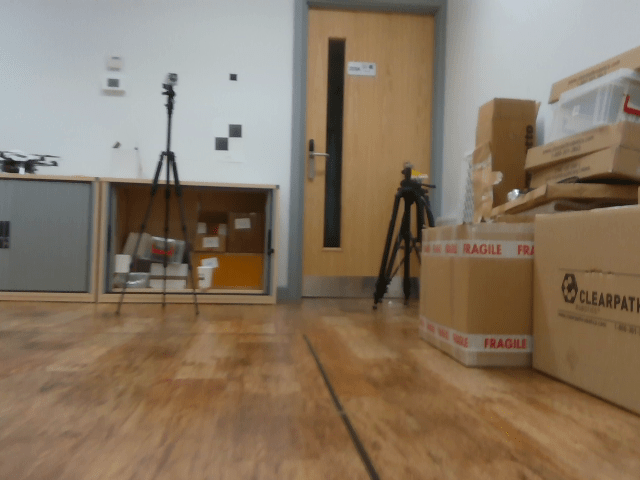}
        \includegraphics[width=0.195\columnwidth]{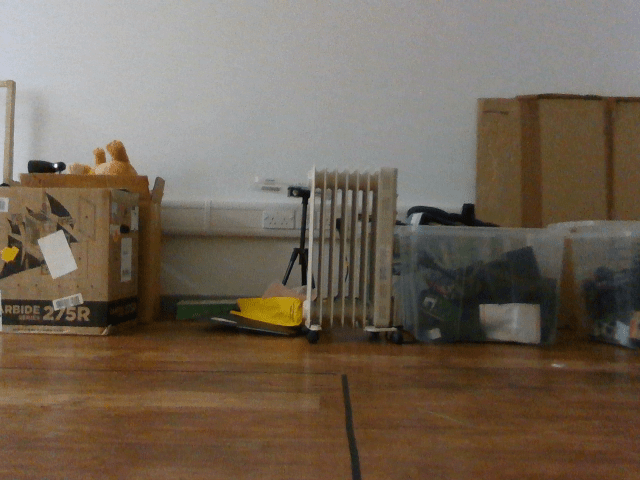}
        \includegraphics[width=0.195\columnwidth]{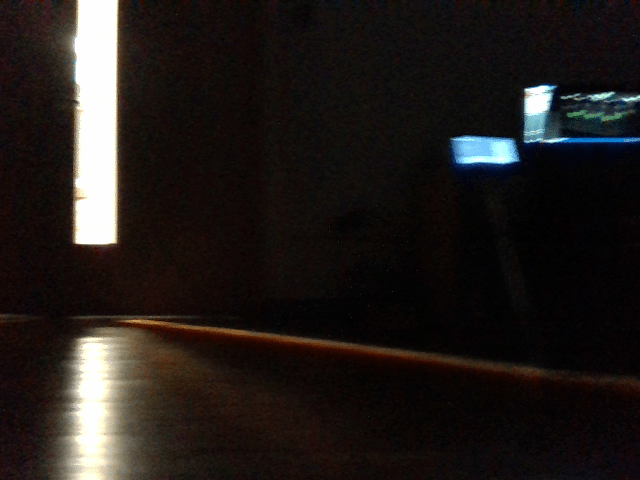}
        \includegraphics[width=0.195\columnwidth]{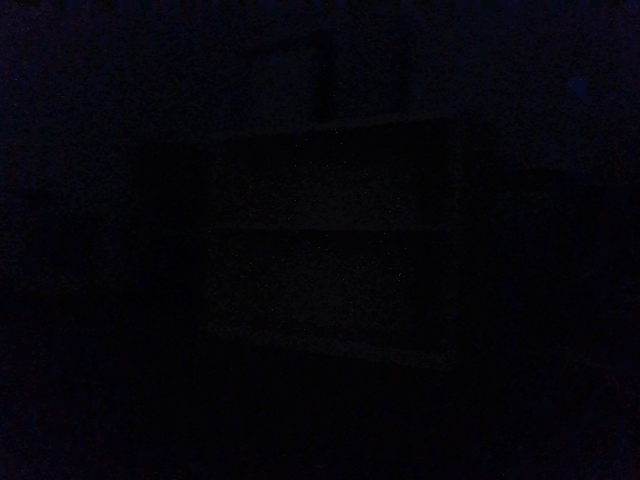} \\
        \includegraphics[width=0.195\columnwidth]{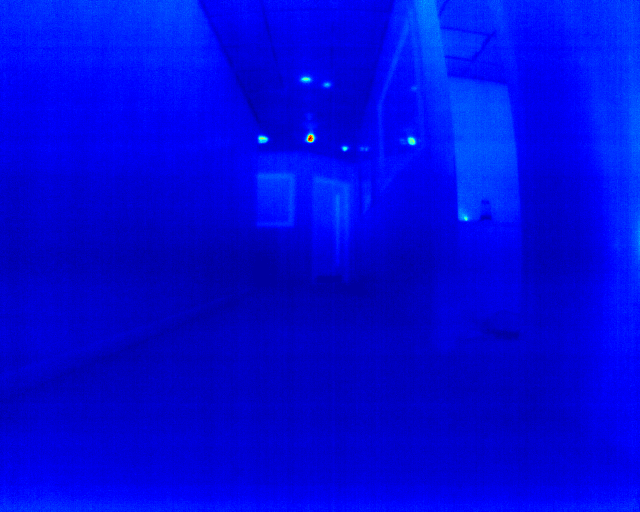}
        \includegraphics[width=0.195\columnwidth]{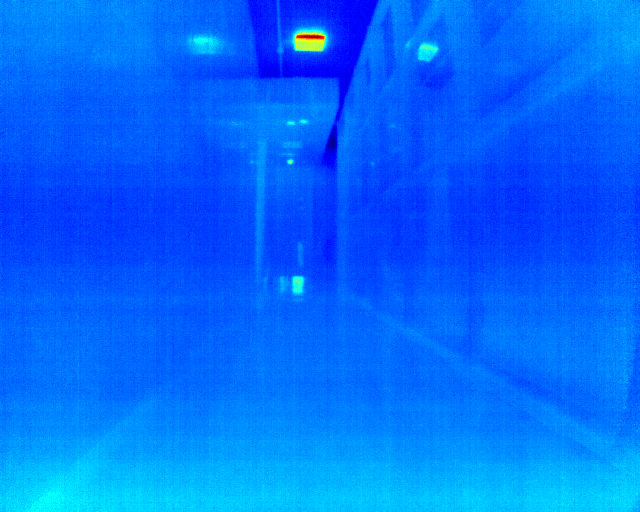}
        \includegraphics[width=0.195\columnwidth]{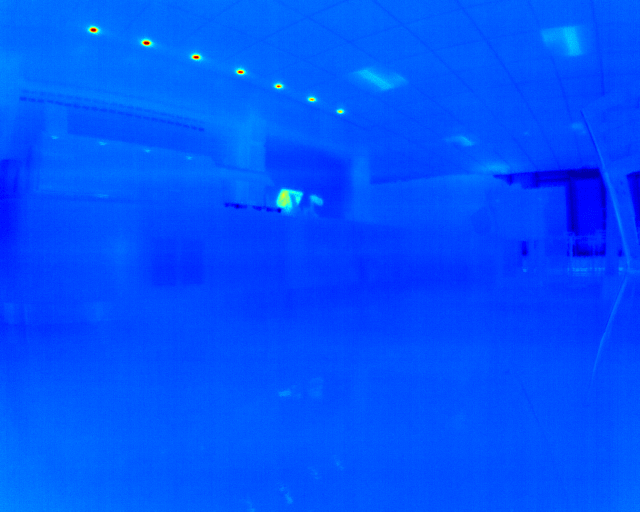}
        \includegraphics[width=0.195\columnwidth]{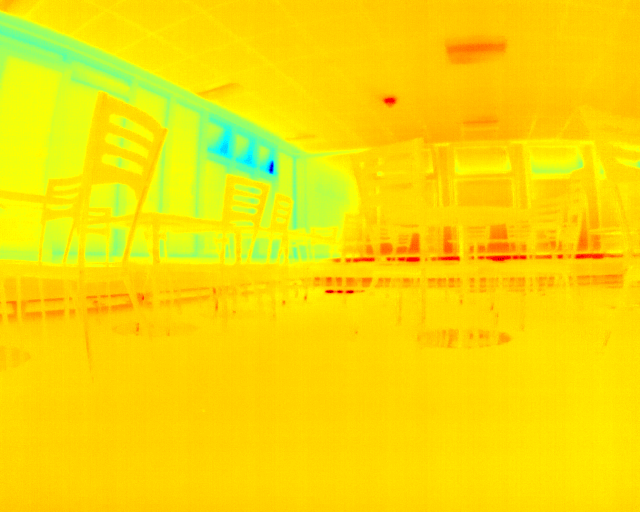}
        \includegraphics[width=0.195\columnwidth]{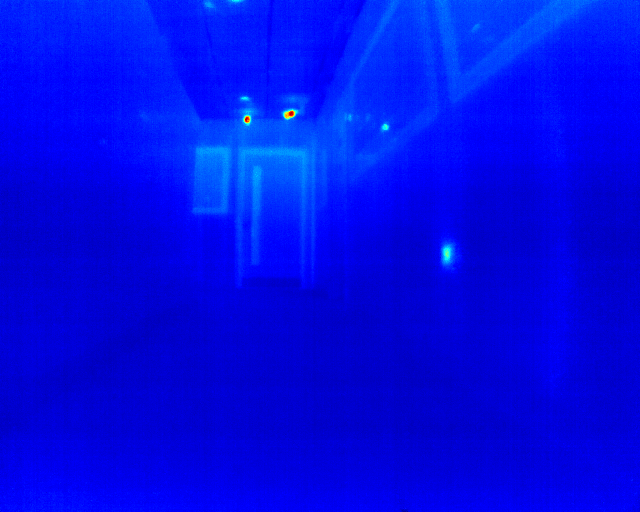}
        \includegraphics[width=0.195\columnwidth]{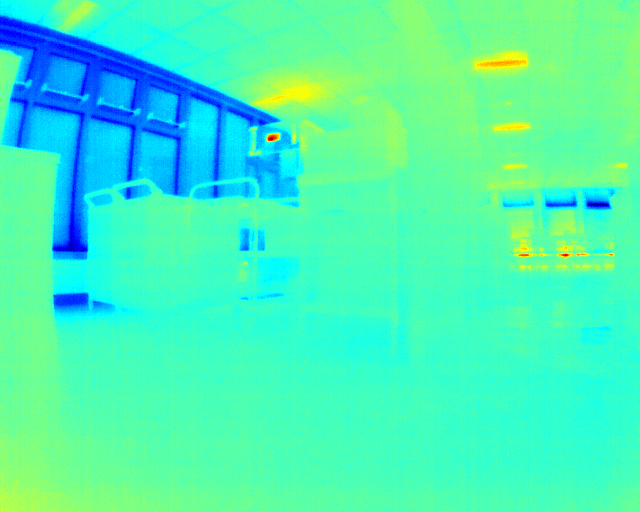}
        \includegraphics[width=0.195\columnwidth]{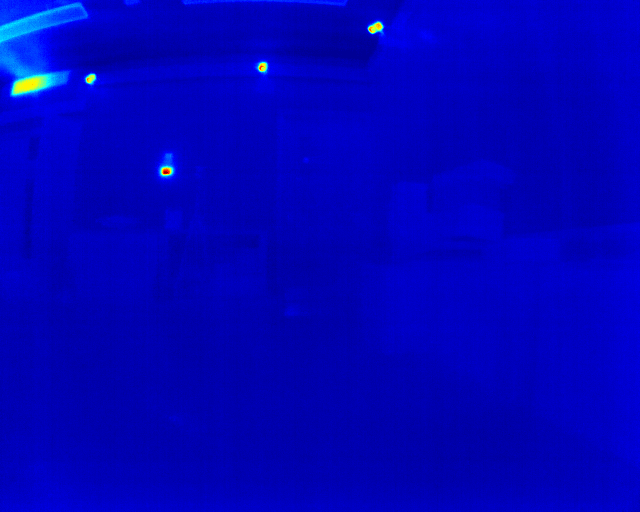}
        \includegraphics[width=0.195\columnwidth]{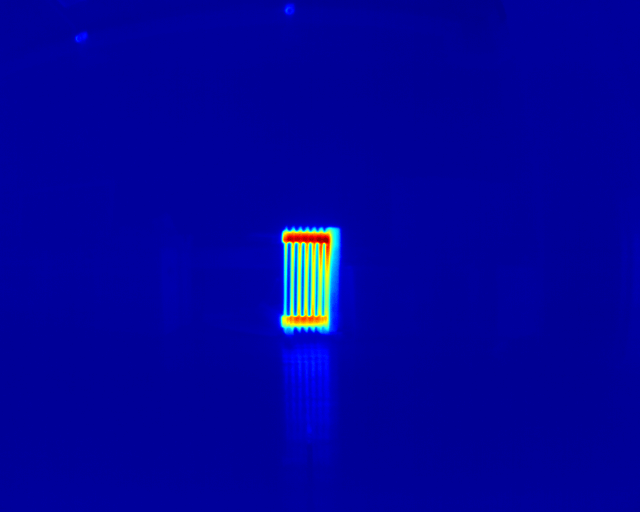}
        \includegraphics[width=0.195\columnwidth]{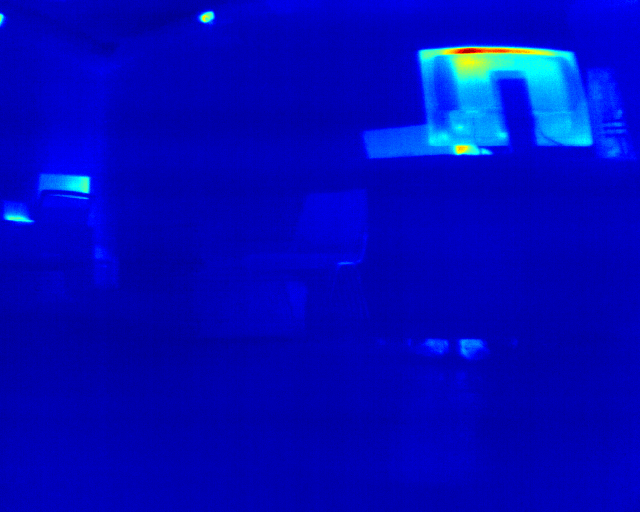}
        \includegraphics[width=0.195\columnwidth]{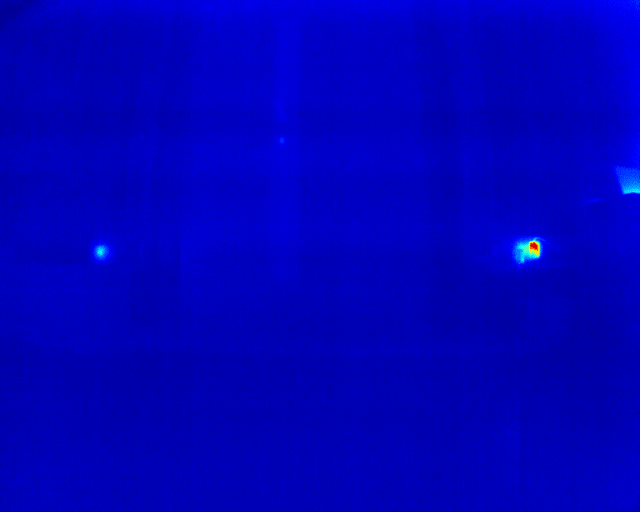}
    \vspace{-0.7cm}
    \caption{Sample images from the in-house dataset. The top and bottom images are from RGB and thermal camera respectively. For clarity, we display the rescaled (normalized) radiometric data instead of the 14-bit raw radiometric data.}
\label{fig:setup_and_sample_data}
\end{figure*}

\subsubsection{Outlier Rejection}
\revise{The outlier rejection module extends GraphTinker \cite{xie2017graphtinker} to our 6-DoF context, which is essentially based on the geometric consistency of loop closure constraints.} Specifically, consider two loop closure proposals detected by the front end, if these two loop closure proposals are both true, then within any reference frame, the two trajectory segments defined by them will form a \emph{sub-loop}. If the two segments are geometrically consistent (e.g., validated through the reverse odometry method as described in \cite{xie2017graphtinker}), one can confirm that both loop closure proposals are true, and mark them as valid (ready to be used by the later graph optimization). Otherwise, at least one of the two loop closure proposals is false positive. In that case we will not push any of them to later optimization, but temporarily keep them in the proposal set for next validation. After all sampled loop closure proposals are examined, the closures with pass rate less than a threshold will be rejected as outliers. In practice, we typically set the threshold less than 1. However, the best setting are varied for different environments as it also highly depend on the accuracy of the individual loop closure constraints.

\subsubsection{Optimization}
\revise{Given the pose graph with odometry constraints and selected (filtered) loop closure constraints, the back end optimizer will minimize Eq. (\ref{eq:slam_objective}) using Levenberg-Marquardt algorithm (i.e., g\textsuperscript{2}o \cite{kummerle2011g}).} To generate optimal solution, the balancing coefficients between odometry and loop closure constraints have to be selected with care. From our experiments, the weight for loop closure constraints is usually set higher than the weight for odometry constraints. For online operation, we typically set $\varrho=0.01$ and $\rho=3$. However, the optimized solution might be obtained by setting different $\varrho$ and $\rho$ for different sequences and environment. This might be the right choice if online operation is not required, e.g. the pose graph optimization can be done in offline fashion as in the Structure from Motion (SfM) works.

\begin{figure}
\centering
    \vspace{-0.1cm}
        \includegraphics[width=0.3\columnwidth]{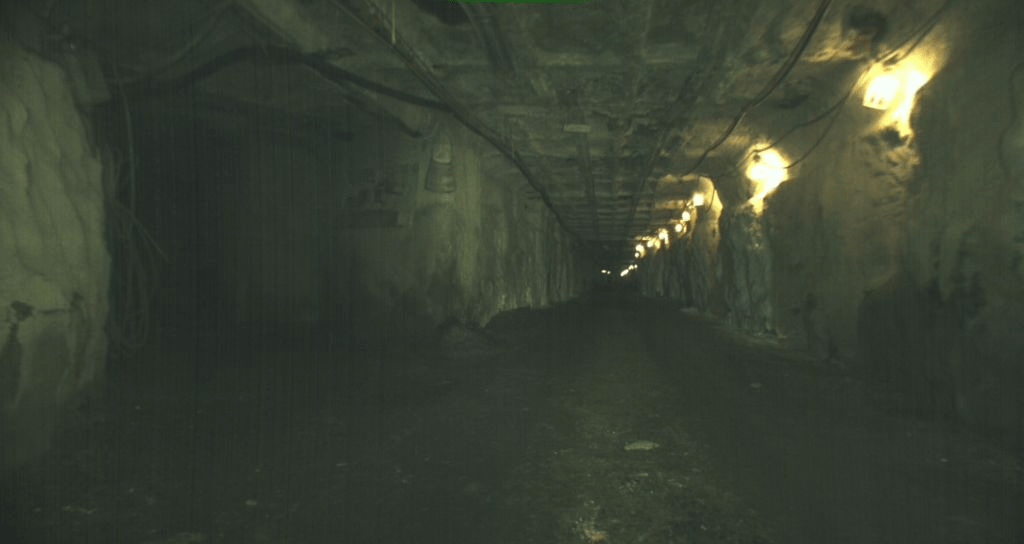}
        \includegraphics[width=0.3\columnwidth]{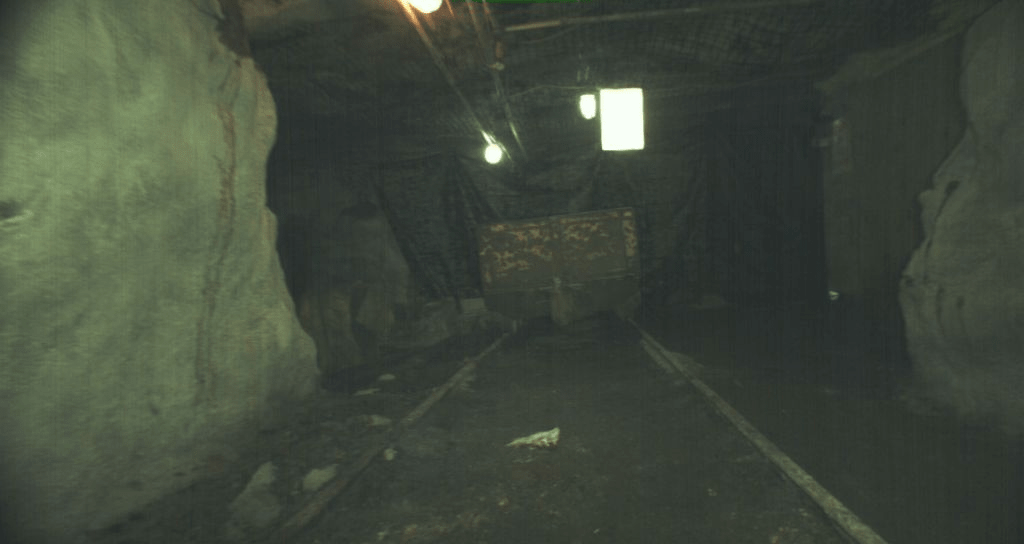}
        \includegraphics[width=0.3\columnwidth]{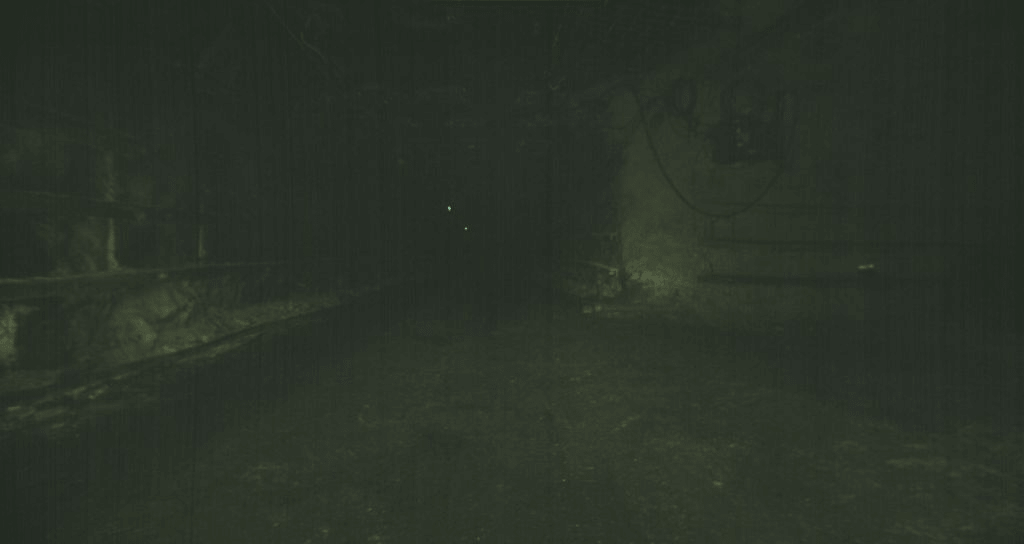}
    \vspace{-0.1cm}
    \caption{\revise{Sample images from Subt-tunnel dataset \cite{rogers2020test}.}}
\label{fig:subt_images}
\end{figure}

\section{Experiments}
\label{sec:experiments}
\subsection{Dataset}
\revise{We use multiple datasets to test our model, including self-collected and public datasets. For the self-collected dataset, we conducted experiments on ground robot and handheld data in indoor environment with around 8 km total trajectory length. For the public dataset, we used SubT-tunnel dataset \cite{rogers2020test}. Each dataset is described as follows.}

\subsubsection{Indoor Ground Robot Data}
We collected our data with Turtlebot 2. We use a Flir Boson 640 thermal camera to capture raw radiometric (14-bit) thermal data at 30~fps with $640\times512$ resolution. For the IMU, we utilize XSens MTI-1 Series running at 100 Hz. We also equip the robot with a Velodyne HDL-32E Lidar running at 10 Hz and an Intel Real Sense Depth camera with $680 \times 480$ RGB resolution (rolling shutter) running at 30 fps. These RGB data are used for training the hallucination network and assisting loop closure detection during training. Note that there are at least 2/3 spatial correspondences between both RGB and thermal images, enabling the training of hallucination network. In total, we have 36 sequences collected in different type of environments (e.g. hall, canteen, office, corridor, etc.), in which we use 23 sequences for training and 13 sequences for testing. For odometry training, we employ inertial-assisted wheel odometry provided by the Turtlebot 2 as the pseudo ground truth. For evaluation, following the practice in \cite{saputra2020deeptio}, we utilize VICON Motion Capture system and Lidar Gmapping\footnote{https://openslam-org.github.io/gmapping.html} to generate the ground truth poses. Lidar is particularly used when we do not have VICON as we collected some of our data in public space. To examine system robustness against different visibility conditions, we collect the data with sufficient illumination (bright), dim, or in darkness. Fig. \ref{fig:setup_and_sample_data} shows the example images for both RGB and thermal cameras.

\subsubsection{Indoor Handheld Data}
\revise{For the handheld data, we built a 3D printed model that utilizes the same set of sensors as with the indoor ground robot data. The only difference is, instead of using Velodyne HDL-32E Lidar, we replaced it with a more lightweight Velodyne Ultra Puck. We collected data in ten distinct floors from $6$ different multistorey buildings including hallways, canteen, common room, building junction, atrium, office, and cluttered store rooms. The smallest floor has an area around $205m^{2}$ while the largest one reaches $1500m^{2}$. For quantitative evaluation, we used lidar SLAM generated by ALOAM \footnote{https://github.com/HKUST-Aerial-Robotics/A-LOAM} as the (pseudo) ground truth. To test the system robustness in adverse lighting condition, we also collected data in a smoke-filled environment firefighter training facility, which is located at Washington DC Fire and EMS Training Academy, United States.}
\subsubsection{SubT-tunnel Data}
\revise{SubT-tunnel dataset \cite{rogers2020test} is a public dataset collected by a participant of DARPA subterranean challenge from CCDC Army Research Laboratory (ARL). The dataset contains synchronized lidar, RGB (stereo), depth, thermal, and IMU, taken from a ground robot moving in a long trajectory in an underground tunnel. The dataset is divided into 2 categories, namely urban circuit and tunnel circuit dataset. Sample images can be seen in Fig. \ref{fig:subt_images}. For this experiment, we only used the tunnel circuit as this data contains usable 14-bit thermal data (10 Hz) captured from Flir Boson camera. In particular, we utilized \texttt{ex\_B\_route1} sequence (54 minutes) for testing while the remaining are used for fine-tuning \revisetwo{(re-training with a lower learning rate)}. The (pseudo) ground truth for this experiment was provided by Lidar OmniMapper \cite{trevor2014omnimapper}.}

\begin{table*}[t]
\centering
\begin{threeparttable}
  \setlength{\tabcolsep}{3pt} 
  \caption{\revisetwo{RMS Relative Pose Errors (RPE) in Indoor Ground Robot Data}}
  \renewcommand{\arraystretch}{1}
  \fontsize{9}{10}\selectfont
  \label{table:evaluation_robot_rpe}
  \begin{tabular}{|c|cc|cccccccc|cccccc|}
    \hline
    \multirow{3}*{Seq} & \multirow{3}*{Lighting} & \multirow{2}*{Length} & \multicolumn{2}{c}{VINet \cite{clark2017vinet}} & \multicolumn{2}{c}{\revise{TI odometry}} & \multicolumn{2}{c}{\revise{TI odometry}} & \multicolumn{2}{c|}{TI odometry} & \multicolumn{2}{c}{VINet \cite{clark2017vinet}} & \multicolumn{2}{c}{{VINS-Mono}} & \multicolumn{2}{c|}{\multirow{2}{*}{Inertial+Wheel}} \\
    
    & & & \multicolumn{2}{c}{(Thermal 14-bit)} & \multicolumn{2}{c}{\revise{8-bit (\textbf{ours})}} & \multicolumn{2}{c}{\revise{w/o hallu. (\textbf{ours})}} & \multicolumn{2}{c|}{14-bit (\textbf{ours})} & \multicolumn{2}{c}{(RGB)} & \multicolumn{2}{c}{\revise{(RGB)}} & \multicolumn{2}{c|}{} \\
    
    & & (m) & $\textbf{t}$ (m) & $\textbf{r}$ ($^{\circ}$) & $\textbf{t}$ (m) & $\textbf{r}$ ($^{\circ}$) &  $\textbf{t}$ (m) & $\textbf{r}$ ($^{\circ}$) & $\textbf{t}$ (m) & $\textbf{r}$ ($^{\circ}$) & $\textbf{t}$ (m) & $\textbf{r}$ ($^{\circ}$) & $\textbf{t}$ (m) & $\textbf{r}$ ($^{\circ}$) & $\textbf{t}$ (m) & $\textbf{r}$ ($^{\circ}$)\\
    \hline
    \hline
    32 & Bright & 31.4 & 0.049 & 2.543 & 0.047 & 2.496 & 0.048 & 2.514 & \textbf{0.039} & \textbf{2.431} & 2.613 & 0.044 &  - & - & 0.029 & 2.426 \\
    33 & Dim & 22.5 & 0.027 & 1.440 & 0.038 & 1.366 & \textbf{0.019} & 1.218 & \textbf{0.019} & \textbf{1.208} & 1.341 & 0.029 & - & - & 0.017 & 1.294 \\
    34 & Dim & 20.7 & 0.023 & 1.406 & 0.044 & 1.358 & 0.022 & 1.344 & \textbf{0.021} & \textbf{1.263} & 1.653 & 0.028 & - & - & 0.017 & 1.531 \\
    37 & Dark & 71.1 & 0.031 & 1.374 & 0.042 & 1.284 & 0.023 & 1.306 & \textbf{0.021} & \textbf{1.233} & 1.809 & 0.033 & - & - & 0.017 & 1.386 \\
    39 & Bright & 16.8 & 0.029 & 1.696 & 0.044 & 1.586 & 0.029 & 1.645 & \textbf{0.028} & \textbf{1.579} & 1.679 & 0.034 & - & - & 0.022 & 1.550 \\
    42 & Dim & 65.1 & 0.029 & 1.714 & 0.049 & 1.755 & \textbf{0.028} & 1.618 & \textbf{ 0.028} & \textbf{1.570} & 1.773 & 0.035 & 0.144 & 0.816 & 0.026 & 1.676 \\
    43 & Dim & 66.5 & 0.029 & 1.713 & 0.039 & 1.727 & 0.027 & 1.614 & \textbf{0.025} & \textbf{1.571} & 1.670 & 0.033 & 0.166 & 0.799 & 0.023 & 1.651 \\
    44 & Dim & 71.1 & 0.035 & 1.727 & 0.048 & 1.685 & 0.031 & 1.763 & \textbf{0.029} & \textbf{1.676} & 1.839 & 0.038 & 0.190 & 0.859 & 0.026 & 1.758  \\
    45 & Bright & 61.3 & 0.034 & 1.674 & 0.046 & 1.633 & 0.030 &	1.655 & \textbf{0.027} & \textbf{1.557} & 1.870 & 0.037 & 0.173 & 0.848 & 0.024 & 1.641 \\
    46 & Bright & 21.7 & 0.032 & 1.517 & 0.052 & 1.516 & \textbf{0.028} & 1.430 & \textbf{0.028} & \textbf{1.353} & 1.518 & 0.038 & 0.150 & 0.695 & 0.038 & 1.452 \\
    47 & Bright & 22.2 & 0.033 & 1.779 & 0.045 & 1.753 & 0.028 & 1.644 & \textbf{0.027} & \textbf{1.577} & 1.756 & 0.035 & 0.173 & 0.859 & 0.018 & 1.808 \\
    48 & Bright & 42.1 & 0.032 & 1.557 & 0.047 & 1.605 & \textbf{0.025} & 1.525 & \textbf{0.025} & \textbf{1.462} & 1.655 & 0.036 & 0.205 & 0.787 & 0.022 & 1.623 \\
    49 & Bright & 81.1 & 0.025 & 0.766 & 0.043 & 0.705 & \textbf{0.019} & 0.695 & 0.022 & \textbf{0.679} & 0.737 & 0.032 & 0.166 & 0.474 & 0.025 & 0.745 \\
    \hline
    \hline
    \multicolumn{3}{|c|}{Mean} & 0.034 & 1.608 & 0.044 & 1.575 & 0.028 & 1.536 & \textbf{0.026 } & \textbf{1.473} & 0.031 & 1.686 & 0.171 & 0.767 & 0.023 & 1.580 \\
    \hline
    \multicolumn{17}{p{400pt}}{\footnotesize \revisetwo{*The bold indicates the most accurate method among algorithms with thermal and inertial data as the input.}}
  \end{tabular}
\end{threeparttable}
\end{table*}

\subsection{Odometry Evaluation in Ground Robot Data}
Odometry constraints is an important factor to yield accurate thermal-inertial SLAM because it is utilized as the initial estimation in the pose graph optimization. \revise{In this section, we study the influence of thermal representation, measure the accuracy and the timing of the odometry factor, and validate the estimated variances. To measure the quality of the odometry estimation, we measure the Root Mean Square (RMS) of Relative Pose Errors (RPE) and Absolute Trajectory Errors (ATE) against ground truth provided by VICON and Lidar Gmapping.}

\begin{figure*}[!ht]
    \centering
    \vspace{-0.4cm}
    \includegraphics[width=15cm,trim=0.7cm 1.9cm 1.1cm 2cm,clip]{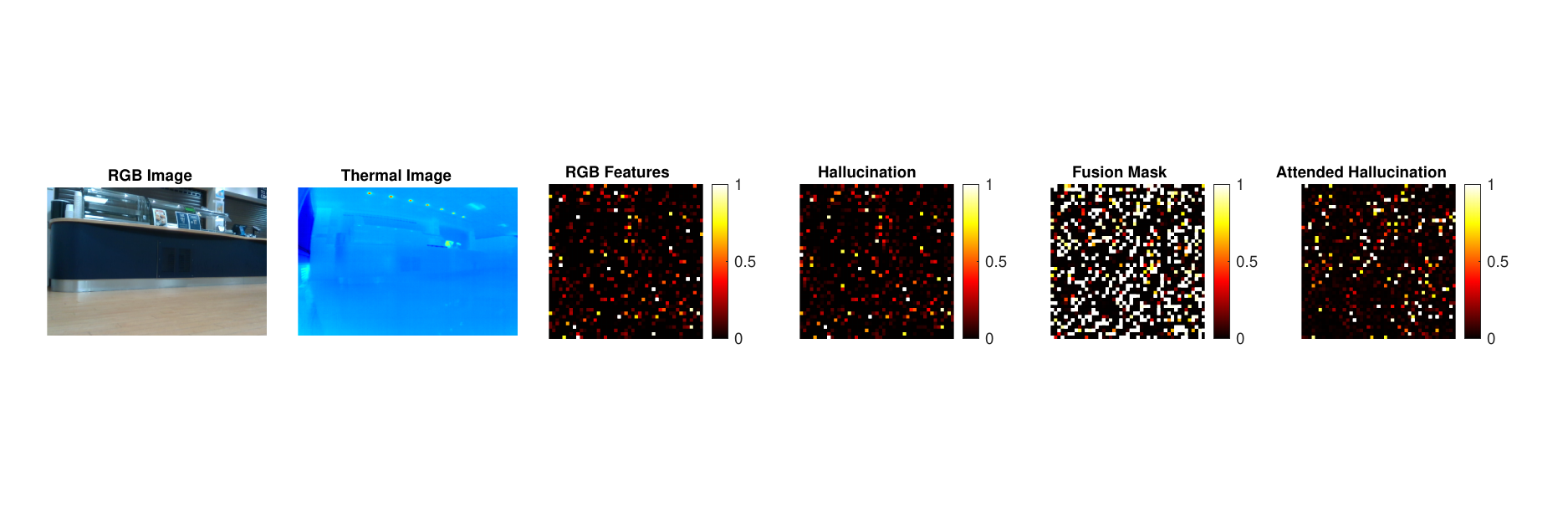} \\
    \vspace{-0.8cm}
    \includegraphics[width=15cm,trim=0.7cm 2cm 1.1cm 2cm,clip]{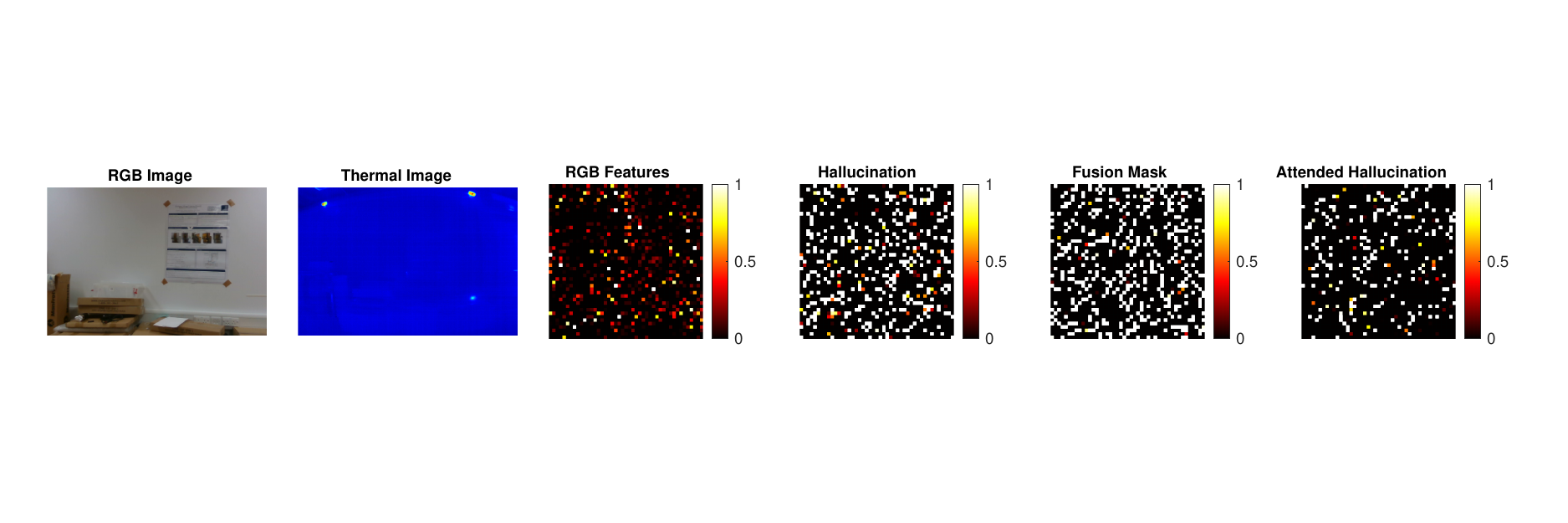}
    \vspace{-0.9cm}
    \caption{\revise{Comparison between RGB features produced by VINet and fake RGB features produced by the hallucination network. \textbf{Top}: example of accurate hallucination, showing generalization in the test data. \textbf{Bottom}: example of erroneous hallucination due to limited thermal features. In this scenario, selective fusion produces less dense fusion mask, relying more on other modality like IMU. From left to right: RGB image, thermal image, original RGB features, hallucinated RGB features, fusion mask for the hallucination features, and the output attended hallucination features.}}
\label{fig:features_visualization}
\end{figure*}

\subsubsection{\revise{The Influence of Thermal Representation}}
\revise{In this section, we investigate the choice of the thermal input by comparing the normalized 8-bit and 14-bit representation. As one can see in Table \ref{table:evaluation_robot_rpe}, the 14-bit TI odometry produces more accurate results than the 8-bit TI odometry for all sequences. Hypothetically this is because our odometry network is devised based on the optical flow network (i.e. FlowNet \cite{dosovitskiy2015flownet}) which extracts pixel displacement features rather than image appearance features. By using the 14-bit representation, it is easier to retain similar gradient information between consecutive frames such that consistent pixel displacement distribution in a short period of time can be extracted, even when frames sub-sampling is performed. On the other hand, the re-scaling process in the 8-bit representation might (slightly) alter the weak gradient information, making it more difficult to extract consistent pixel displacement distribution for the network to learn. Nevertheless, the accuracy differences between 8-bit and 14-bit are marginal, indicating that the 8-bit representation is also usable for deep odometry estimation.}

\subsubsection{\revise{The Importance of Hallucination and Selective Fusion}}
\begin{figure}
    \vspace{-0.5cm}
    \subfloat[Seq 39]{
        \includegraphics[width=4.3cm,trim=0.8cm .5cm 1cm .5cm,clip]{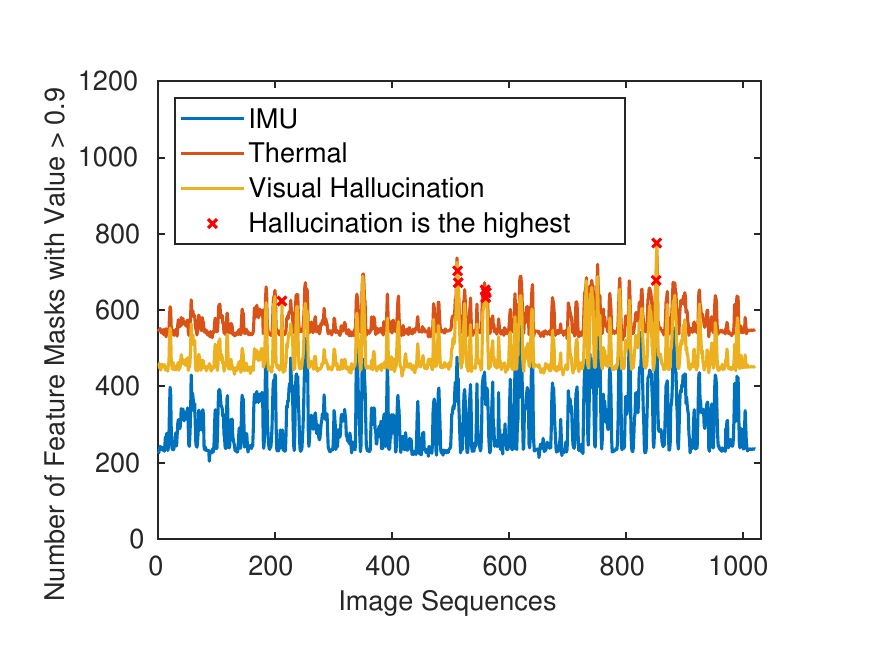}}
    \subfloat[Poster data]{
        \includegraphics[width=4.3cm,trim=0.8cm .5cm 1cm .5cm,clip]{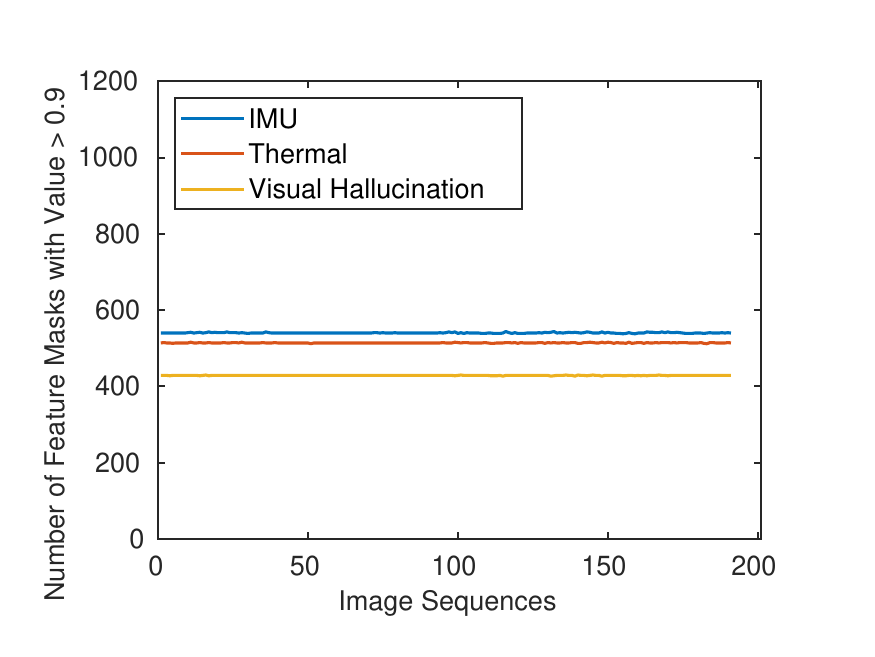}
        \vspace{-0.3cm}}
    \caption{\revise{Selective fusion mask for mobile robot data in (a) Seq 39 and (b) poster data. We plot the total number of masks for each feature modality with value higher than 0.9, indicating the importance of the features.}}
\label{fig:sf_mask}
\end{figure}

\revise{To understand the importance of hallucination network and the selective fusion mechanism, we plot the output feature representation, comparing RGB features, hallucination features, fusion mask, and attended hallucination features as seen in Fig. \ref{fig:features_visualization}. Fig. \ref{fig:features_visualization} (top) shows the hallucinated RGB features generated from the test data in the canteen sequence (Seq 39), which is the only canteen sequence we have (no other indoor structure that replicate this environment). It shows that the hallucination network produces accurate fake RGB features in this new environment, showing the generalization capability. We believe that as long as there are enough thermal features as the starting information, the hallucination network can perform meaningful hallucination by interpolating the RGB features. It can be seen as well from Fig. \ref{fig:sf_mask} (a) that in Seq 39, there are many cases when the selective fusion network used more hallucination features than other features (thermal and IMU), showing the importance of hallucination to improve the odometry accuracy. This is also supported by the quantitative results in terms of RPE as shown in Table \ref{table:evaluation_robot_rpe}.

However, there is always a case that hallucination network produce erroneous results. The example is shown in Fig. \ref{fig:features_visualization} (bottom) when there is not enough thermal gradient information to hallucinate rich RGB features. This happens when the camera faces a poster that has a similar temperature with the wall. In this case, selective fusion module will rely on the other modalities, i.e. IMU, in order to produce meaningful odometry. As we can see from Fig. \ref{fig:sf_mask} (b), in poster data, selective fusion utilizes more IMU features than other modalities and places the hallucination as the least useful features.}

\subsubsection{Accuracy}
\begin{figure}
    \vspace{-0.3cm}
    \includegraphics[width=1\columnwidth]{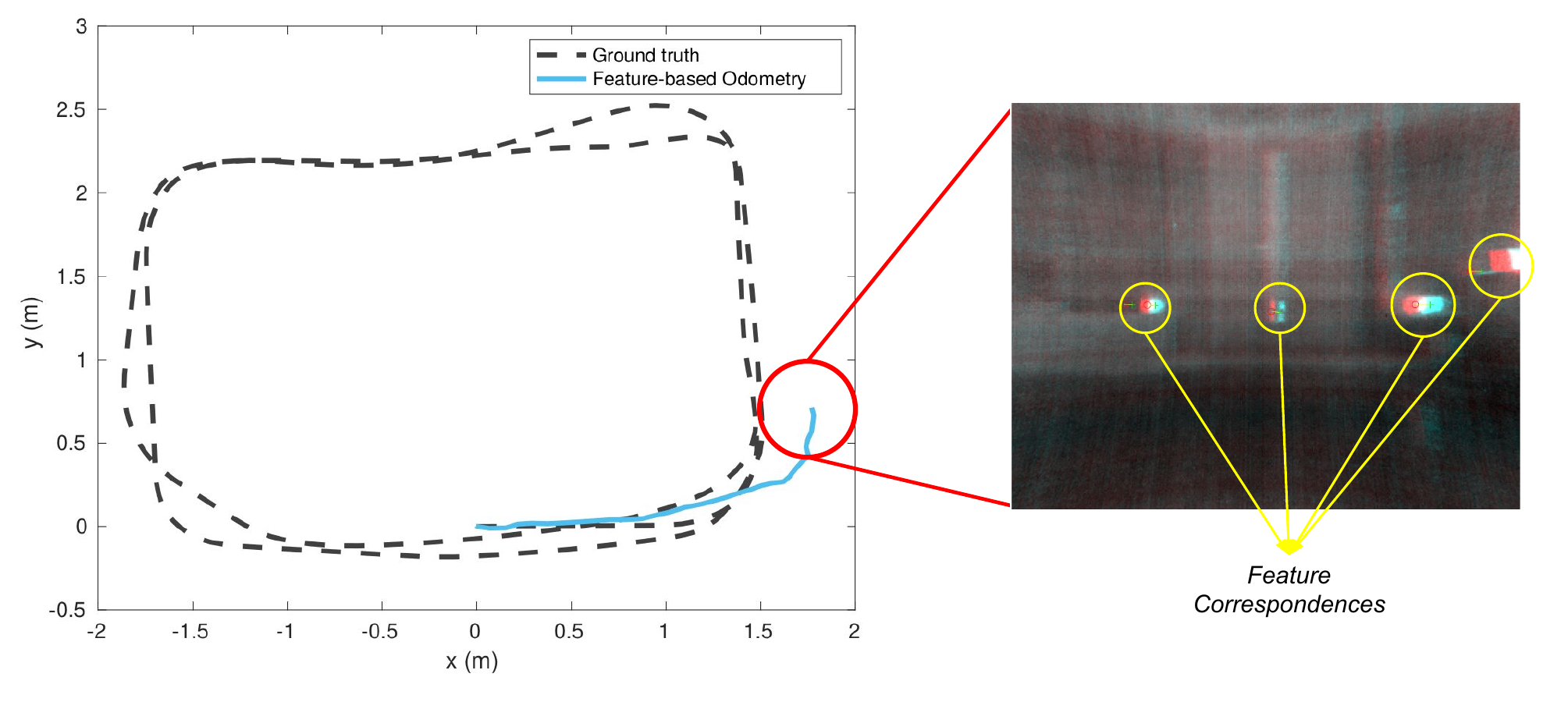}
    \vspace{-0.7cm}
    \caption{\revise{Feature-based approach (SURF) using standard descriptors loses tracks on scenes with poor thermal gradients. In this example not enough correspondences are matched between consecutive images.}}
\label{fig:feature_based_odom}
\end{figure}

\revise{To measure the quantitative performance of our odometry model, we report both RPE and ATE in Table \ref{table:evaluation_robot_rpe} and Table \ref{table:evaluation_robot_ate} respectively. We also compare our model with VINet \cite{clark2017vinet} applied on RGB and thermal, VINS-Mono \cite{qin2018vins} applied on RGB, and IMU assisted wheel odometry. As shown in Table \ref{table:evaluation_robot_rpe}, our neural odometry produces more accurate results compared to VINet. VINet applied on RGB falls short possibly due to the variation in lighting condition (e.g. dim and darkness), yielding sub-optimal performances. VINet applied on thermal generates less accurate estimation due to the difficulty of abstracting thermal data without the help from hallucination network. Our neural thermal-inertial odometry produces consistent results either from the perspective of RPE or ATE, and comparable to IMU assisted wheel odometry and VINS-Mono RGB (SLAM), showing the importance of hallucination network and feature selection in multi modal sensor fusion. \revisetwo{Note that in scenarios with benign lighting, VINS-Mono typically yield more accurate rotation as it exploits the RGB images which have richer features for accurate rotation estimation. Nevertheless, VINS-Mono fails to initialize or loses tracks due to unstable frame rate, abrupt motion, and the presence of dynamic objects (people) in front of the RGB camera, particularly in Seq 32, 33, 34, 37 and 39.} We also tried to run VINS-Mono with thermal camera (14-bit) but it lose tracks after couple of seconds to a minute due to abrupt motion and lack of features in the corridor. To investigate further the difficulty of tracking thermal features in our data, we tried to perform odometry estimation using strong feature matching algorithm like SURF \cite{bay2006surf} instead of using KLT tracker as demonstrated by VINS-Mono. The pose is then estimated by using five-point algorithm \cite{nister2004efficient} and bundle adjustment based on Computer Vision Toolbox applied in Matlab. For this purpose, we normalized 14-bit images by re-scaling it into 256 intensity around the median of thermal value. As we can see from Fig. \ref{fig:feature_based_odom}, SURF-based monocular thermal odometry loses tracks, specially following large changes in viewpoint and poor thermal gradients.}

\begin{figure}
    \vspace{-0.3cm}
        \includegraphics[width=4.3cm,trim=0.8cm .5cm 1cm .5cm,clip]{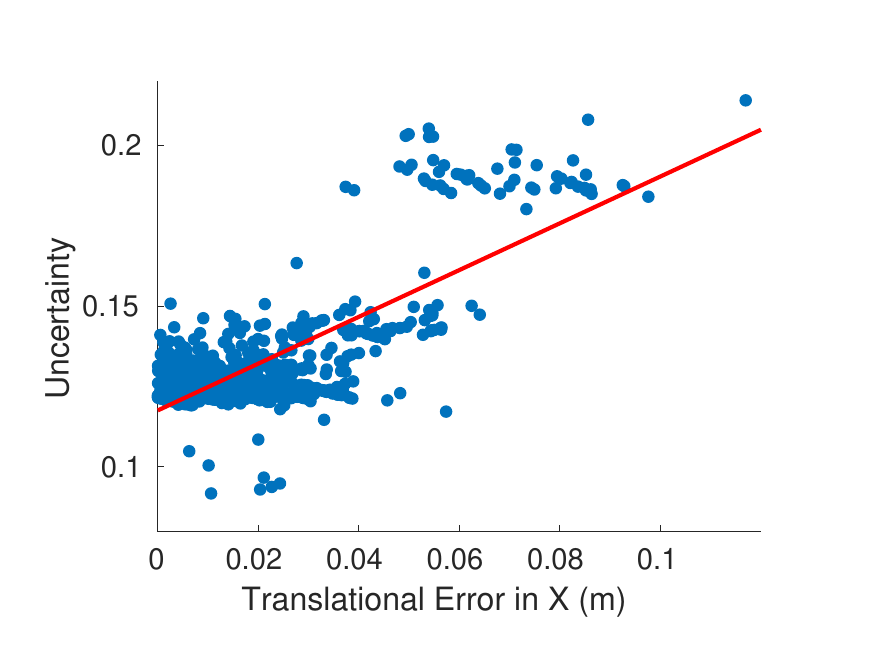}
        \includegraphics[width=4.3cm,trim=0.8cm .5cm 1cm .5cm,clip]{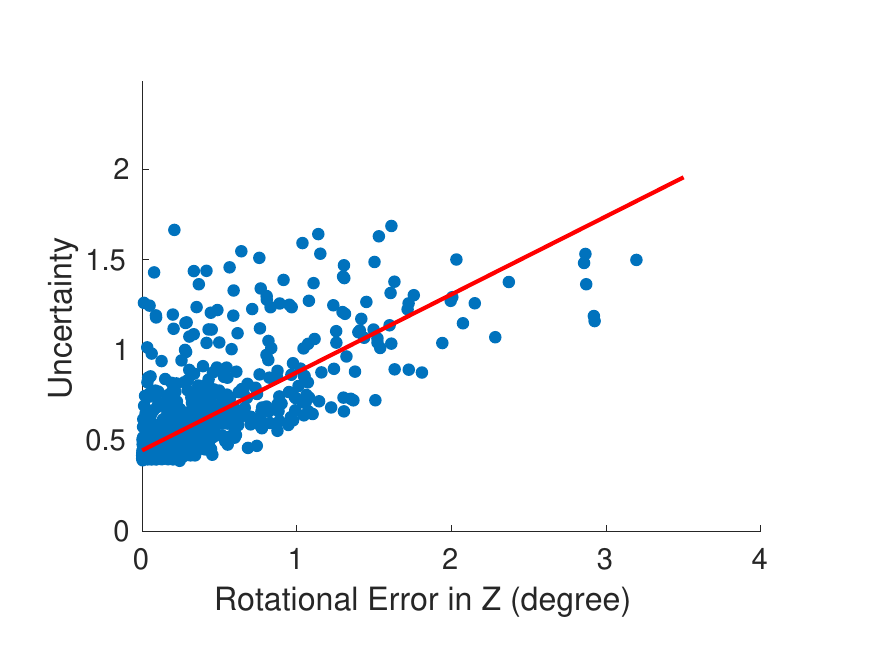}
        \vspace{-0.3cm}
    \caption{Translational errors (in X axis) and rotational errors (w.r.t Z axis) against uncertainty (standard deviation).}
\label{fig:uncertainty_vs_errors}
\end{figure}

\subsubsection{Probabilistic Estimates}
\label{sec:prob_estimates}
\revise{To validate the proposed probabilistic approach, we first analyse the importance of modelling the odometry output through a mixture of Gaussian, followed by interpreting the output variances. Table \ref{table:influence_mixing_coefficients} shows the influence of the number of Gaussian (indicated by the number mixing coefficients $K$) on accuracy. As it can be seen, by increasing the number of $K$, the accuracy also increases, showing that the network models the 6-DoF pose distribution more accurately by using a mixture of Gaussian instead of a single Gaussian ($K=1$). Nevertheless, at some point, the accuracy saturates or even degrades.}

Fig. \ref{fig:uncertainty_vs_errors} plots the uncertainty (variance) estimate against translational and rotational errors. We show the comparison for rotation in Z axis as the other axes (X and Y) mostly remain unchanged during operation. For translation, most changes happen in X and Y directions while translation in Z axis are almost zero since the robot moves in flat surface. The figure shows that the approximate uncertainty is correlated with the odometry error, validating our approach.

\begin{figure}
    \subfloat[Seq 46]{
         \includegraphics[width=0.9\columnwidth]{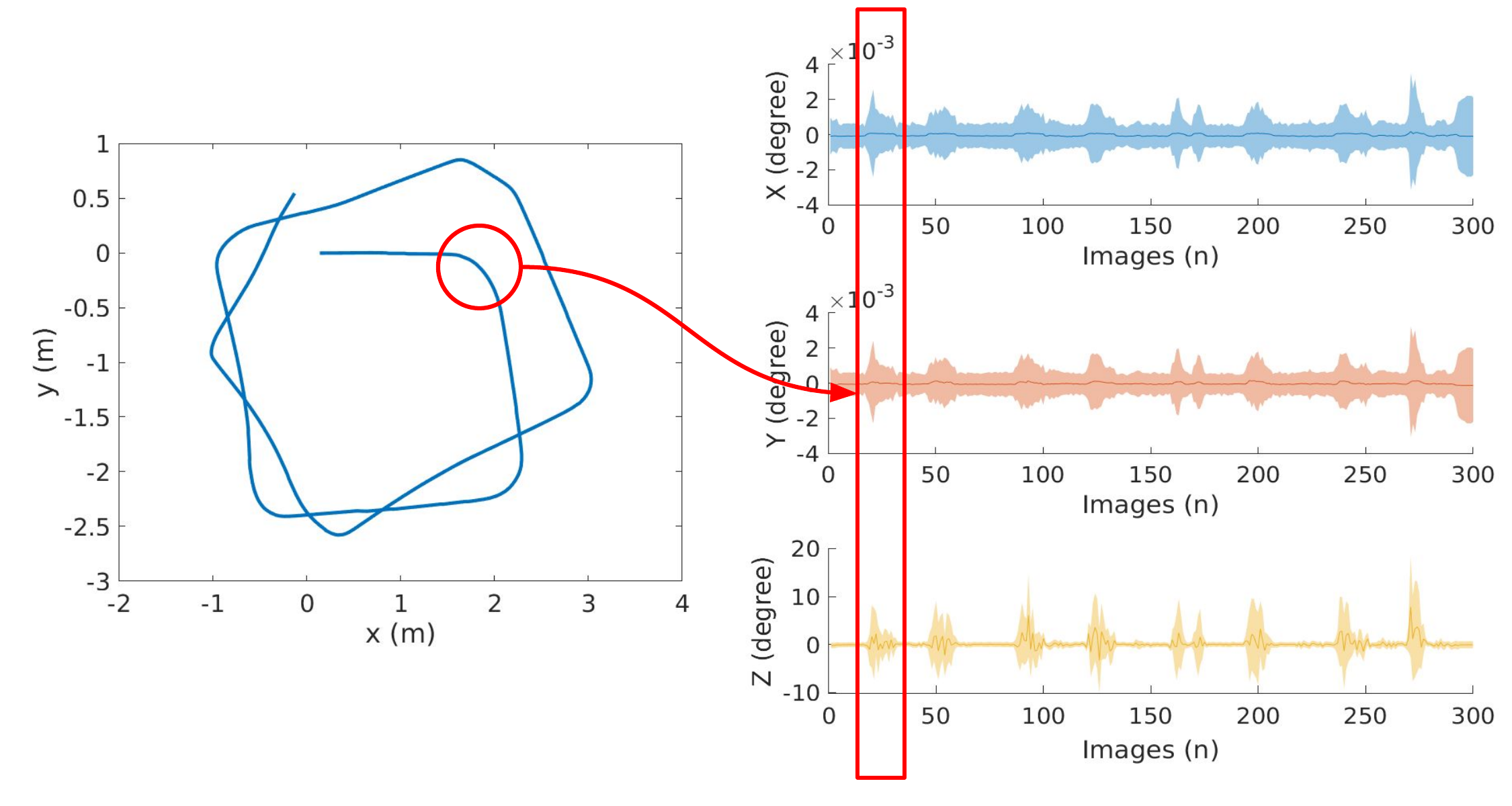}
        }
        \vspace{-0.4cm}
    \\    
    \subfloat[Seq 49]{
        \includegraphics[width=0.9\columnwidth]{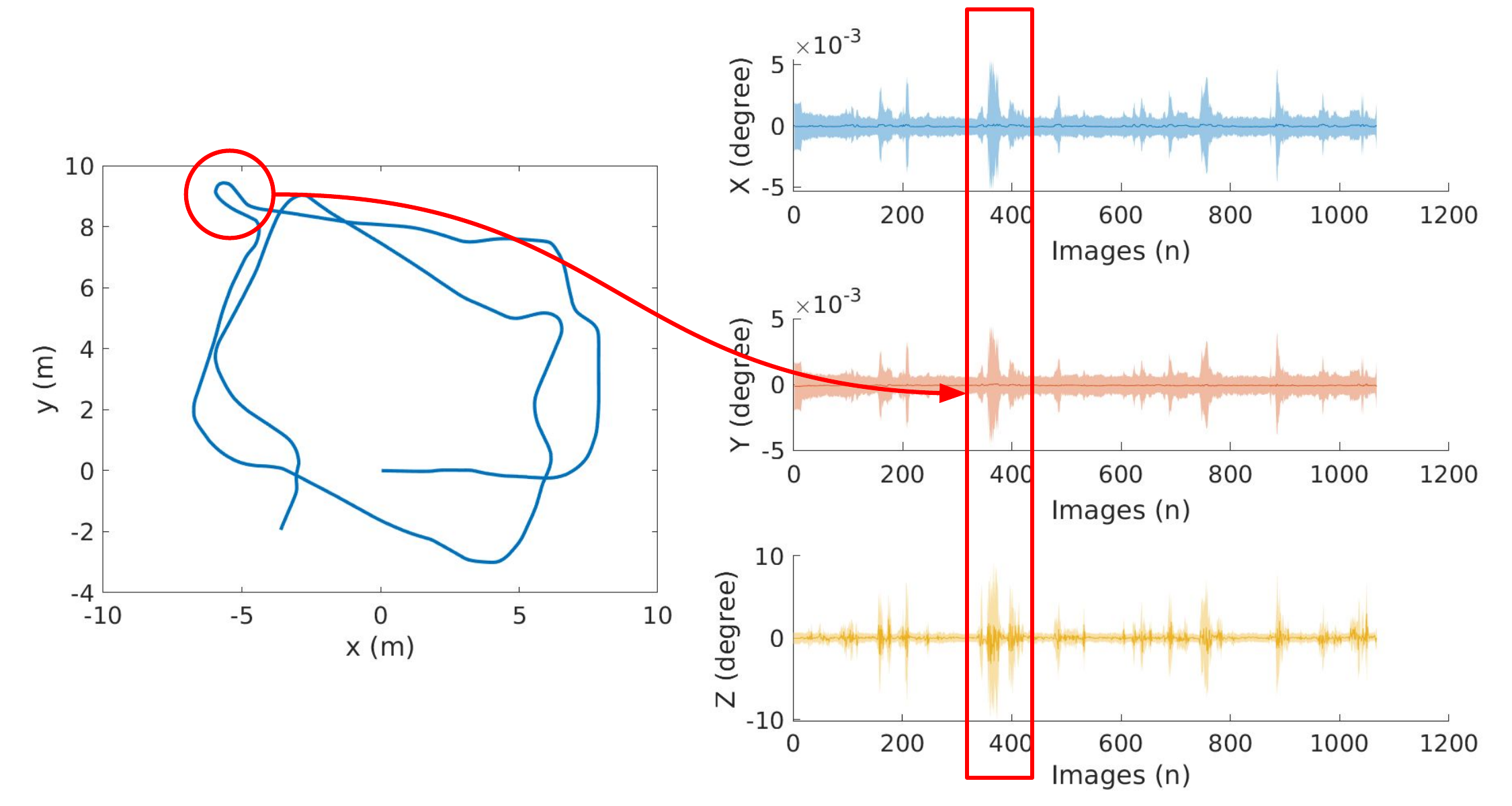}
        }
    \vspace{-0.1cm}
    \caption{\revise{Relative pose errors from the neural odometry networks against $3\sigma$ variances interval.}}
\label{fig:uncertainty_estimates_6dof}
\end{figure}

\revise{To have a better understanding in which condition the network produces larger uncertainty, we plot the rotational uncertainty in X, Y, and Z direction for Seq 46 and Seq 49 in Fig. \ref{fig:uncertainty_estimates_6dof}. Following the practice in \cite{wang2018end}, for better visualization, we draw the rotational errors against $3\sigma$ variance interval. As one can see, the rotational errors are located within the variance intervals, verifying the meaningful of uncertainty estimation using MDN. In Seq 46, we can see that the network yields larger variance when the mobile agents perform large rotation (around 90 degree rotation in Z axis). In Seq 49, we can also observe that the largest uncertainty takes place when the robot performs U-turn, which typically generates the largest error in odometry estimation.} This is interesting since this ability to approximate the variance is learnt during training without supervision on uncertainty. In this sense, we can validate that the uncertainty estimation can be used as a valid constraint for SLAM optimization.

\begin{table}[t]
\centering
\begin{threeparttable}
  \caption{\revise{The Influence of The Number of Mixing Coefficients ($K$) on Accuracy (ATE)}}
  \renewcommand{\arraystretch}{1}
  \setlength\tabcolsep{6pt}
  \fontsize{9}{10}\selectfont
  \label{table:influence_mixing_coefficients}
  \begin{tabular}{|c|c|c|c|c|c|}
    \hline
    \multirow{2}*{ATE} & \multicolumn{5}{c|}{The Number of $K$} \\
    \cline{2-6}
    & 1 & 5 & 10 & 15 & 20 \\
    \hline
    \hline
    Mean (m) & 0.793 & 0.725 & \textbf{0.609} & 1.051 & 0.862 \\
    Std (m) & 0.449 & 0.435 & \textbf{0.357} & 0.441 & 0.606 \\
    \hline
  \end{tabular}
\end{threeparttable}
\end{table}

\subsubsection{Computation Cost}
The neural odometry model was trained on an NVIDIA TITAN V GPU. It required approximately 6-18 hours for training the hallucination network and around 6-20 hours for training the remaining networks. The model contains 273 millions of parameters, requiring 547 MB of disk space. \revise{To generate a single prediction, the model requires approximately 0.5s in a standard CPU, which is typically slower than the real-time implementation of visual-inertial odometry (e.g., VINS-Mono \cite{qin2018vins}) which can produce the camera pose between 0.05-0.1s. Nevertheless, our model can be executed for up to 26 Hz (0.039s required for a single inference) in a powerful GPU like NVIDIA TITAN V.}


\begin{figure*}
    \includegraphics[width=2\columnwidth]{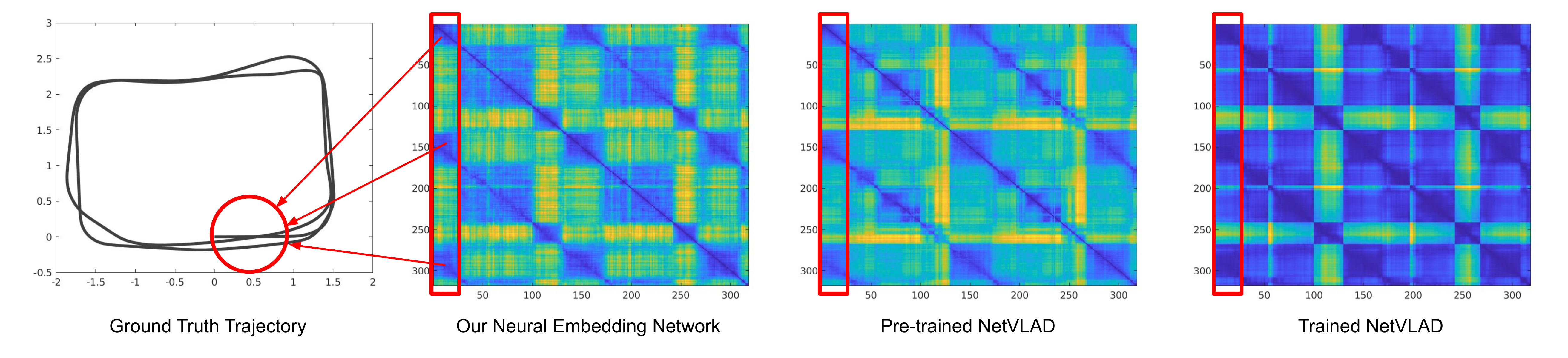}
    \vspace{-0.5cm}
    \caption{Similarity matrix produced by our neural embedding network on Seq 2, compared to NetVLAD pre-trained on RGB and re-trained on thermal. Blue and yellow color indicates the most similar and the most dissimilar pair. Note that our model can produce distinct embedding to identify loop closure.}
\label{fig:similarity_matrix}
\end{figure*}

\begin{figure}
        \vspace{-0.8cm}
        \subfloat[BoTW \cite{tsintotas2019probabilistic} on RGB]{
        \includegraphics[width=4.3cm,trim=0.8cm .5cm 1cm 1.1cm,clip]{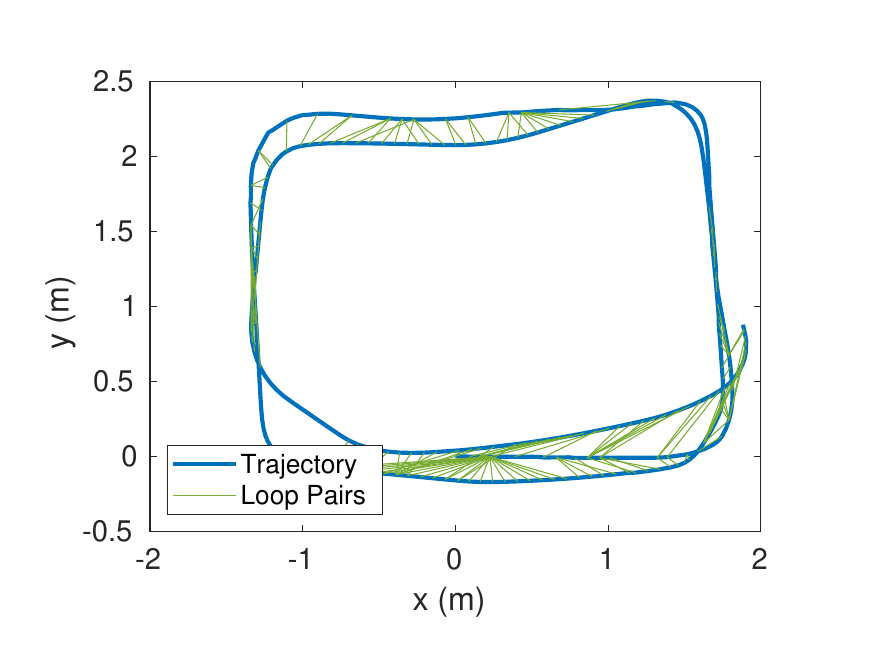}}
        \subfloat[BoTW \cite{tsintotas2019probabilistic} on Thermal]{
        \includegraphics[width=4.3cm,trim=0.8cm .5cm 1cm 1.1cm,clip]{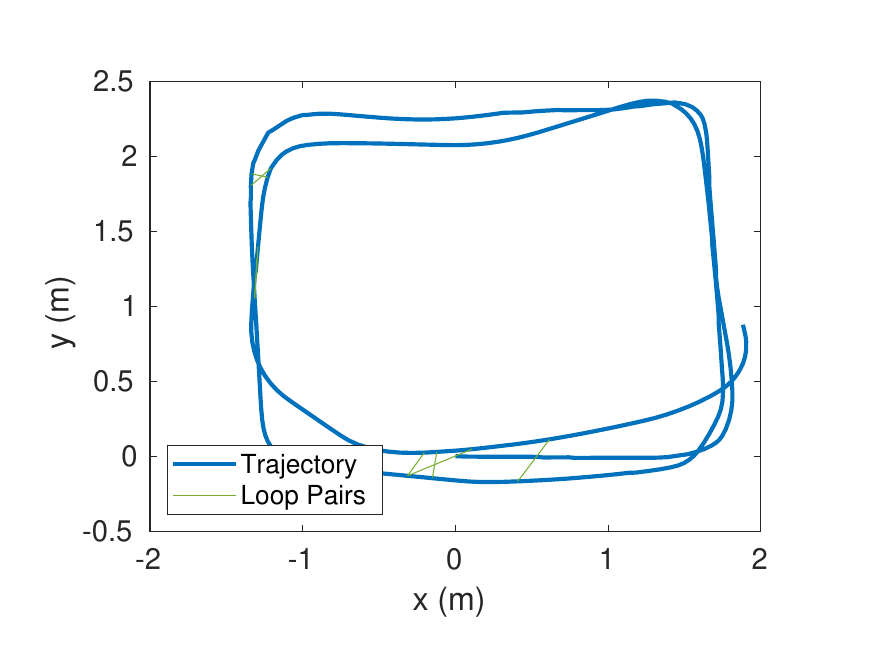}}
        \vspace{-0.4cm}
        \\ 
        \subfloat[Trained NetVLAD (14-bit) \cite{arandjelovic2016netvlad}]{
        \includegraphics[width=4.3cm,trim=0.8cm .5cm 1cm .5cm,clip]{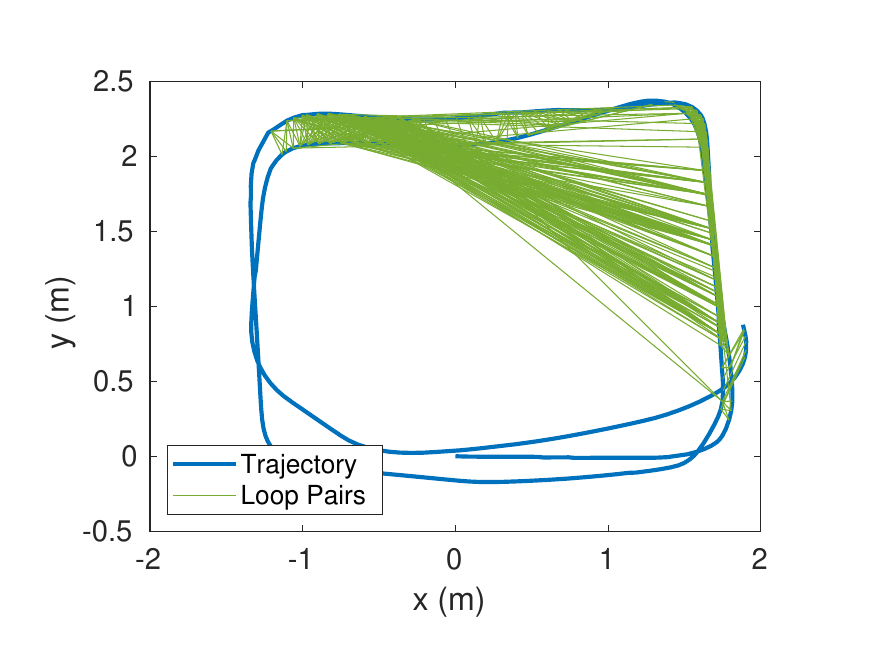}}
        \subfloat[Trained NetVLAD (8-bit) \cite{arandjelovic2016netvlad}]{
        \includegraphics[width=4.3cm,trim=0.8cm .5cm 1cm .5cm,clip]{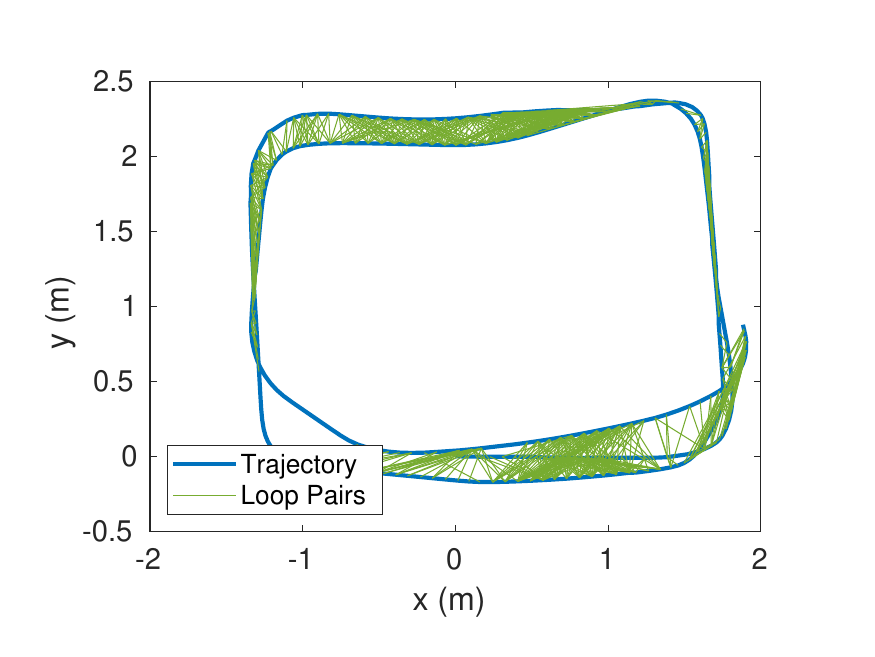}}
        \vspace{-0.4cm}
        \\
        \subfloat[Our model (14-bit)]{
        \includegraphics[width=4.3cm,trim=0.8cm .5cm 1cm .5cm,clip]{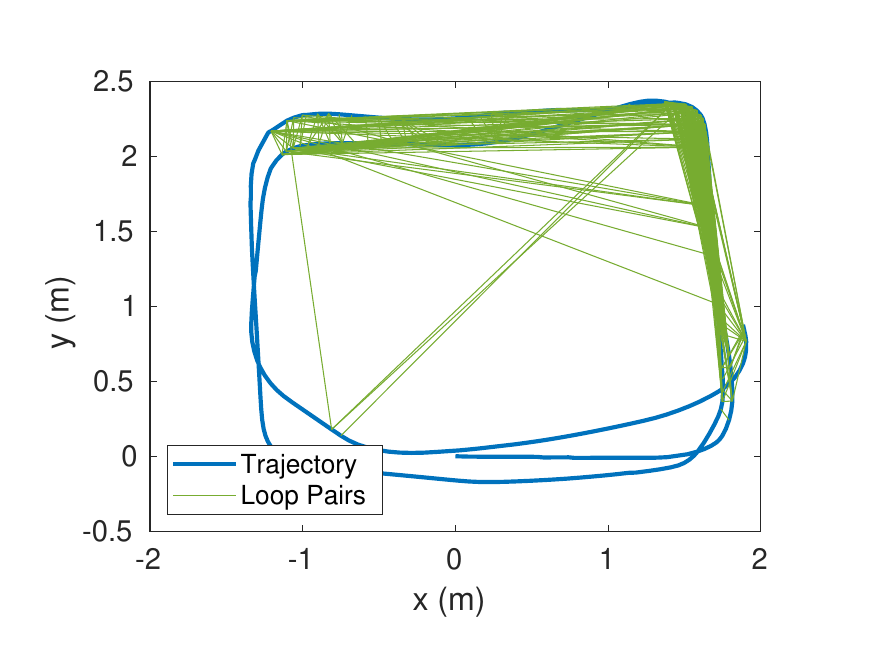}}
        \subfloat[Our model (8-bit)]{
        \includegraphics[width=4.3cm,trim=0.8cm .5cm 1cm .5cm,clip]{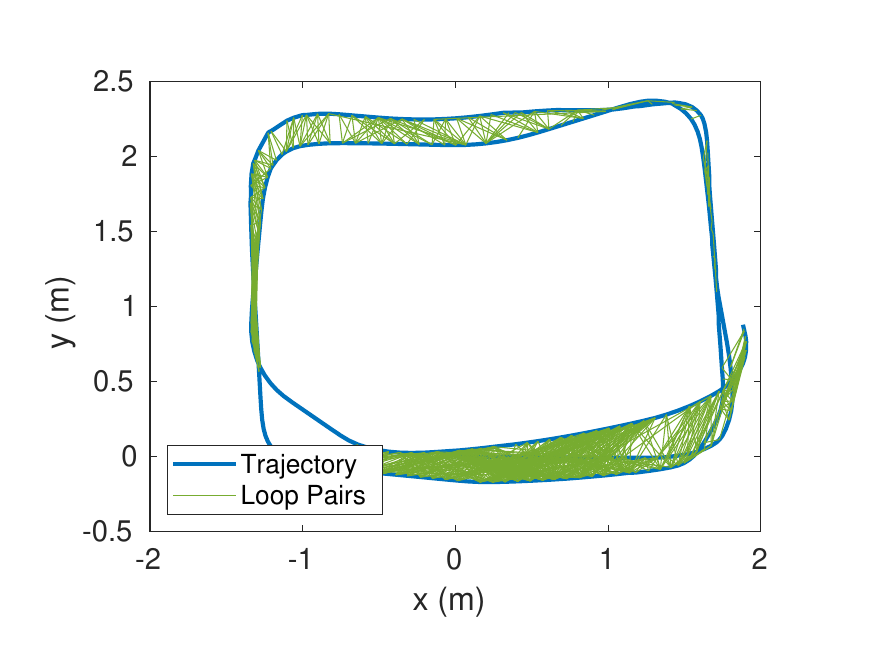}}
        \vspace{-0.05cm}
    \caption{\revise{Loop closure detection on Seq 33 from BoTW (applied on RGB and thermal imagery), trained NetVLAD (on thermal), and our neural embedding networks (either using 14-bit or 8-bit representation). Our model can produce similar performance to NetVLAD (8-bit) but with only 0.4\% space requirement to store the embedding features.}}
\label{fig:loop_pair}
\end{figure}

\subsection{Validating Loop Closure Detection and Loop Closure Constraints in Indoor Ground Robot Data}

\subsubsection{\revise{Qualitative and Quantitative Results for Loop Closure Detection}}
We perform qualitative validation of our loop closure detection by plotting the detected loop pair on top of the output odometry. For comparison, we also show the loop closure detection from state-of-the-art feature-based approach (i.e. BoTW \cite{tsintotas2019probabilistic}) and deep learning approach for place recognition (i.e. NetVLAD \cite{arandjelovic2016netvlad}). For NetVLAD, we generate the results from the pre-trained model from Pittsburgh dataset (RGB images) and also from the re-trained weights on our thermal images.

Before plotting the output loop closure pair, we inspect what the neural embedding network learn during training and compare it with NetVLAD, either applied on RGB (pre-trained) or applied on thermal (re-trained). Fig. \ref{fig:similarity_matrix} depicts the similarity matrix generated by measuring the cosine distance between embedding vectors on Seq 33. When we compare few images in the beginning of sequence with all other images in the whole sequence (red square area in Fig. \ref{fig:similarity_matrix}), our neural embedding network identifies two areas with the highest similarity (excluding the adjacent frames). If we look at the ground truth trajectory, those two areas belong to the same place when the mobile agent re-visit the starting point. This indicates that our network can produce meaningful and distinct embedding vectors, identifying loop closure by clustering the same places into similar embedding. While NetVLAD can also identify two similar regions, they are less distinctive (fewer area with strong dissimilarity/yellow colored). We presume that this is because the performance of NetVLAD largely depends on the clustering algorithm. If the algorithm wrongly clusters local features due to the similar characteristic of particular thermal features (two different RGB patches might look similar on thermal as it lacks texture), it might classify two different scenes as an identical ones (with some degree of certainty).

Fig. \ref{fig:loop_pair} depicts the detected loop pair on the output odometry given by BoTW (on RGB and on thermal), NetVLAD (on thermal), and our loop closure detection. As it can be seen, BoTW clearly produces robust performance on RGB. However, it performs badly on thermal imagery, yielding a very small number of correctly detected loop. This emphasizes the difficulty of performing data association on thermal images due to the lack of robust features. Our model, as expected, can perform very well, detecting large number of positive loop pairs. Nevertheless, by carefully setting the threshold, NetVLAD (thermal) can also produce a similar number of positive loop pair. \revise{However, this is done with the cost of large space requirement and slow computation time as described in Section \ref{sec:loop_closure_effifiency}.}

\begin{figure}
\centering
    \vspace{-0.1cm}
    \includegraphics[width=0.65\columnwidth]{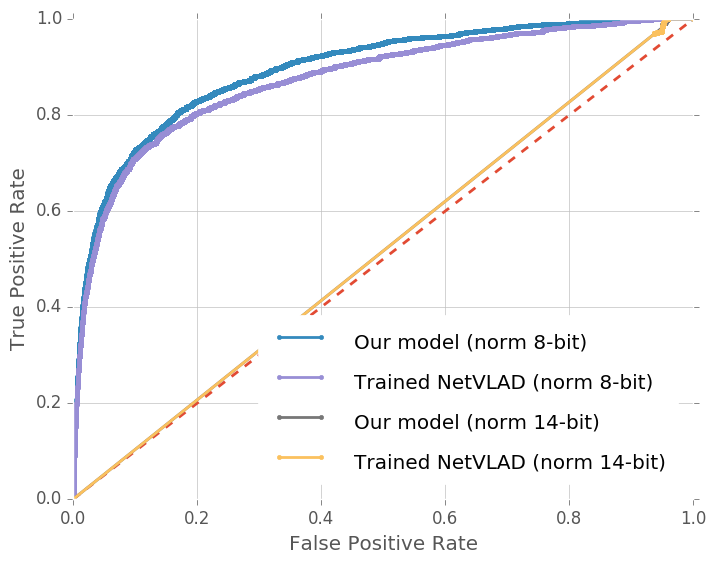}
    \vspace{-0.1cm}
    \caption{\revise{ROC curves between Neural Embedding (our model) and trained NetVLAD applied on thermal data for all sequences in ground robot data.}}
\label{fig:roc_curve}
\end{figure}

\revise{For quantitative experiments, we compare our loop closure detection with respect to BoTW applied on RGB. We plot a Receiver Operating Characteristic (ROC) curve to measure the trade-off between sensitivity (true positive rate-TPR) and specificity (false positive rate-FPR) for every possible cut-off as it has been used by previous work on place recognition \cite{torralba2003context, lategahn2013learn, dube2017segmatch}. As it can be seen from Fig. \ref{fig:roc_curve}, our model (normalized 8-bit) produces slightly better performance than the NetVLAD (normalized 8-bit). Given $20\%$ FPR, our model obtains around $82\%$ TPR, showing a good trade-off between TPR and FPR.}

\subsubsection{\revise{The Impact of Thermal Representation for Loop Closure Detection}}
\revise{In Fig. \ref{fig:loop_pair} (c) and (e), we also display the loop pair detected from both NetVLAD and our embedding model by using normalized (between 0 and 1) 14-bit representation. For this purpose, we alter the discrepancy threshold $\mathcal{S}_{ij}$ such that we can obtain a similar amount of loop pair compared to the normalized 8-bit representation. As it can be seen, either by using NetVLAD or our embedding model, training loop closure detection by using normalized 14-bit thermal representation produces many wrong loop pair and becomes unusable. This happens most likely because both models utilize the pre-trained ResNet as the feature extractor which relies heavily on the image appearance properties depicted by 8-bit RGB images. Quantitative measures shown by ROC curves in Fig. \ref{fig:roc_curve} also indicate that 14-bit models produce very bad performance (close to random guess), further reinforcing the qualitative results in Fig. \ref{fig:loop_pair} (c) and (e).}

\subsubsection{Efficiency of Neural Embedding Network}
\label{sec:loop_closure_effifiency}
Table \ref{table:efficiency_neuloop} shows the disk space required (MB) to save the embedding vectors from NetVLAD compared to our model. Our model requires much less space as it can represents a thermal image by using only 128 vectors. On the other hand, NetVLAD needs to produce 32768-dimensional VLAD vectors, which requires almost 7 GB disk space to save embedding vectors for all sequences. On the other hand, our model only need 27.3 M, $0.4\%$ from what NetVLAD requires. \revise{In terms of runtime efficiency, NetVLAD takes 2.12 second (in CPU) to generate the embedding vectors from a single image, while our model takes only 0.63 second ($3.4 \times$ faster). However, this is again slower than the real time implementation of loop closure detection in VINS-Mono (15-25Hz), although our model can also reach real time performance (27Hz, 0.037s per single inference) when it was tested in powerful TITAN V GPU.}

\begin{table}[t]
\centering
\begin{threeparttable}
  \setlength{\tabcolsep}{5pt}
  \caption{Space Requirement (MB) to Save Embedding Features in Ground Robot Data}
  \renewcommand{\arraystretch}{1}
  \setlength\tabcolsep{6pt}
  \fontsize{9}{10}\selectfont
  \label{table:efficiency_neuloop}
  \begin{tabular}{|c|c|c||c|}
    \hline
    Seq & Files & NetVLAD & Our \\
    \hline
    \hline
    32 & 2019-10-24-18-22-33 & 248 & \textbf{1.1} \\
    33 & 2019-11-23-15-54-25 & 266 & \textbf{1} \\
    34 & 2019-11-23-15-52-53 & 244 & \textbf{0.9} \\
    37 & 2019-11-23-15-59-12 & 362 & \textbf{1.4} \\
    39 & 2019-11-04-20-29-51 & 855 & \textbf{3.3} \\
    42 & 2019-11-22-10-10-00 & 789 & \textbf{3.1} \\
    43 & 2019-11-22-10-14-01 & 782 & \textbf{3.2} \\
    44 & 2019-11-22-10-22-48 & 829 & \textbf{3.2} \\
    45 & 2019-11-22-10-26-42 & 728 & \textbf{2.8} \\
    46 & 2019-11-22-10-34-57 & 252 & \textbf{1} \\
    47 & 2019-11-22-10-37-42 & 263 & \textbf{1} \\
    48 & 2019-11-22-10-38-47 & 472 & \textbf{1.8} \\
    49 & 2019-11-28-15-40-10 & 895 & \textbf{3.5} \\
    \hline
    \hline
    \multicolumn{2}{|c|}{Total} & 6985 & \textbf{27.3} \\
    \hline
  \end{tabular}
\end{threeparttable}
\end{table}

\subsubsection{Validating Loop Closure Constraints}
Being able to correctly detect loop pair is indeed an important stage in SLAM pipeline. However, estimating accurate poses between the loop closure constraints is actually more important. Even if we have large number of true positive loop pair, if the majority of relative poses among them are largely erroneous, it will badly impact the back end optimization, making the trajectory estimation even worse. To this end, we will validate the accuracy of our neural loop closure network. To have a better perspective on the accuracy, we compare our result with the standard feature-based pose estimation. Note that we have to re-scale the pose estimation generated by feature-based approach by using ground truth pose, since the estimation is correct only up to a scale. \revisetwo{On the other hand, our model learns to implicitly estimate the scale by learning it from the ground truth during training.}

\begin{figure}
    \vspace{-0.1cm}
    \subfloat[Feature-based approach]{
         \includegraphics[width=4.3cm,trim=0.7cm .5cm 1cm .5cm,clip]{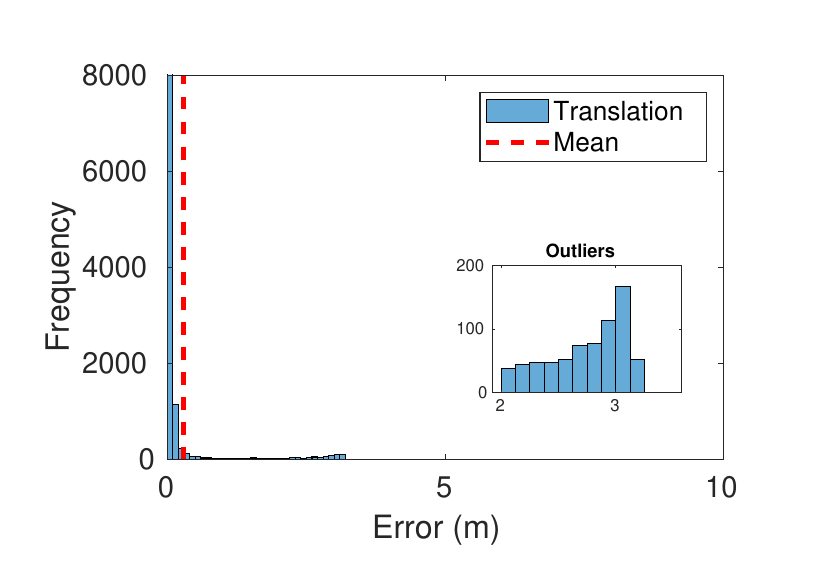}
         \includegraphics[width=4.3cm,trim=0.7cm .5cm 1cm .5cm,clip]{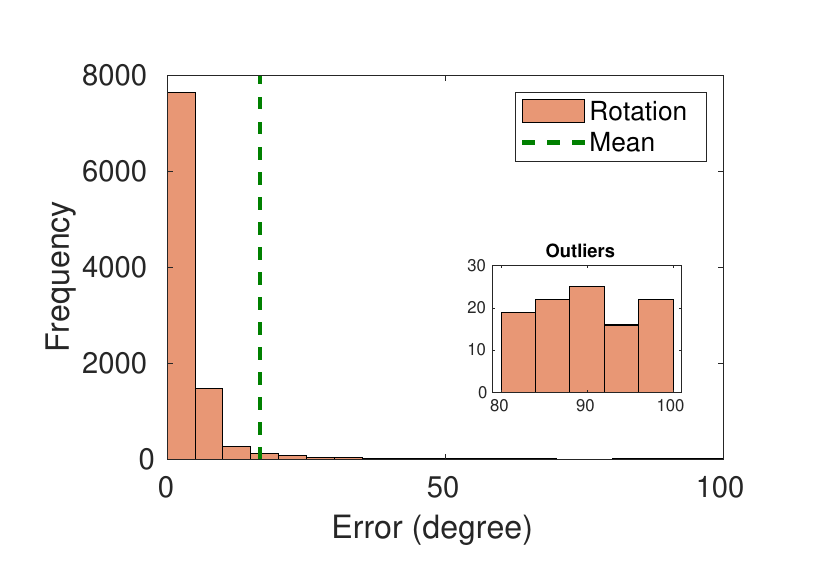}
        }
        \vspace{-0.4cm}
    \\    
    \subfloat[Neural loop closure]{
        \includegraphics[width=4.3cm,trim=0.7cm .5cm 1cm .5cm,clip]{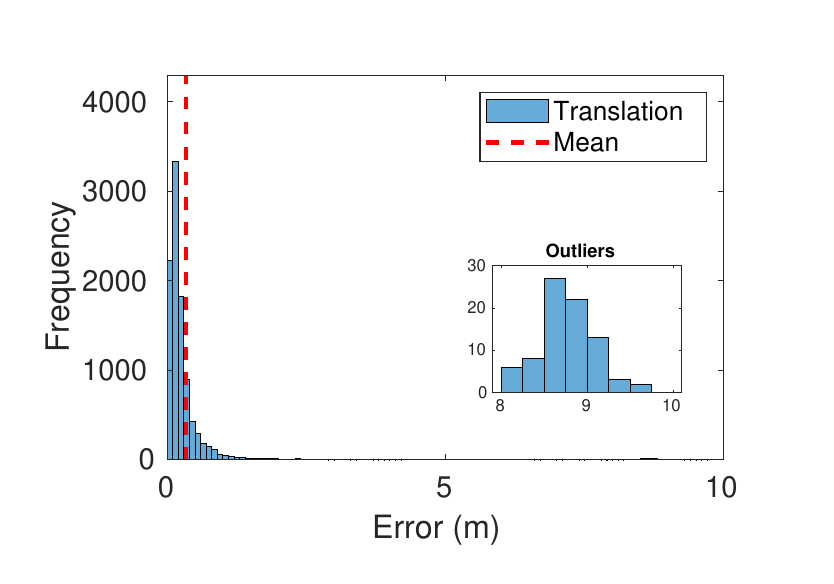}
        \includegraphics[width=4.3cm,trim=0.7cm .5cm 1cm .5cm,clip]{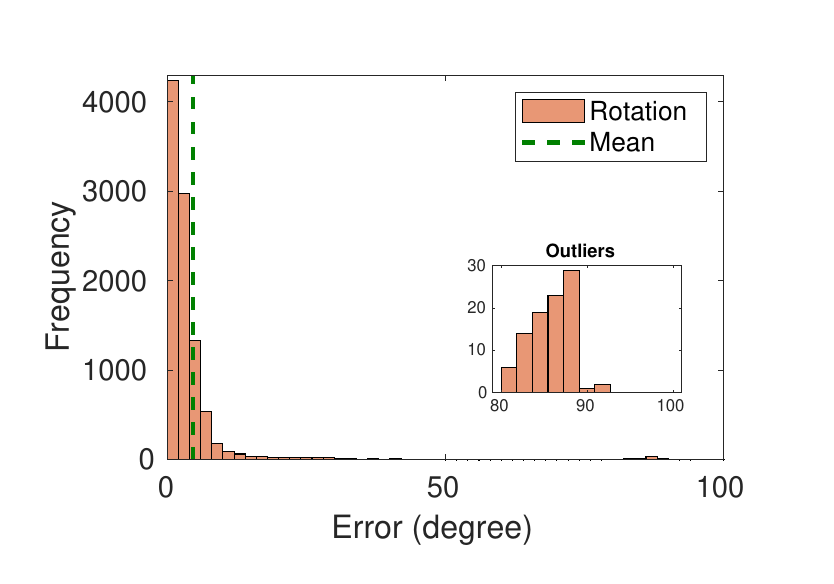}
        }
    \vspace{-0.1cm}
    \caption{\revise{Comparison of error distribution between neural loop closure and SURF-based pose estimation approach for all ground robot sequences. Note that the translation estimation of SURF-based approach are scaled using the ground truth and only $89\%$ of error data are displayed as the remaining fail to obtain sufficient correspondences.}}
\label{fig:loop_pose_err}
\end{figure}

\revise{Fig. \ref{fig:loop_pose_err} (a) and (b) describes the error distribution for translation and rotation component for all ground robot sequences generated by feature-based approach (SURF and 5-point algorithm) and neural loop closure respectively. Note that the translation estimation for SURF-based pose estimation is scaled with the ground truth. As one can see, our model produces robust and accurate results compared to the feature-based approach with around 0.344 m and 4.581 degree translation and rotation error respectively. However, there are conditions when the network produces large error possibly when it faces with NUC, large baseline scenario, or heavily featureless scene, in which hallucination network cannot even help. This is why outlier rejection in the back end becomes important component to generate accurate SLAM estimation.}

\subsection{SLAM Performance in Ground Robot Data}
\subsubsection{Accuracy}
\begin{figure}[!ht]
    \centering
    \vspace{-0.5cm}
    \subfloat[Seq 37]{
        \hspace{-0.5cm}
        \includegraphics[width=4.7cm,trim=1cm .9cm 1cm .5cm,clip]{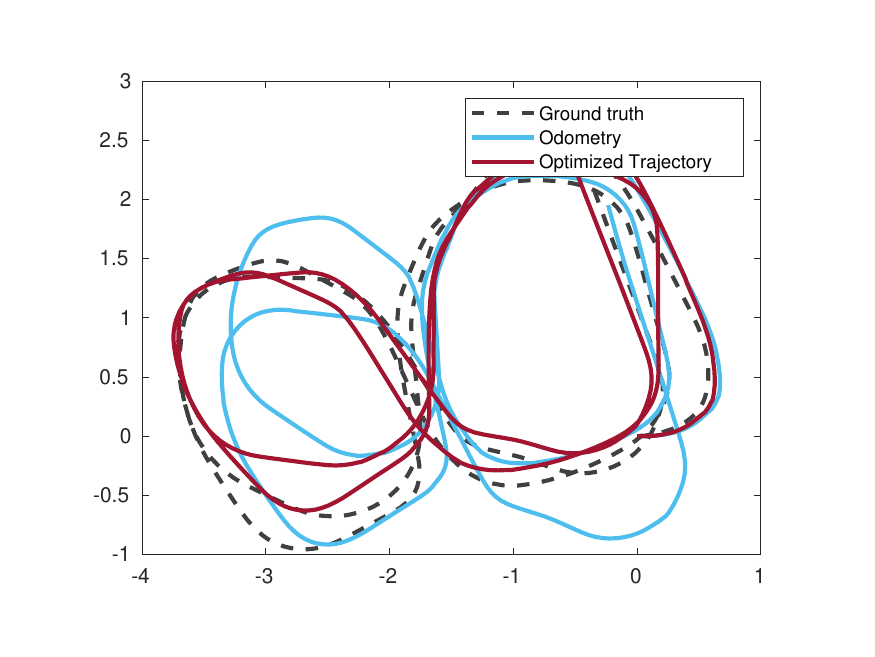}
        \hspace{-0.5cm}}
    \subfloat[Seq 39]{
        \includegraphics[width=4.7cm,trim=1cm .9cm 1cm .5cm,clip]{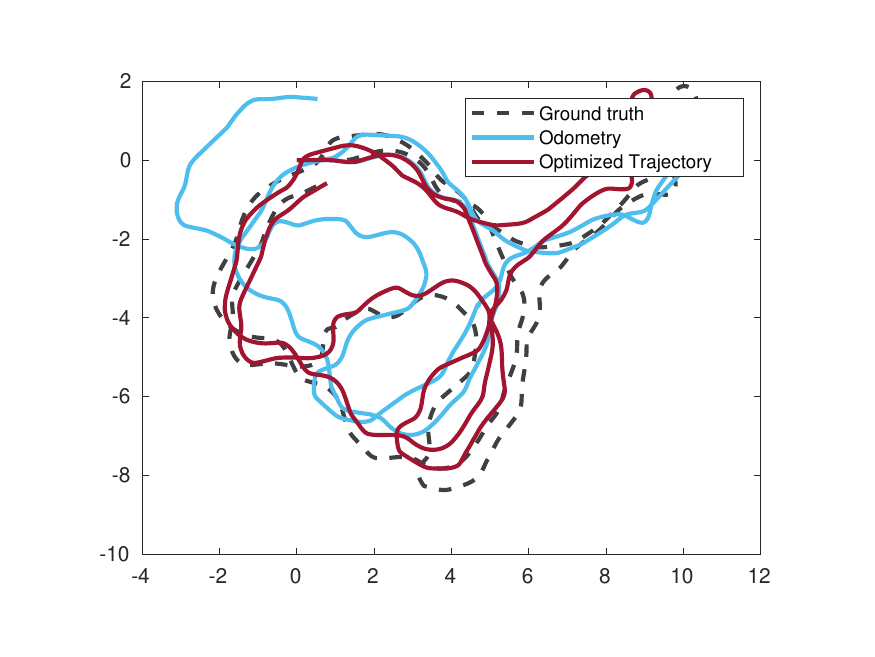}} \\
        \vspace{-0.5cm}
    \subfloat[Seq 42]{
        \hspace{-0.5cm}
        \includegraphics[width=4.7cm,trim=1cm .9cm 1cm .3cm,clip]{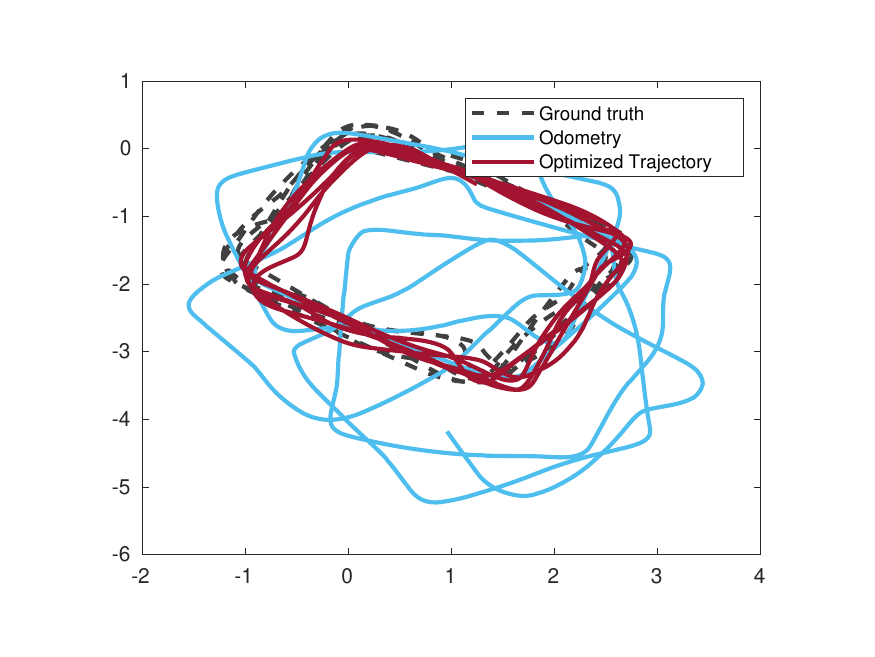}
        \hspace{-0.5cm}}
    \subfloat[Seq 43]{
        \includegraphics[width=5cm,trim=1cm .9cm 0.3cm .3cm,clip]{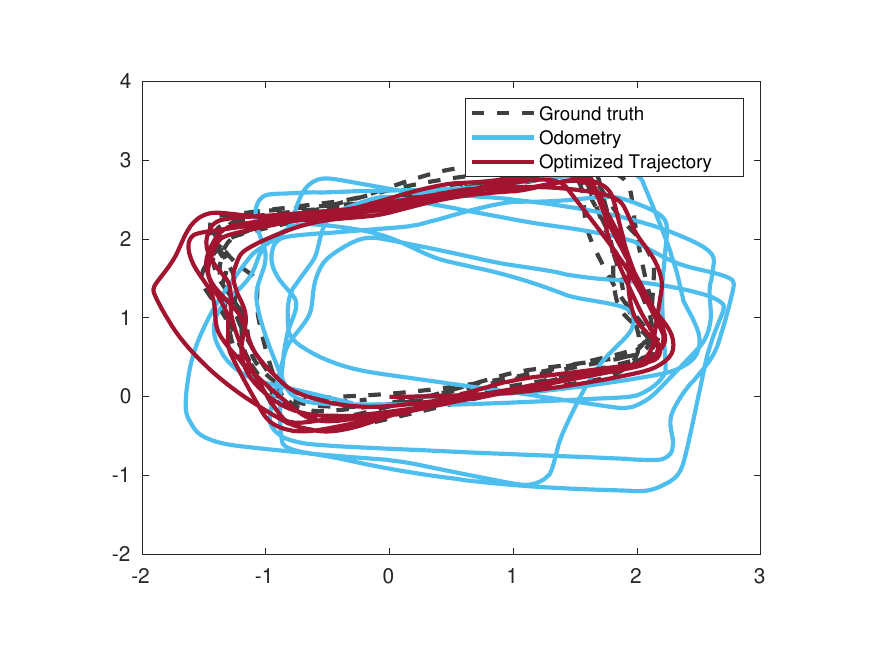}
        \hspace{-0.5cm}} \\
        \vspace{-0.5cm}
    \subfloat[Seq 46]{
        \hspace{-0.8cm}
        \includegraphics[width=4.7cm,trim=1cm .9cm 1cm .3cm,clip]{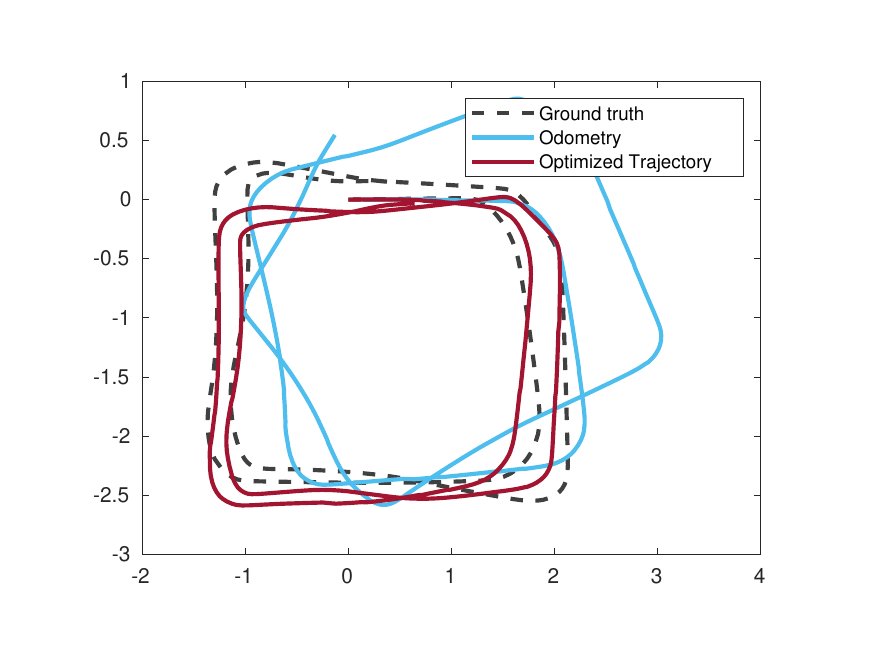}
        \hspace{-0.5cm}}
    \subfloat[Seq 49]{
        \includegraphics[width=4.7cm,trim=1cm .9cm 1cm .3cm,clip]{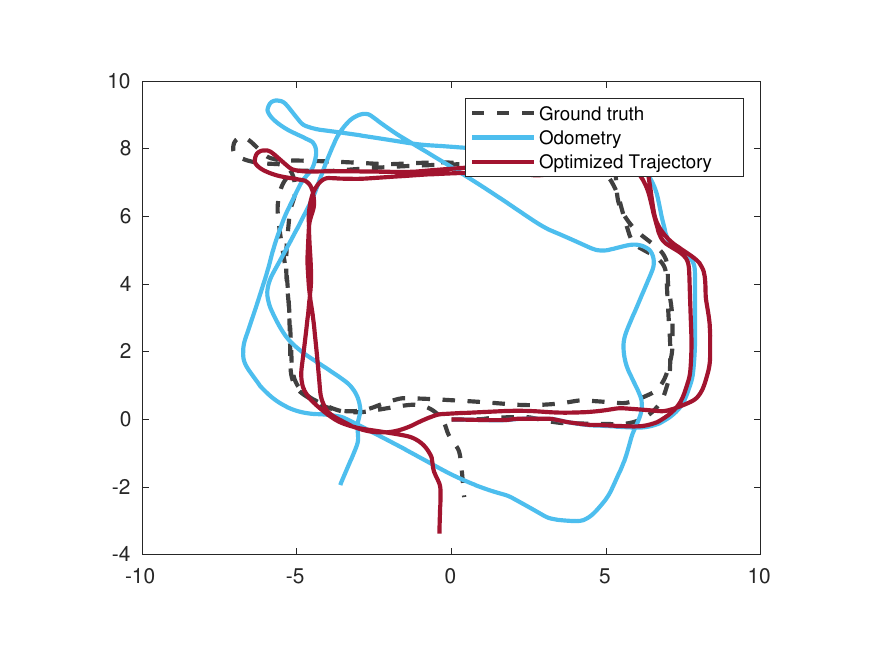}
        \hspace{-0.5cm}}
    \caption{Qualitative result of TI-SLAM system in some ground robot data. Both odometry and optimized odometry are generated from thermal-inertial data.}
\label{fig:trajectories_groundrobot}
\end{figure}

Table \ref{table:evaluation_robot_ate} lists the ATE of TI-SLAM for all test sequences. It can be seen that the complete SLAM system produces much better accuracy than the output trajectory from neural thermal-inertial odometry, showing an increase of $53.9\%$. In some sequences, the loop closure constraints can even improve the accuracy more than $70\%$, especially for a long trajectory that contains many loop pair (Seq 42, Seq 43, and Seq 46). \revise{On average, it yields 0.281 m errors, an order of magnitude smaller than VINet applied on thermal and better than VINS-Mono (RGB). Note that we even align VINS-Mono trajectory with the ground truth using \cite{umeyama1991least}. In some sequences, TI-SLAM even generates more accurate performance than the IMU assisted wheel odometry (e.g. Seq 39, Seq 42, Seq 46 and Seq 48), showing the efficacy of our thermal-inertial SLAM system.} To have better qualitative perspective, Fig. \ref{fig:trajectories_groundrobot} shows some output trajectories.

\begin{table}[t]
\centering
\begin{threeparttable}
  \renewcommand{\arraystretch}{1}
  \setlength\tabcolsep{1pt}
  \caption{\revise{RMS Absolute Trajectory Errors (m) in Indoor Ground Robot Data}}
  \fontsize{9}{10}\selectfont
  \label{table:evaluation_robot_ate}
  \begin{tabular}{|c|cc|ccc|cc|}
    \hline
    Seq & VINet & VINet & TI & \multirow{2}*{TI-SLAM} & \multirow{2}*{Gain} & \revise{VINS\textsuperscript{*}} & IMU+ \\
     & (RGB) & (Thermal) & odometry & & & \revise{(RGB)} & Wheel \\ 
    \hline
    \hline
    32 & 1.453 & 2.195 & 0.308 & \textbf{0.240} & 22.1\% & - & 0.123 \\
    33 & 0.565 & 2.032 & 0.289 & \textbf{0.182} & 37\% & - & 0.067 \\
    34 & 1.583 & 0.804 & 0.364 & \textbf{0.275} & 24.5\% & - & 0.073 \\
    37 & 1.931 & 2.309 &  0.249 & \textbf{0.164} & 33.9\% & - & 0.076 \\
    39 & 5.309 & 5.975 &  0.916 & \textbf{0.448} & 51\% & - & 0.546 \\
    42 & 2.670 & 1.880 & 1.257 & \textbf{0.141} & 88.8\% & 0.351 & 0.270 \\
    43 & 1.543 & 2.819 &  0.592 & \textbf{0.121} & 79.5\% & 0.531 & 0.109 \\
    44 & 2.478 & 2.498 &  0.813 & \textbf{0.419} & 48.5\% & 0.691 & 0.188 \\
    45 & 2.022 & 2.329 &  0.526 & \textbf{0.352} & 33.1\% & 0.620 & 0.328 \\
    46 & 1.424 & 0.713 &  0.478 & \textbf{0.143} & 70.1\% & 0.240 & 0.160 \\
    47 & 1.182 & 1.348 &  0.393 & \textbf{0.246} & 37.5\% & 0.260 & 0.238 \\
    48 & 1.542 & 1.925 &  0.423 & \textbf{0.369} & 12.5\% & 0.783 & 0.505 \\
    49 & 3.218 & 10.47 &  1.311 & \textbf{0.550} & 58\% & 1.324 & 0.428 \\
    \hline
    \hline
    Mean & 2.071 & 2.869 & 0.609 & \textbf{0.281} & 53.9\% & 0.600 & 0.239 \\
    \hline
  \end{tabular}
  \begin{tablenotes}
  \fontsize{8}{10}\selectfont
  \item[\revise{*The trajectory is aligned with GT using \cite{umeyama1991least}}.]
  \end{tablenotes}
\end{threeparttable}
\end{table}

\subsubsection{The importance of uncertainty estimates}
In this section, we inspect the importance of incorporating uncertainty using our MDN model. We take Seq 43 for instance and replace the uncertainty estimation on that sequence with a simple identity matrix, either applied on the odometry or the loop closure constraints. Note that all other setting remain the same. Table \ref{table:importance_uncertainty} shows the output from this experiment. As expected, employing MDN covariance on both odometry and loop closure constraints can significantly increase the performance gain. This happens as our model estimates the uncertainty according to the input images and motion dynamics, which better reflects the actual condition compared to fix (identity) covariance. We can also observe that injecting covariance on the odometry produces better performance than injecting it on the loop constraints. This is possibly because odometry constraints are much denser than the loop closure constraints, in which a better uncertainty estimation on the denser data might lead to easier optimization.

\begin{table}[t]
\centering
\begin{threeparttable}
  \caption{The Impact of Incorporating Covariance from MDN}
  \renewcommand{\arraystretch}{1}
  \setlength\tabcolsep{6pt}
  \fontsize{9}{10}\selectfont
  \label{table:importance_uncertainty}
  \begin{tabular}{|cc|c|c|c|}
    \hline
    \multicolumn{2}{|c|}{Applying MDN covariance on} & Odometry & SLAM & \multirow{2}*{Gain} \\
    Odometry & Loop & ATE & ATE & \\
    \hline
    \hline
    - & - & 0.592 & 0.348 & 41.3\% \\
    \checkmark & - & 0.592 & 0.172 & 71.1\% \\
    - & \checkmark & 0.592 & 0.221 & 62.2\% \\
     \checkmark & \checkmark & 0.592 & 0.121 & 79.5\% \\
    \hline
  \end{tabular}
\end{threeparttable}
\end{table}

\subsubsection{The impact of uncertainty weight and scale}
\begin{figure}
\centering
    \vspace{-0.5cm}
    \includegraphics[width=7.cm,trim=0.5cm .2cm 0.5cm .5cm,clip]{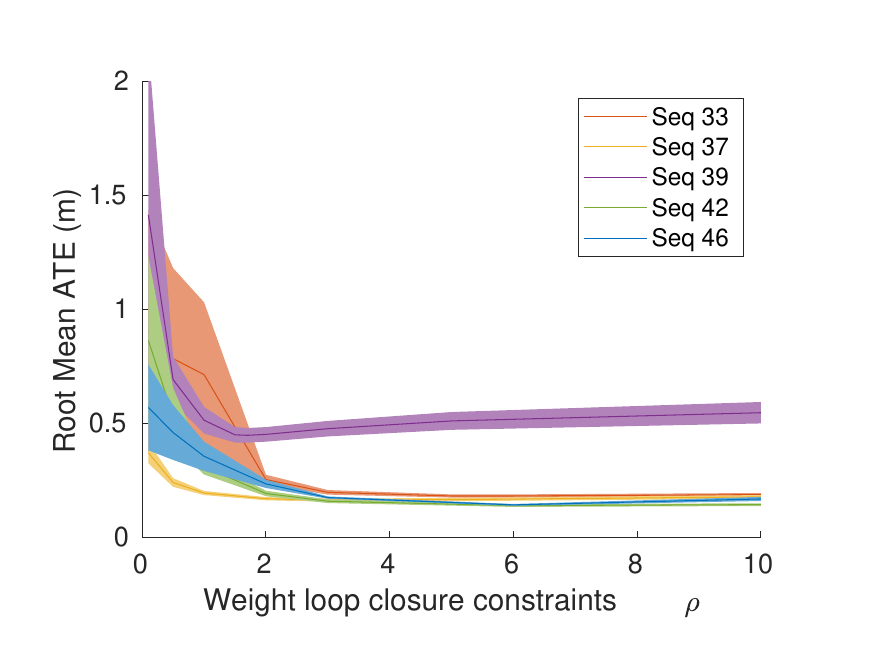}
    \vspace{-0.5cm}
    \caption{The impact of uncertainty weights on Seq 33, 37, 39, 42, and 46. In this experiment, we fix every other parameters except the covariance weight $\rho$ in loop closure constraints.}
\label{fig:ate_vs_weight_uncertainty}
\end{figure}

To understand the importance of balancing the weight between odometry and loop closure constraints by scaling the covariance, we measure the ATE of our model while changing the covariance scale from loop closure constraints $\rho$ and fixing other parameters. Fig. \ref{fig:ate_vs_weight_uncertainty} shows the result of this study. As we have discussed in Section \ref{sec:slam_backend}, in general, we have to set larger weight in loop closure constraints to make the pose graph optimization work and improve the accuracy of odometry estimation. This reflects in Fig. \ref{fig:ate_vs_weight_uncertainty} that the error typically lower when we set $\rho > 2$ and larger when we set $\rho < 2$. However, if we set $\rho$ too large, this also generates sub-optimal performance as the effect of loop closure constraints becomes too dominant (as seen in Seq 39 in Fig. \ref{fig:ate_vs_weight_uncertainty}). Nevertheless, the impact on ATE is not as bad as setting small $\rho$.

\begin{figure*}[!ht]
    \centering
    \vspace{-0.5cm}
    \subfloat[Seq 38]{
        \hspace{-0.5cm}
        \includegraphics[width=4.5cm,trim=1cm .9cm 1cm .5cm,clip]{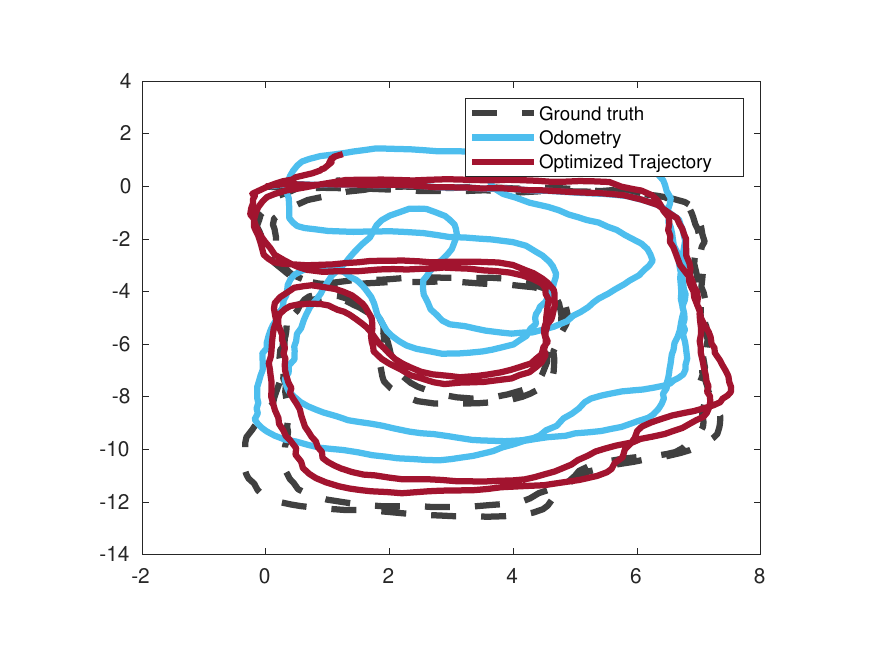}
        \hspace{0cm}}
    \subfloat[Seq 40]{
        \includegraphics[width=4.5cm,trim=1cm .9cm 1cm .5cm,clip]{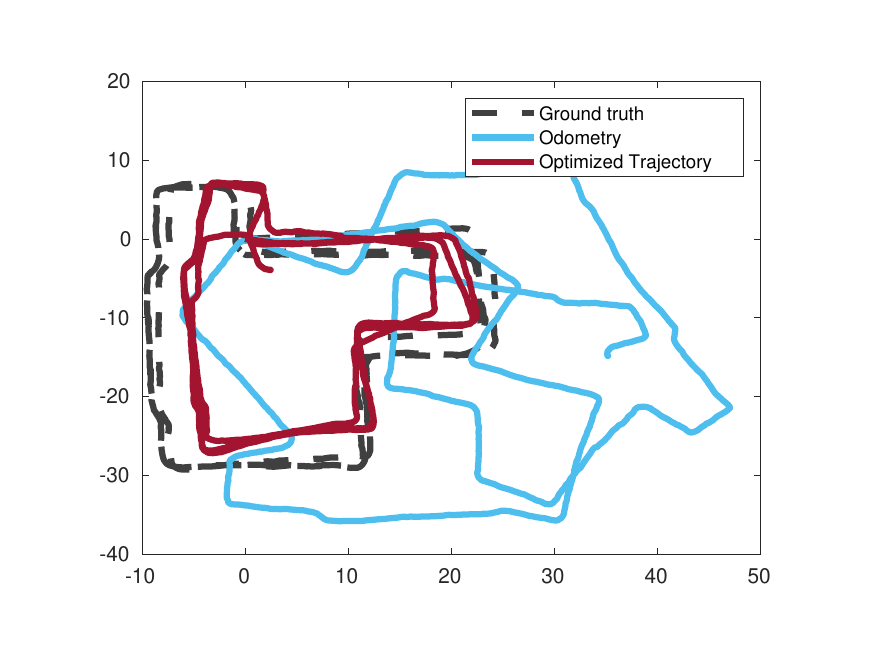}}
        \vspace{0cm}
    \subfloat[Seq 42]{
        \hspace{-0.5cm}
        \includegraphics[width=4.5cm,trim=1cm .9cm 1cm .3cm,clip]{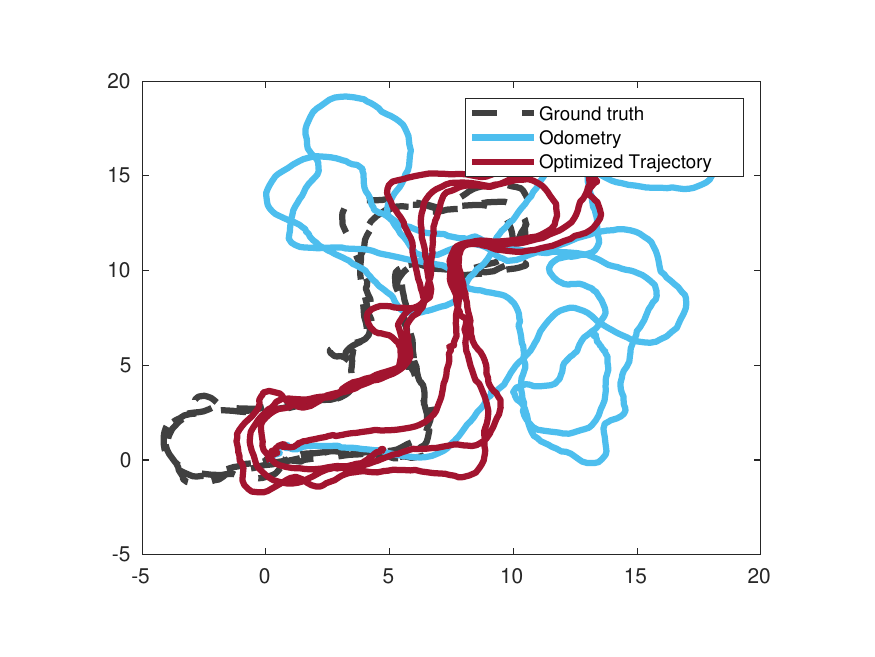}
        \hspace{0cm}}
    \subfloat[Firefighter Facility]{
        \hspace{-0.5cm}
        \includegraphics[width=4.5cm,trim=0.5cm .1cm 1cm 1.5cm,clip]{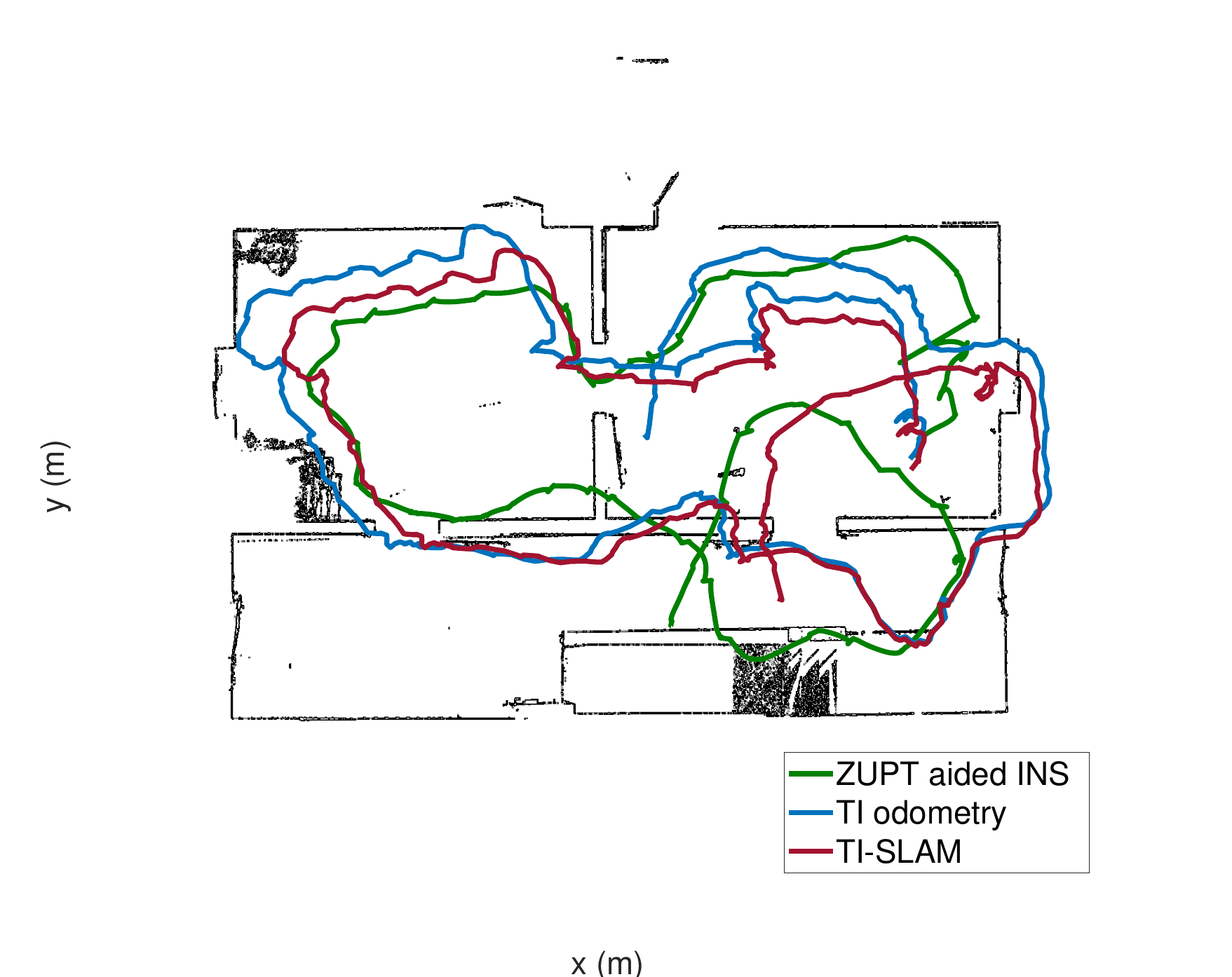}
        \hspace{0cm}}
    \caption{\revise{(a)-(c) Qualitative result of TI-SLAM system in the large scale handheld data. Both odometry and optimized odometry are generated from thermal-inertial data. (d) Test in real emergency scenario with smoke-filled environment. We qualitatively compared TI-SLAM with ZUPT aided INS as RGB, depth, or lidar-based odometry/SLAM system does not work. Note that the floor plan was generated with Lidar SLAM prior to the testing.}}
\label{fig:trajectories_handheld}
\end{figure*}

\subsection{\revise{SLAM Performance in Handheld Data}}
\subsubsection{\revise{SLAM Accuracy}}
\revise{For the evaluation in the handheld data, we fine-tune our neural odometry and neural loop closure in the handheld data and keep the neural embedding network as it is. We need to fine-tune the neural odometry and neural loop closure as the dataset was collected with different sensor placement which impacts the pose estimation accuracy as the extrinsic parameters change. Table \ref{table:evaluation_handheld_ate} lists the ATE of TI-SLAM for all test sequences in handheld data. As we can see, our TI-SLAM produces the most accurate trajectories compared to VINet (RGB), VINet (Thermal), and TI-SLAM. Overall, by incorporating loop closure detection and robust pose graph optimization, we can improve the accuracy of TI odometry for up to $62.76\%$. Fig. \ref{fig:trajectories_handheld} (a)-(c) show some qualitative results from the handheld evaluation.}

\begin{table}[t]
\centering
\begin{threeparttable}
  \renewcommand{\arraystretch}{1}
  \setlength\tabcolsep{3pt}
  \caption{\revise{RMS Absolute Trajectory Errors (m) in Handheld Data}}
  \fontsize{9}{10}\selectfont
  \label{table:evaluation_handheld_ate}
  \begin{tabular}{|cc|cc|ccc|c|}
    \hline
    \multirow{2}*{Seq} & Length & VINet & VINet & TI & \multirow{2}*{TI-SLAM} & Gain \\
     & (m) & (RGB) & (Thermal) & odometry & & ($\%$) \\
    \hline
    \hline
    35 & 142 & 10.964 & 6.999 & 3.272 & \textbf{1.033} & 68.43 \\
    36 & 62 & 2.866 & 2.727 & 1.076 & \textbf{0.661} & 38.64 \\
    37 & 37 & 7.602 & 4.166 & 1.527 & \textbf{0.538} & 64.76 \\
    38 & 104 & 7.788 & 4.754 & 1.641 & \textbf{0.444} & 72.90 \\
    39 & 76 & 5.062 & 1.933 & 2.075 & \textbf{0.921} & 55.57 \\
    40 & 182 & 64.195 & 26.725 & 17.544 & \textbf{4.175} & 76.20 \\
    42 & 115 & 12.815 & 17.696 & 5.472 & \textbf{1.859} & 66.03 \\
    43 & 314 & 9.355 & 11.522 & 2.569 & \textbf{1.038} & 59.58 \\
    \hline
    \hline
    \multicolumn{2}{|c|}{Mean} & 15.081 & 9.565 & 4.397 & \textbf{1.334} & 62.76 \\
    \hline
  \end{tabular}
\end{threeparttable}
\end{table}

\subsubsection{\revise{Test in Smoke-filled Environment}}
\begin{figure}
\centering
    \vspace{-0.8cm}
    \subfloat[Before smoke]{
         \includegraphics[width=4cm,trim=3.8cm 8cm 5cm 8cm,clip]{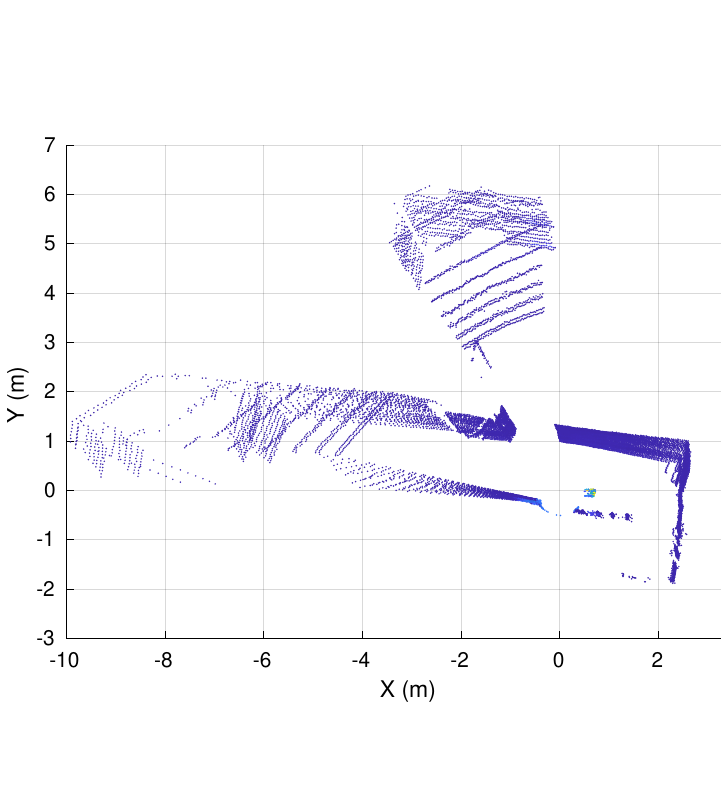}
        }
    \subfloat[After smoke]{
        \includegraphics[width=4cm,trim=3.8cm 8cm 5cm 8cm,clip]{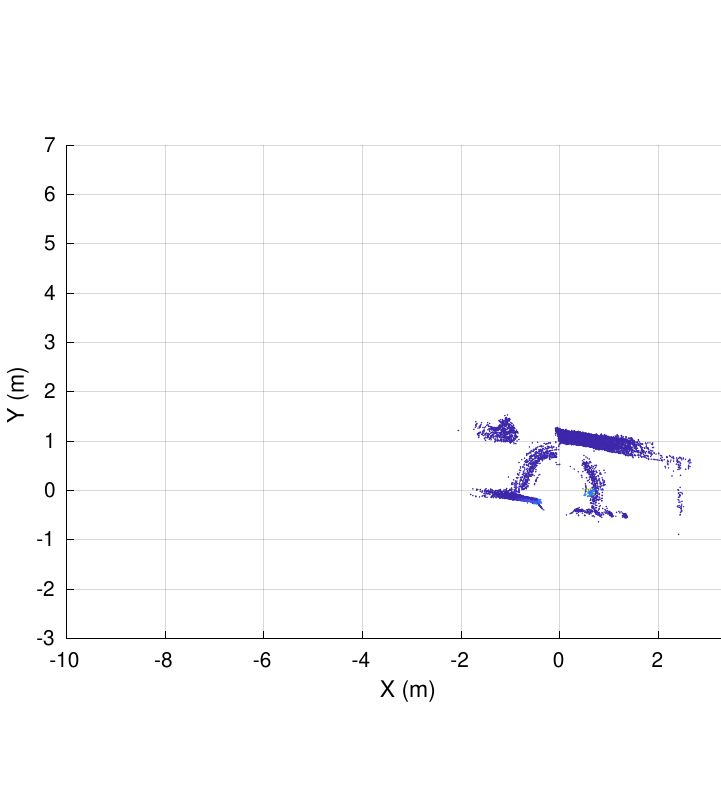}
        }
    \caption{\revise{Lidar point cloud (a) before and (b) after the environment is filled with smoke. As can be observed from (b) that the smoke blocks the lidar signal, creating barrier in front of the device.}}
\label{fig:lidar_smoke}
\end{figure}

\revise{For the evaluation in smoke-filled environment, we compare our approach with a zero-velocity-aided (ZUPT) Inertial Navigation System (INS) \cite{wahlstrom2019zero} as none of RGB- and Lidar-based odometry/SLAM can work \cite{bijelic2019seeing}. As can be seen from Fig. \ref{fig:lidar_smoke} (a) and (b), the smoke blocks the lidar signal, creating a half-sphere barrier in front of the device and degrading the lidar odometry/SLAM algorithm\footnote{https://www.youtube.com/watch?v=EZ1gpetEN8c}. In that sense, without the availability of (pseudo) ground truth, only qualitative results are provided. Nevertheless, we can see from Fig. \ref{fig:trajectories_handheld} (d) that our TI-SLAM produces similar trajectory with ZUPT aided INS. It is good to note that the loop closure detection plays important roles of correcting the drift of TI odometry.}

\subsection{\revise{SLAM Performance in SubT-tunnel Data}}
\begin{figure*}[!ht]
    \centering
    \vspace{-0.2cm}
    \hspace{0cm}
    \includegraphics[width=4.4cm,trim=0.1cm .1cm 0.1cm .5cm,clip]{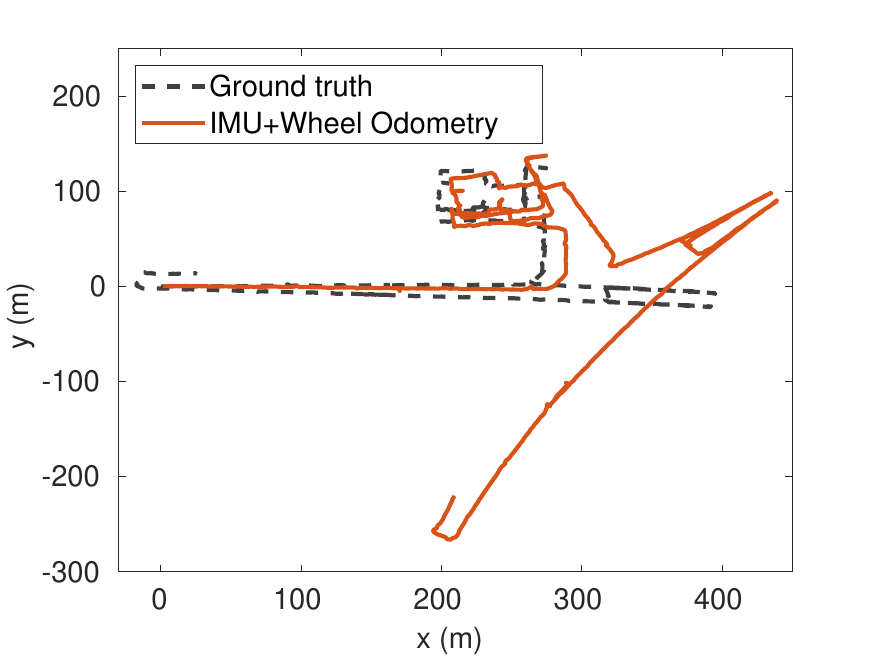}
    \includegraphics[width=4.4cm,trim=0.1cm .1cm 0.1cm .5cm,clip]{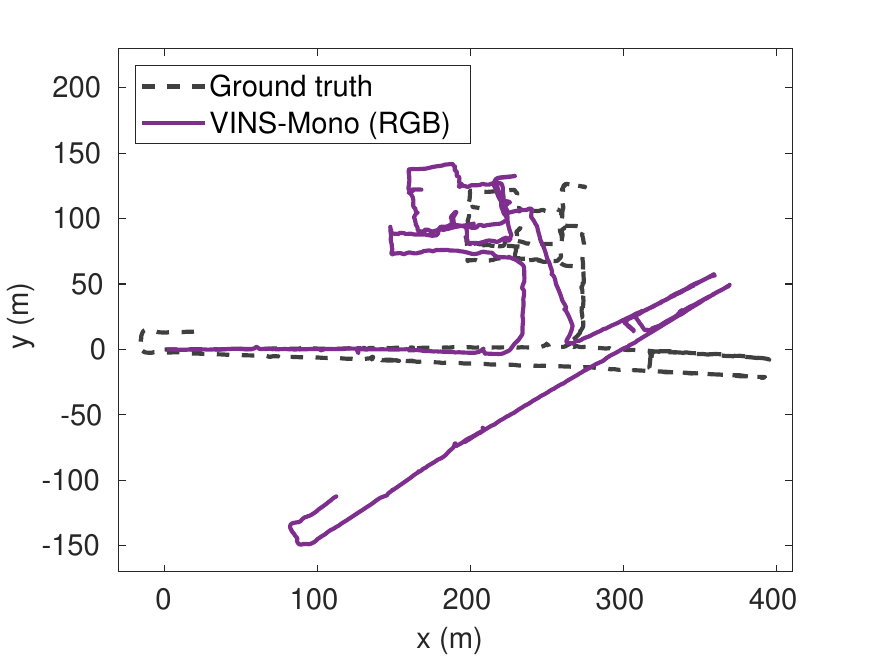}
    \includegraphics[width=4.4cm,trim=0.1cm .1cm 0.1cm .5cm,clip]{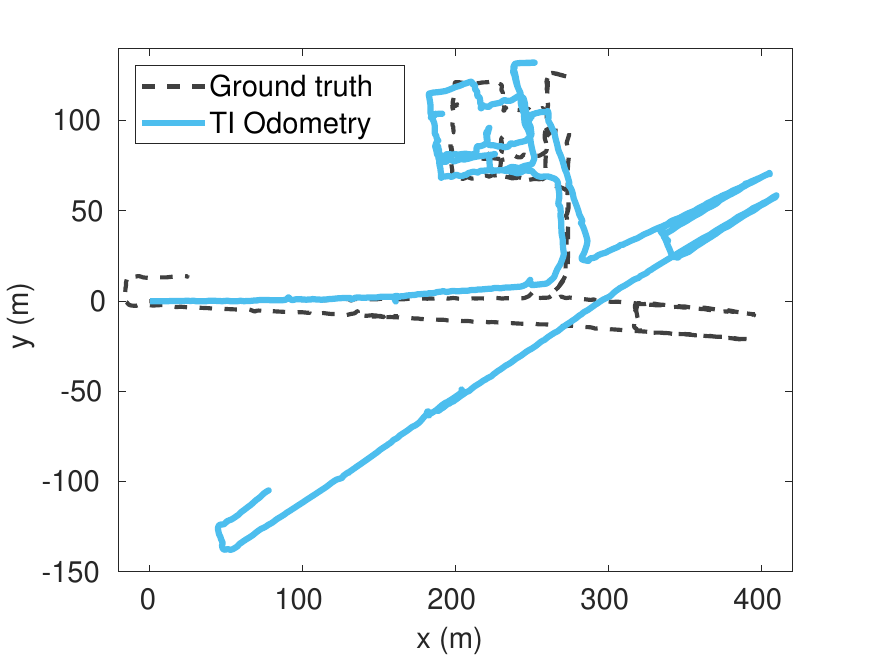}
    \includegraphics[width=4.4cm,trim=0.1cm .1cm 0.1cm .5cm,clip]{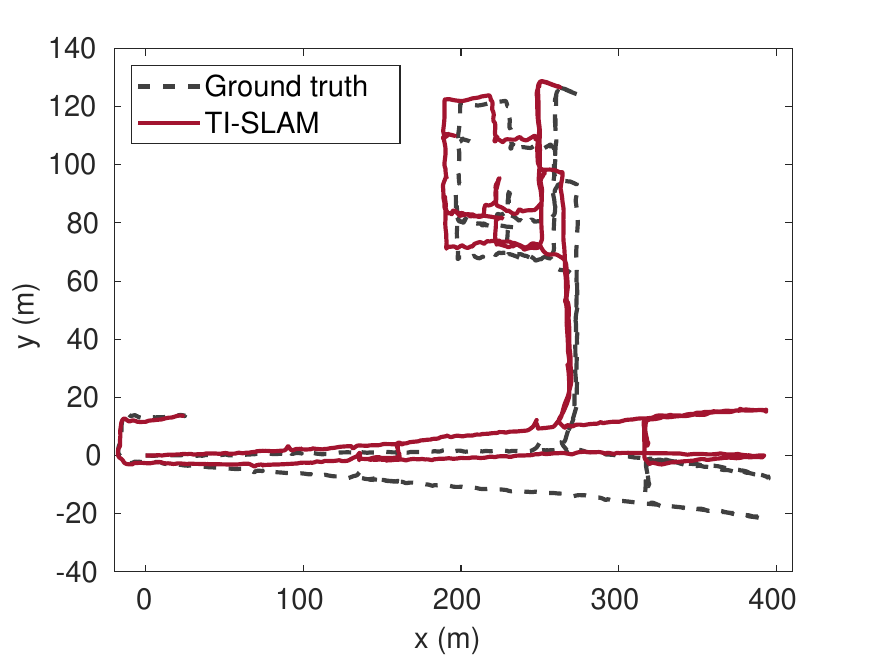}
    \vspace{-0.2cm}
    \caption{\revise{Test in a long mission (54 minutes) 
    in \texttt{ex\_B\_route1} sequence from SubT-tunnel dataset. The trajectory generated by inertial+wheel, VINS-Mono (RGB), TI-odometry (ours), and TI-SLAM (ours) are depicted.}}
\label{fig:trajectories_subt}
\end{figure*}

\revise{For the experiment in SubT-tunnel dataset, we use our in-house ground robot as the base model and fine tune the SLAM front end in sequences \texttt{sr\_B\_route1} and \texttt{sr\_B\_route2}. The output trajectory and the ATE can be seen from Fig. \ref{fig:trajectories_subt} and Table \ref{table:evaluation_subt_ate}. For comparison, we provide the result from IMU assisted wheel odometry, VINet (thermal), VINS-Mono (RGB) \cite{qin2018vins}, and our TI odometry. Despite the fact that in some areas the tunnel has no illumination (complete darkness), we still can utilize VINS-Mono (RGB) as the robot is equipped with four LED illuminators. On the other hand, VINS-Mono (thermal), either using 8-bit or 14-bit representation, loses tracks after running for about a minute or two due to abrupt motion, the lack of thermal features, and low frame rate (10Hz).

As you can see in Table \ref{table:evaluation_subt_ate}, compared to the state-of-the-art visual-inertial odometry and SLAM algorithms (e.g., VINet, VINS-Mono), TI-SLAM produces more consistent trajectory for a long mission (54 minutes) in poorly illuminated scene with diverse motion types (e.g., U-turn, stop motion, etc.). TI-SLAM is even much better than IMU assisted wheel odometry which typically performs better than our model in shorter indoor mission as can be seen in Table \ref{table:evaluation_robot_ate}. Our loop closure detection and loop closure constraints estimation again provide an important role as TI odometry drift significantly in the long mission. By closing the loop, the accuracy can be improved by around $40\%$ in \texttt{ex\_B\_route1} sequence. Nevertheless, it is good to note that our approach is running offline while VINS-Mono has to respect realtime constraints.}

\begin{table}[t]
\centering
\begin{threeparttable}
  \renewcommand{\arraystretch}{1}
  \setlength\tabcolsep{1pt}
  \caption{\revise{RMS Absolute Trajectory Errors (m) in a Long Mission in SubT-tunnel Data}}
  \fontsize{9}{10}\selectfont
  \label{table:evaluation_subt_ate}
  \begin{tabular}{|c|ccc|cc|}
    \hline
     & VINet & IMU+ & VINS-Mono & TI & TI-SLAM \\
     & (thermal) & Wheel & (RGB) & Odometry &  \\
    \hline
    \hline
    Mean & 156.41 & 42.276 & 30.242 & 32.048 & \textbf{19.277} \\
    Std. Dev. & 88.948 & 48.315 & 20.723 & 17.184 & \textbf{10.909} \\
    \hline
  \end{tabular}
\end{threeparttable}
\end{table}

\section{\revise{Lesson Learnt, Limitations, and Future Work}}

\revise{Despite the fact that TI-SLAM can work well in our test scenarios, there are limitations and lessons learnt that can be used as a ground for future exploration. \revisetwo{First, for thermal-inertial odometry, accurate scale estimation remains a challenge in some scenarios and pose graph SLAM cannot completely fix it.} This problem becomes apparent especially in longer mission as depicted in hand-held (Fig. \ref{fig:trajectories_handheld} (c)) and SubT-tunnel experiment (Fig. \ref{fig:trajectories_subt}). Incorporating range sensor like millimeter wave radar \cite{lu2020milliego} can be potentially used to alleviate this problem. 

Second, the framework currently operates in offline fashion as real-time performance remain an open problem for deep networks, especially when it is executed in standard CPU. In real-time scenarios, deep network compression and acceleration \cite{saputra2019distilling, choudhary2020comprehensive} can be used in the future investigation. 

Third, as TI-SLAM brings together deep learning approach and conventional pose graph optimization, tuning hand-crafted parameters are required in the SLAM back end, e.g., scaling covariance between odometry and loop closure constraints, choosing a threshold to reject false positive loop constraints, etc. In this case, typically, there are no general hand-crafted parameters that can maximally perform for individual sequence. For online operation, parameters that generate the best average result are usually used while for offline operation, these parameters can be tuned individually for each sequence. In that sense, training a model that can automatically predict the back end parameters would be a viable future direction. Finally, domain adaptation also remains a challenge for deep odometry network. While the embedding network is usually more generalized in cross domain scenarios (e.g., the neural embedding network trained in ground robot data can be directly used for testing in handheld data), odometry and loop closure network need to be tuned for different domain as the intrinsic and extrinsic parameters between sensors might change. Designing a DNN model that can learn to estimate these intrinsic and extrinsic parameters during operation can be an interesting topic for future research direction.}

\section{Conclusion}
\label{sec:conclusion}
In this paper, we have demonstrated the first complete thermal-inertial SLAM system. Our key approach enabling full thermal-inertial SLAM is the usage of probabilistic neural networks to abstract noisy sensor data such that it will be more amenable for SLAM inference. \revise{By combining this neural abstraction in the SLAM front end with a robust graph-based optimization in the SLAM back end, we can generate an accurate trajectory estimation for different scenarios including handheld (firefighting) and ground robot motion in indoor and underground tunnel}. Future research directions include designing online thermal-inertial SLAM system such that it can work within real time constraints and incorporating a range-based sensor to produce more accurate results in arbitrary environments.


{
\appendix[Training Details]
\subsection{Neural Thermal-Inertial Odometry}
As we have mentioned in Sec. \ref{sec:tio_learning_mechanism}, to train neural thermal-inertial odometry, we use Eq. (\ref{eq:hallucination_loss}) as the objective function in the first stage of training. Adam optimizer with a 0.0001 learning rate is used for maximum of 200 epochs during this training process. \revise{Before training, we normalize the input 14-bit radiometric data (into between 0 and 1) by using the maximum and minimum radiometric value extracted from the training dataset. We then subtract it with the mean over the training dataset.} We randomly cut the training sequence into small batches of consecutive pair ($n = 8$) to obtain better generalization. We also sub-sample the input such that we operate on around 5 fps to provide sufficient parallax between consecutive frames. In the second stage, we continue to train the network by using Eq. (\ref{eq:mdn_pose_loss}) for 200 epochs with RMSProp. We set 0.001 for the initial learning rate and then drop it by $25\%$ after every 25 epochs. We also follow this procedure to train neural loop closure network.

\subsection{Neural Embedding Network}
For the neural embedding network, we train the network using Eq. \ref{eq:triplet_loss} as the objective for a maximum of 200 epochs using Adam optimizer with 0.0001 initial learning rate. We set the threshold of adjacent frames empirically as 18. \revise{Before training, we normalize the raw 14-bit radiometric data (into between 0 and 1) using the maximum and minimum radiometric value extracted from the training dataset. We then convert back the normalized radiometric data into a grayscale image (8-bit) and copy the channels into three such that it can replicate the standard RGB images consumed by the ResNet50.} We only use a batch size of 3  during training to fit it in our GPU.
}

\bibliographystyle{IEEEtran}
\bibliography{IEEEabrv,references}


\begin{IEEEbiography}[{\includegraphics[width=1in,height=1.25in,clip,keepaspectratio]{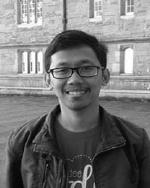}}]{Muhamad Risqi U. Saputra} is currently an Assistant Professor in Data Science at Monash University, Indonesia. Previously, he was a postdoctoral research associate in Computer Science Department, University of Oxford, UK. He also obtained his DPhil degree from Oxford. Before coming to Oxford, he received his bachelor and master degrees from the Department of Electrical Engineering and Information Technology, Universtas Gadjah Mada, Indonesia. His main research interests revolves around machine learning, computer vision, and cyber-physical systems.
\end{IEEEbiography}

\begin{IEEEbiography}[{\includegraphics[width=1in,height=1.25in,clip,keepaspectratio]{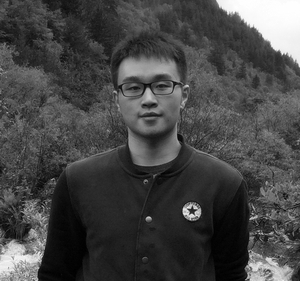}}]{Chris Xiaoxuan Lu}
is currently an Assistant Professor (UK Lecturer) in the School of Informatics at University of Edinburgh. Before that he did both PhD study and post-doctoral research in the Department of Computer Science, University of Oxford. His research interest lies in Cyber Physical Systems, Robotics and Autonomous Systems and Artificial Intelligence of Things (AIoT).
\end{IEEEbiography}

\begin{IEEEbiography}[{\includegraphics[width=1in,height=1.25in,clip,keepaspectratio]{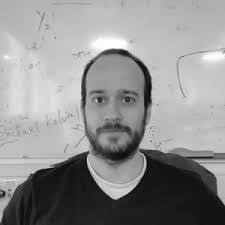}}]{Pedro Porto B. de Gusmao}
currently holds a senior research associate position at the Department of Computer Science and Technology, University of Cambridge, UK. Previously, he was a postdoctoral researcher at the Cyber-Physical Systems group, University of Oxford. He received bachelors degree in Telecommunication Engineering from the University of Sao Paulo and masters degree on the same field from the Politecnico di Torino in a double degree program. In 2017, he obtained his PhD from the same Politecnico. His research interests include computer vision, machine learning and signal processing.
\end{IEEEbiography}

\begin{IEEEbiography}[{\includegraphics[width=1in,height=1.25in,clip,keepaspectratio]{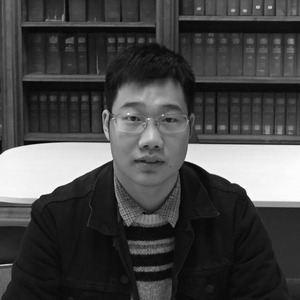}}]{Bing Wang}
is currently PhD student at Department of Computer Science, University of Oxford. Before that, he obtained his BEng Degree at Shenzhen University, China. His research interest lies in camera localization, feature detection, description and matching, and cross-domain representation learning.
\end{IEEEbiography}

\begin{IEEEbiography}[{\includegraphics[width=1in,height=1.25in,clip,keepaspectratio]{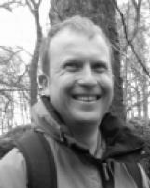}}]{Andrew Markham} is an Associate Professor at the Department of Computer Science, University of Oxford. He obtained his BSc (2004) and PhD (2008) degrees from the University of Cape Town, South Africa. He is the Director of the MSc in Software Engineering. He works on resource-constrained systems, positioning systems, in particular magneto-inductive positioning and machine intelligence.
\end{IEEEbiography}

\begin{IEEEbiography}[{\includegraphics[width=1in,height=1.25in,clip,keepaspectratio]{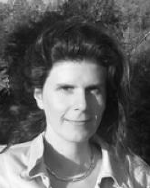}}]{Niki Trigoni}
is a Professor at the Department of Computer Science, University of Oxford. She is currently the director of the EPSRC Centre for Doctoral Training on Autonomous Intelligent Machines and Systems, and leads the Cyber Physical Systems Group. Her research interests lie in intelligent and autonomous sensor systems with applications in positioning, healthcare, environmental monitoring and smart cities.
\end{IEEEbiography}
 




\vfill

\end{document}